\documentclass[10pt,journal,compsoc]{IEEEtran}
\usepackage{amsmath,amsfonts}
\usepackage{multirow}
\usepackage{array}
\usepackage[caption=false,font=normalsize,labelfont=sf,textfont=sf]{subfig}
\usepackage{textcomp}
\usepackage{stfloats}
\usepackage{url}
\usepackage{verbatim}
\usepackage{wrapfig}
\usepackage{pgfplots}
\usepackage{enumitem, multirow, xspace}
\usepackage{graphicx}
\usepackage{amsmath}
\usepackage{booktabs}
\usepackage{multirow}
\usepackage{autobreak}
\usepackage{makecell}
\usepackage{cancel}
\usepackage{hyperref}
\usepackage[ruled,vlined]{algorithm2e}
\usepackage[numbers, sort]{natbib}
\usepackage[commandnameprefix=ifneeded]{changes}
\usepackage{etoolbox}
\usepackage{amsmath}
\usepackage{amssymb}
\usepackage{mathtools}
\usepackage{amsthm}
\usepackage{bbding}
\usepackage{threeparttable}
\usepackage{colortbl}
\usepackage{color}
\usepgfplotslibrary{groupplots}

\theoremstyle{plain}

\theoremstyle{definition}

\theoremstyle{remark}

\makeatletter

\begin{document}

\title{Monte Carlo Neural PDE Solver for Learning PDEs via Probabilistic Representation}

\author{Rui Zhang, Qi Meng, Rongchan Zhu, Yue Wang, Wenlei Shi, Shihua Zhang, \\  Zhi-Ming Ma, Tie-Yan Liu,~\IEEEmembership{Fellow,~IEEE}
\thanks{Corresponding author: Qi Meng, Email: meq@amss.ac.cn}
\thanks{Rui Zhang is with Gaoling School of Artificial Intelligence, Renmin University of China, Beijing, 100872, China. Email: rayzhang@ruc.edu.cn.}
\thanks{Qi Meng, Shihua Zhang and Zhi-Ming Ma are with the Academy of Mathematics and Systems Science, Chinese Academy of Sciences (CAS), Beijing 100190, China. E-mail: \{meq, zsh\}@amss.ac.cn, mazm@amt.ac.cn.}
\thanks{Rongchan Zhu is with Beijing Institute of Technology, Beijing 100081, China. E-mail: zhurongchan@126.com.}
\thanks{Yue Wang and Tie-Yan Liu are withZhongguancun Academy, Beijing 100094, China. E-mail: is.yuewang@gmail.com, tie-yan.liu@outlook.com.}
\thanks{ Wenlei Shi is with Institute of Computing Technology, Chinese Academy of Sciences, Beijing 100190, China. E-mail: shiwenlei22b@ict.ac.cn}}

\markboth{IEEE Trans on Pattern Analysis and Machine
Intelligence,~Vol.~xx, No.~x, 2025}%
{Shell \MakeLowercase{\textit{et al.}}: MCNP}


\IEEEtitleabstractindextext{
\begin{abstract}
In scenarios with limited available data, training the function-to-function neural PDE solver in an unsupervised manner is essential. However, the efficiency and accuracy of existing methods are constrained by the properties of numerical algorithms, such as finite difference and pseudo-spectral methods, integrated during the training stage. These methods necessitate careful spatiotemporal discretization to achieve reasonable accuracy, leading to significant computational challenges and inaccurate simulations, particularly in cases with substantial spatiotemporal variations. To address these limitations, we propose the Monte Carlo Neural PDE Solver (MCNP Solver) for training unsupervised neural solvers via the PDEs' probabilistic representation, which regards macroscopic phenomena as ensembles of random particles. Compared to other unsupervised methods, MCNP Solver naturally inherits the advantages of the Monte Carlo method, which is robust against spatiotemporal variations and can tolerate coarse step size. In simulating the trajectories of particles, we employ Heun's method for the convection process and calculate the expectation via the probability density function of neighbouring grid points during the diffusion process. These techniques enhance accuracy and circumvent the computational issues associated with Monte Carlo sampling. Our numerical experiments on convection-diffusion, Allen-Cahn, and Navier-Stokes equations demonstrate significant improvements in accuracy and efficiency compared to other unsupervised baselines.
\end{abstract}

}

\maketitle
\section{Introduction}
\IEEEPARstart{S}{olving} partial differential equations (PDEs) is essential for understanding and modeling various physical phenomena, including fluid dynamics~\cite{anderson1995computational}, heat transfer~\cite{rohsenow1998handbook}, and quantum mechanics~\cite{howe1980quantum}. Traditional numerical methods for PDEs, such as finite difference, finite element, and spectral methods, have achieved remarkable successes and are widely used across various industries~\cite{larsson2003partial}. However, they still have certain limitations. For instance, these methods often require precise spatiotemporal discretization, which can be computationally expensive, particularly in scenarios with high spatiotemporal variability~\cite{courant1967partial}. Additionally, they may struggle with complex geometries and multiscale problems, where the need for specialized mesh designing and treatment can complicate the simulation process and increase user difficulty~\cite{huang2010adaptive, weinan2011principles}.

Recently, deep learning (DL) methods have emerged as a promising approach to addressing scientific computation challenges for PDE solving, offering a new perspective beyond traditional numerical techniques~\cite{karniadakis2021physics}. By harnessing the representation power of deep neural networks, DL-based techniques have successfully overcome several limitations inherent in classical numerical strategies. For instance, physics-informed neural networks (PINNs)~\cite{raissi2019physics, doi:10.1126/science.aaw4741} and deep energy methods (DEM)~\cite{samaniego2020energy, nguyen2021parametric} respectively utilize the strong or weak form of PDEs to construct loss functions, providing flexibility in handling complex boundary problems without the need for mesh discretization. Additionally, DL-based reduced-order modeling (ROM)~\cite{chen2021physics, fresca2021comprehensive} improves the accuracy and efficiency of traditional ROM techniques, particularly in tackling nonlinear and multiscale problems. However, these methods are typically designed for PDEs with fixed initial or forcing fields, necessitating retraining of the neural networks when handling new initial fields.

Beyond these achievements, the function-to-function neural PDE solvers have risen as another new paradigm for simulating physical systems, which leverage neural networks as surrogate models to approximate the solutions of a family of PDEs~\cite{sanchez2020learning, wandel2021learning,lu2021learning, DBLP:conf/iclr/LiKALBSA21}. Along this direction, several studies have proposed diverse network architectures for neural PDE solvers~\cite{sanchez2020learning,li2020neural,lu2021learning,brandstetter2022message,zhang2024deciphering}. These solvers can be trained using supervised~\cite{sanchez2020learning,lu2021learning, li2020neural} or unsupervised approaches~\cite{wandel2021learning,wang2021learning,li2021physics}, employing pre-generated data or PDE information to construct training targets, respectively. The unsupervised training approach is essential for DL-based PDE solvers, particularly in scenarios with limited available or high-quality data. To address this issue, some studies~\cite{wandel2021learning,shi2022lordnet,li2021physics} borrow techniques from classical numerical solvers to construct training targets. For instance, the MAC grid physics-constrained network and its 3D extension~\cite{wandel2021learning, wandel2021teaching}, low-rank decomposition network (LordNet)~\cite{shi2022lordnet} and physics-informed neural operator (PINO)~\cite{li2021physics} integrate finite difference or pseudo-spectral methods with neural networks during the training stage. However, these traditional Eulerian methods require fine meshes or step sizes for stable simulations. Therefore, the performance and efficiency of corresponding neural PDE solvers are also limited by time and space discretization, particularly when handling high spatiotemporal variations. Furthermore, these methods often necessitate additional loss functions to enforce boundary conditions, introducing extra complexity in hyperparameter tuning and computational demands.

To this end, we introduce the \textbf{Monte Carlo Neural PDE Solver (MCNP Solver)}, a novel approach for training neural solvers from a probabilistic perspective, which views macroscopic phenomena as ensembles of random movements of microscopic particles~\cite{yang1949kinetic}. For a PDE system with probabilistic representation, we construct the loss function of the MCNP Solver between two sequential PDE fields $u_t$ and $u_{t+\Delta t}$ via the relationship derived from the Monte Carlo approximation. To ensure the efficiency and accuracy of the MCNP Solver, we develop several techniques when combining the Monte Carlo approximation with the training of the deep neural network. Specifically, in simulating the corresponding stochastic difference equations (SDE), we use the Heun's method to obtain a more accurate result when handling the convection process. During the diffusion process, we approximate the mathematical expectation via the probability density function (PDF) of neighbouring grid points to eliminate sampling a large number of particles in Monte Carlo methods. Compared to other unsupervised neural solvers, such as LordNet~\cite{shi2022lordnet} and PINO~\cite{li2021physics}, the MCNP Solver naturally inherits the advantages of Monte Carlo methods. It can tolerate coarse step size~\cite{giraldo1997comparison,mimeau2021review}, thereby reducing training costs and accumulated errors arising from temporal discretization. Additionally, it can efficiently handle high-frequency spatial fields due to its derivative-free property~\cite{10.2307/2098715, Maire2012MonteCA}. Moreover, the boundary conditions are automatically encoded into the stochastic process of particles~\cite{10.2307/2098715, Maire2012MonteCA}, eliminating the need to introduce extra loss terms to satisfy such constraints. 

In summary, we make the following contributions:

1. We propose the MCNP Solver, an innovative unsupervised approach for training neural solvers that can be applied to PDE systems with probabilistic representation. We also devise several strategies to boost performance and efficiency during the convection and diffusion processes in SDE simulations.

2. Our experiments involving the convection-diffusion, Allen-Cahn, and Navier-Stokes equations demonstrate significant improvements in accuracy and efficiency over other unsupervised neural solvers, particularly for simulation tasks with complex spatiotemporal variations and coarse step sizes. Furthermore, we conduct experiments to solve 2D fractional diffusion on a disk, extending the MCNP Solver to mesh-free and fractional Laplacian scenarios.

3. Beyond comparisons with unsupervised learning methods, we comprehensively compare MCNP Solver with other widely-used PDE solvers, including the Eulerian solver, Monte Carlo methods, and the supervised training approach. We delve into a detailed discussion on each method's strengths, weaknesses, and application scopes.

The structure of this paper is as follows: Section~\ref{sec:related} introduces related works on two types of neural PDE solvers. Section~\ref{sec:method} provides a detailed overview of the probabilistic representation of the PDEs and the proposed MCNP Solver. Sections~\ref{sec:exp} and \ref{sec:addexp} present the experimental results of the MCNP Solver in comparison with unsupervised learning methods and other popular PDE solvers, respectively. Finally, Section~\ref{sec:future} summarizes our method, discusses its limitations and outlines potential future work.

\section{Related Work}\label{sec:related}

In this section, we introduce two primary categories of neural PDE solvers as follows. The first category is designed to learn a location-to-value mapping for a specific PDE, such as PINN~\cite{raissi2019physics}. The second category is designed to learn a function-to-function mapping for a family of PDEs, such as Fourier neural operator (FNO) and DeepONet~\cite{DBLP:conf/iclr/LiKALBSA21, lu2021learning}. In this paper, we target the second task and aim to learn neural PDE solvers between functional spaces that can generalize to different PDE conditions over a distribution.

\subsection{The Location-to-Value Neural PDE Solver} 
The location-to-value PDE solver utilizes the neural network to approximate the solution for a specific PDE with fixed initial and boundary conditions. The input of the neural network is the location and time $(\boldsymbol{x}, t)$, and the output is the corresponding solution $u(\boldsymbol{x}, t)$. To this end, PINNs have been proposed to construct the loss function using the equation and boundary residuals via the automatic differentiation regime. They are widely employed for solving forward or inverse problems~\cite{raissi2019physics,chen2021physics,karniadakis2021physics,qzhao2022graphpde}. Recently, PINNs have made significant progress in addressing scientific problems based on PDEs, such as Navier-Stokes equations~\cite{doi:10.1126/science.aaw4741,margenberg2022neural}, Schrödinger equations~\cite{hermann2020deep, li2021neural}, Allen-Cahn equations~\cite{mattey2022novel, karlbauer2022composing}. Beyond the original PINNs, several methods have been proposed further to enhance the solver accuracy and efficiency of PINNs. For instance, by incorporating more advanced optimization algorithms, the convergence speed of PINNs can be significantly accelerated~\cite{JIN2021109951, lu2022sobolev}. Also, several improved architectures can boost PINNs' capacity to fit high-frequency signals~\cite{CiCP-28-1970, bu2021quadratic}. Some works have sought to improve the performance of PINNs through the design of novel activation functions~\cite{gnanasambandam2023self}, adaptive collocation strategies~\cite{anitescu2019artificial} and adaptive hyperparameter selection methods~\cite{xiang2022self}. Instead of PINNs, some works utilize the probabilistic representation to train neural networks~\cite{han2020derivative, guo2022monte, zhang2022drvn}, which can efficiently handle high-dimensional or fractional PDEs~\cite{han2018solving, richter2021solving, guo2022monte,richter2022robust,nusken2021interpolating}. Furthermore, some studies design loss functions based on other numerical methods, such as the finite volume method~\cite{bezgin2021data}, finite element method~\cite{mitusch2021hybrid, pantidis2023integrated}, and energy-based method~\cite{samaniego2020energy, nguyen2021parametric, wang2022cenn}. Notably, the aforementioned location-to-value methods require retraining neural networks when encountering a PDE with new initial conditions, which can be time-consuming. In this paper, we aim to learn a function-to-function PDE solver that can generalize over a distribution. 

\subsection{The Function-to-Function Neural PDE Solver}
This kind of neural PDE solver has been proposed to learn mappings between functional spaces, such as mapping a PDE's initial condition to its solution~\cite{lu2021learning}. Works like DeepONet~\cite{lu2021learning} and its variants~\cite{seidman2022nomad,venturi2023svd,lee2023hyperdeeponet} encode the initial conditions and queried locations using branch and trunk networks, respectively. Additionally, Fourier Neural Operator (FNO)~\cite{DBLP:conf/iclr/LiKALBSA21} and its variants~\cite{li2022geofno,rahman2023uno,tran2023factorized} explore learning the operator in Fourier space, an efficient approach for handling different frequency components. {Several studies have employed graph neural networks~\cite{ummenhofer2019lagrangian, sanchez2020learning,li2020neural, brandstetter2022message} or transformers~\cite{cao2021choose, li2023transformer} as the backbone models of neural solvers to adapt to complex geometries.} However, these methods require the supervision of ground truth data generated via accurate numerical solvers, which can be time-consuming in general. {To this end, some studies aim to train the neural PDE solvers without the supervision of data~\cite{wandel2021learning,wang2021learning,shi2022lordnet,li2021physics}.} For example, \cite{wang2021learning} proposed PI-DeepONets, which utilize the PDE residuals to train DeepONets in an unsupervised way. Similarly, \cite{huang2022metaautodecoder} proposed Meta-Auto-Decoder, a meta-learning approach to learn families of PDEs in the unsupervised regime. {Furthermore, the MAC grid physics-constrained network~\cite{wandel2021learning}, LordNet~\cite{shi2022lordnet}, and PINO~\cite{li2021physics} borrow techniques from traditional Eulerian methods, such as the finite difference and pseudo-spectral methods, and utilize the corresponding residuals as training loss, respectively. In addition to the above approaches, several advanced methods have been proposed to further improve the accuracy of spatiotemporal derivative calculations in unsupervised solver learning methods. For instance, the spline-PINN methods~\cite{wandel2022spline, sun2022bayesian} use Hermite spline kernels to interpolate the spatiotemporal field continuously, and the SP-PINN~\cite{navaneeth2023stochastic} and its extension~\cite{navaneeth2024physics} employ stochastic projection to enhance the precision of spatial gradient computation. Compared to these unsupervised methods, the MCNP Solver integrates physical knowledge via a novel probabilistic perspective, leveraging the strengths of Lagrangian approaches. This approach allows the neural PDE solver to exhibit robustness against spatiotemporal variants and tolerate coarse step size. Also, the boundary conditions and geometric shapes of the PDEs are naturally embedded into the particles' random walk, eliminating the need for extra constraints in the loss function. Finally, the MCNP Solver can efficiently handle fractional Laplacian operators~\cite{kozubowski2006fractional}, addressing a gap in the existing unsupervised methods.}

\section{Methodology}\label{sec:method}
This section introduces the methodology part of the MCNP Solver. Section~\ref{sec:fk} presents the Monte Carlo method and its corresponding theory. Section~\ref{sec:mcloss} presents the overall framework of the neural Monte Carlo loss, and Section~\ref{sec:sde} provides corresponding details in simulating the convection and diffusion processes in neural Monte Carlo loss.

\subsection{Preliminary}\label{sec:fk}

\begin{figure*}[!t]
\begin{center}
\centerline{\includegraphics[width=16cm]{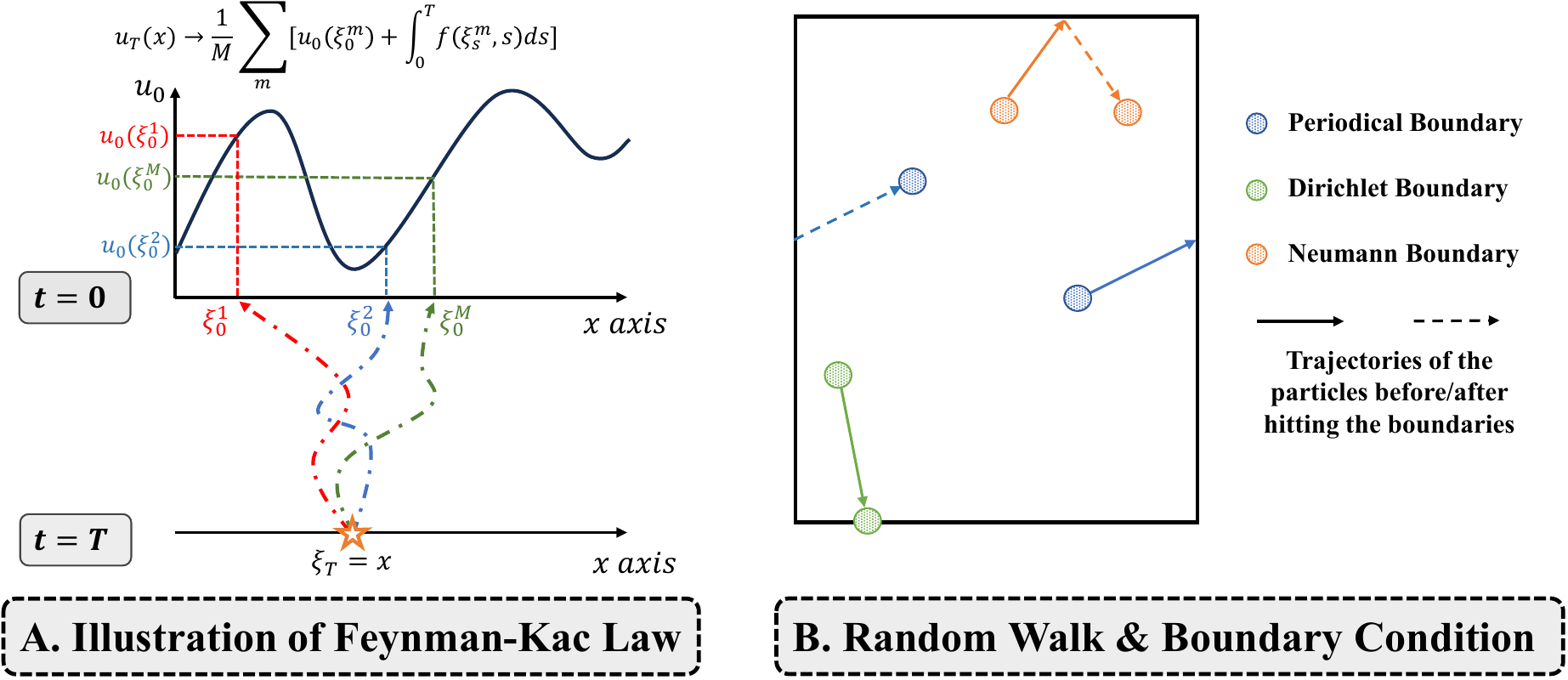}}
\caption{{\textbf{{Illustration of Feynman-Kac law and the random walks of particles when hitting different boundaries.}} \textbf{A: }$M$ particles $\{{\boldsymbol{\xi}}_{s}^m\}_{m=1}^M$ begin at the location $\boldsymbol{x}$, and conduct the random walk according to Eq.~\ref{eq:bsde} from $t=T$ to $t=0$. When $t=0$, the particles query the value at $u_0$, and their average value can be considered as the approximation of $u_T(x)$ (Eq.~\ref{eq:proba_rep_u}). \textbf{B: } Monte Carlo methods can naturally encode boundary conditions in the random walk of particles. For periodical/Dirichlet/Neumann boundary conditions, the random walks of particles need to be pulled back/stopped/reflected when hitting the boundaries, respectively.}}
\label{fk}
\end{center}
\vspace{-10pt}
\end{figure*}

In this paper, we consider the general convection-diffusion equation defined as follows:
\begin{equation}\label{eq:main_pde}
\begin{aligned}
    &\frac{\partial u}{\partial t} = \boldsymbol{\beta}[u]( \boldsymbol{x}, t)\cdot {\nabla u} + \kappa {\Delta u} + f(\boldsymbol{x}, t),\\
    &u(\boldsymbol{x},0)=u_0(\boldsymbol{x}),
\end{aligned}
\end{equation}
where $\boldsymbol{x} \in \Omega \subset \mathbb{R}^d$ and $t$ respectively denote the spatial variable and the temporal variable, $\boldsymbol{\beta}[u](\boldsymbol{x},t) \in \mathbb{R}^d$ is a vector-valued mapping from $(u, \boldsymbol{x}, t)$ to $\mathbb{R}^d$, $\kappa \in \mathbb{R}^+$ is the diffusion parameter, and $f(\boldsymbol{x}, t) \in \mathbb{R}$ denotes the force term. {$\nabla u$ and $\Delta u$ demote the gradient and Laplacian of $u$, respectively. $u_0(\boldsymbol{x})$ represents the initial condition.} Many well-known PDEs, such as the Allen–Cahn and Navier-Stokes equations, can be viewed as a special form of Eq.~\ref{eq:main_pde}.

For such PDEs with the form of Eq.~\ref{eq:main_pde}, the Feynman-Kac formula provides the relationship between the PDEs and the corresponding probabilistic representation~\cite{pardoux1992backward, pardoux1999forward, han2018solving}. In detail, we can use the time inversion (i.e., $\tilde{u}(\boldsymbol{x},t)=u(\boldsymbol{x},T-t), \tilde{f}(\boldsymbol{x},t)=f(\boldsymbol{x},T-t)$) to the PDE as:
\begin{equation}\label{eq:backward_pde}
\begin{aligned}
    &\frac{\partial \tilde{u}}{\partial t} = -\boldsymbol{\beta}[\tilde{u}](\boldsymbol{x}, t)\cdot  {\nabla \tilde{u}} - \kappa  \Delta \tilde{u} - \tilde{f}(\boldsymbol{x}, t), \\
    &\tilde{u}(\boldsymbol{x},T)=u_0(\boldsymbol{x}).
\end{aligned}
\end{equation}
Applying the Feynman-Kac formula~\cite{mao2007stochastic} to the terminal value problem Eq.~\ref{eq:backward_pde}, we have
\begin{equation}\label{eq:proba_rep}
    \tilde{u}_{{0}}(\boldsymbol{x}) = \mathbb{E}\left[\tilde{u}_{T}(\tilde{\boldsymbol{\xi}}_{T}) + \int_{0}^{{T}} \tilde{f}(\tilde{\boldsymbol{\xi}}_s, s)ds\right],
\end{equation}
where $\tilde{\boldsymbol{\xi}}_s\in\mathbb{R}^d$ is a random process starting at $\boldsymbol{x}$, and moving from $t=0$ to $t=T$, which satisfies:
\begin{equation}\label{eq:bsde}
\begin{aligned}
& d\tilde{\boldsymbol{\xi}}_{s} = \boldsymbol{\beta}[\tilde{u}](\tilde{\boldsymbol{\xi}}_s, s)ds + \sqrt{ 2\kappa} d\boldsymbol{B}_s,\\
         &\tilde{\boldsymbol{\xi}}_{{0}}= \boldsymbol{x},
\end{aligned}
\end{equation}where $\boldsymbol{B}_s$ is the $d$-dimensional standard Brownian motion. Applying time inversion $t\rightarrow T-t$ to Eq.~\ref{eq:proba_rep} and letting $\boldsymbol{\xi}$ be the inversion of $\tilde{\boldsymbol{\xi}}$, we have
\begin{align}\label{eq:proba_rep_u}
    &{u}_{{T}}(\boldsymbol{x}) = \mathbb{E}_{\boldsymbol{\xi}}\left[{u}_{0}(\boldsymbol{\xi}_{0}) + \int_{0}^{{T}} {f}(\boldsymbol{\xi}_s, s)ds\right].
\end{align}

We illustrate the diagram of Feynman-Kac law in the 1D case in Fig.~\ref{fk}.A, where the mathematical expectation in Eq.~\ref{eq:proba_rep_u} can be naturally approximated via the Monte Carlo sampling. Feynman-Kac formula can automatically encode boundary conditions into the random walk of particles, including the periodical, Dirichlet and Neumann boundary conditions, as discussed in Fig.~\ref{fk}.B. Apart from Eq.~\ref{eq:main_pde}, some other PDEs can also be handled via the Feynman-Kac formula after specific processing, like wave equations~\cite{dalang2008feynman} and spatially varying diffusion equations~\cite{sawhney2022grid}. 

Based on the Feynman-Kac law, various Monte Carlo methods have been developed to solve PDEs with the form of Eq.~\ref{eq:main_pde}~\cite{chorin2009stochastic, pham2015feynman}. For linear PDEs, Monte Carlo methods can directly simulate the corresponding SDEs and obtain the solution of the targeted PDEs with Eq.~\ref{eq:proba_rep_u}. For nonlinear PDEs (like Navier-Stokes and Allen-Cahn equations), we cannot simulate the SDEs in Eq.~\ref{eq:bsde} or the mathematical expectation in Eq.~\ref{eq:proba_rep_u} directly because the unknown $u$ is required during the simulation. To this end, some advanced numerical algorithms have been developed, such as the random vortex method~\cite{qian2022random} and the branching diffusion method~\cite{henry2014numerical}. Compared to the traditional Eulerian methods, Monte Carlo methods are more adaptable to significant spatiotemporal variations due to their Lagrangian nature, which is widely used for simulating the breaking wave and turbulent problems~\cite{irisov2011numerical, kramer2001review}. Moreover, Monte Carlo methods have been proven to be less restrictive in terms of step size constraints via the analysis of the Courant number~\cite{cottet2000vortex, rossi2015numerical}. However, the accuracy of Monte Carlo methods is limited by the number of particles, which can introduce severe computational time and memory issues~\cite{mimeau2021review}. In the following, we will introduce the MCNP Solver, which can be efficiently trained while inheriting the advantages of the Monte Carlo simulation method.

\subsection{Neural Monte Carlo Loss}\label{sec:mcloss}
Given a PDE in the form of Eq.~\ref{eq:main_pde} and the distribution of initial conditions $\mathcal{D}_0$, the objective of MCNP Solver is to learn a functional mapping $\mathcal{G}_{\theta}$ with parameter $\theta$ that can simulate the subsequent fields for all initial fields $u_0\sim \mathcal{D}_0$ at time $t\in[0, T]$. The inputs and outputs of $\mathcal{G}_{\theta}$ are given as:
\begin{equation}
    \begin{aligned}
        \mathcal{G}_{\theta}: \mathcal{D}_0 \times [0, T] &\to  \mathcal{D}_{[0, T]},\\
        \left(u_0, t\right) &\mapsto u_t,
    \end{aligned}
\end{equation}
where $\mathcal{D}_{[0, T]}$ denotes the joint distribution of the field after $t=0$. In this paper, we are interested in the evolution of PDEs at fixed coordinate system $\{\boldsymbol{x}_p\}_{p=1}^P \in \Omega$, which aligns with the settings in~\cite{DBLP:conf/iclr/LiKALBSA21, brandstetter2022message}. Consequently, the inputs and outputs of the solver $\mathcal{G}_{\theta}$ are solution values at $P$ grid points, i.e., each $u_t$ is represented by a $P$-dimensional vector $[u_t(\boldsymbol{x}_1), \cdots, u_t(\boldsymbol{x}_P)]$. 

In practice, we use neural networks~\cite{DBLP:conf/iclr/LiKALBSA21} to construct the mapping $\mathcal{G}_{\theta}$ and the trained model served as the MCNP Solver. Unlike other supervised learning algorithms~\cite{DBLP:conf/iclr/LiKALBSA21, lu2021learning, brandstetter2022message}, MCNP Solver is trained in an unsupervised way, i.e., only utilizing the physics information provided by PDEs. To this end, we consider training the solver via the relationship between $u_{t}$ and $u_{t+\Delta t}$ (where $0\leq t<t+\Delta t \leq T$) derived by the aforementioned probabilistic representation. Considering Eq.~\ref{eq:proba_rep_u}, the optimal parameter $\theta^*$ for the neural PDE solver $\mathcal{G}_{\theta}$ should satisfy the following equation at each $\boldsymbol{x}_p$ after optimization:
\begin{equation}\label{eq:sde_g}
\begin{aligned}
        &\mathcal{G}_{\theta^*}(u_{0}, t+\Delta t)(\boldsymbol{x}_p)  \\= &\mathbb{E}_{\boldsymbol{\xi}_p}\left[\mathcal{G}_{\theta^*}(u_{0}, t)(\boldsymbol{\xi}_{p,t}) +\int_{t}^{t+\Delta t}f(\boldsymbol{\xi}_{p,s},s) ds\right],
\end{aligned}
\end{equation}
where $\boldsymbol{\xi}_{p,s} (s\in[t,t+\Delta t]$) is the inverse version of stochastic process in Eq.~\ref{eq:bsde} as follows:
\begin{equation}\label{eq:sde_MCNP}
\begin{aligned}
 &  d\boldsymbol{\xi}_{p,s} =  - \boldsymbol{\beta}[{u}](\boldsymbol{\xi}_{p,s}, s)ds - \sqrt{ 2\kappa} d\boldsymbol{B}_s,
 \\ &\boldsymbol{\xi}_{p,{t+\Delta t}}= \boldsymbol{x}_p. 
\end{aligned}
\end{equation}
{To search for $\theta^*$, we convert Eq.~~\ref{eq:sde_g} to the optimization objective, and thus, the neural Monte Carlo loss can be written as follows:}
\begin{equation}\label{eq:loss_nl}
\begin{aligned}
       \mathcal{L}_{\operatorname{MCNP}}(\mathcal{G}_{\theta}|u_0)=  \sum_{t=0}^{T-\Delta t}\sum_{p=1}^P
           \Bigg\| \mathcal{G}_{\theta}(u_{0}, t+\Delta t)(\boldsymbol{x}_p) -\\
        \mathbb{E}_{\boldsymbol{\xi}_p}\left[ \mathcal{G}_{\theta}( u_0, t)(\boldsymbol{\xi}_{p,t})+  \int_{t}^{t+\Delta t}f(\boldsymbol{\xi}_{p,s}, s) ds\right]\Bigg\|_2,
\end{aligned}
\end{equation} 
where $\mathcal{G}_{\theta}(u_0, 0)$ is directly set to be $u_0$ in experiments. During the training stage, we uniformly sample $B$ initial states $u_0$ from $\mathcal{D}_0$ per epoch to ensure the generalizability.

\begin{figure*}[!t]
\begin{center}
\centerline{\includegraphics[width=17cm]{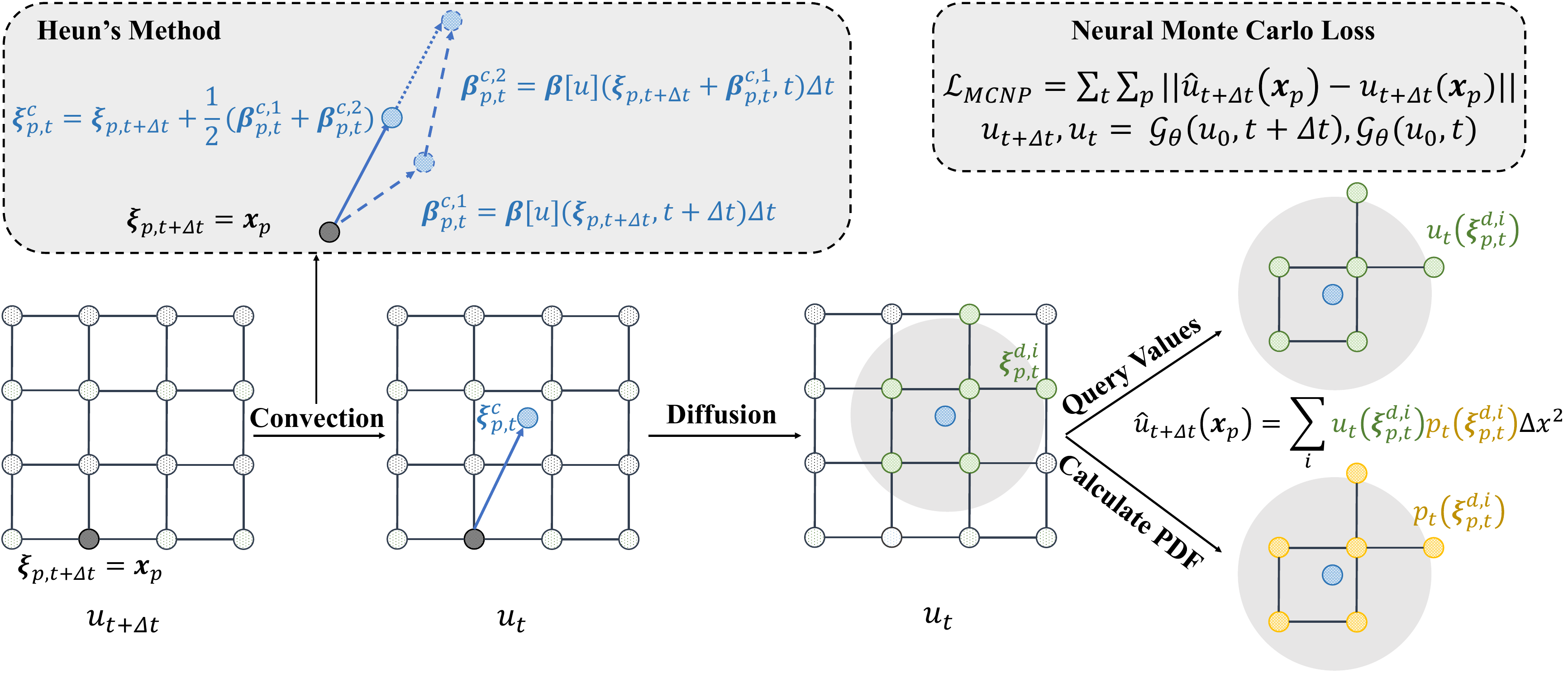}}
\vspace{-5pt}
\caption{{\textbf{Illustration of neural Monte Carlo loss. }We construct the training loss via the relationship between ${u}_{t}$ and $u_{t+\Delta t}$ given by the Feynman-Kac law. Given a grid point at $\boldsymbol{x}_p$, we split the random walk in Eq.~\ref{eq:bsde_time_inver} as the convection and diffusion parts. For the convection process, we utilize the Heun's method to simulate the particle moving from time $t+\Delta t$ to $t$ driven by the drift term $\boldsymbol{\beta}$. For the diffusion process, we calculate the mathematical expectation in Eq.~\ref{eq:sde_g} through the probability density function (PDF) of neighbouring grid points to eliminate sampling using Monte Carlo methods, where $p_{t}(\boldsymbol{\xi}_{p,t}^{d, i})$ denotes the transition probability for the particle moving from $\boldsymbol{\xi}_{p,t}^{c}$ to $\boldsymbol{\xi}_{p,t}^{d, i}$ driven by the diffusion effect. Please note that we omit external forcing $f$ in the figure for simplification.}}
\label{frame}
\end{center}
\vspace{-10pt}
\end{figure*}

\subsection{The Simulation of SDE}\label{sec:sde}
When calculating the loss function defined in Eq.~\ref{eq:loss_nl}, a crucial step is to approximate the expectation over random samples $\boldsymbol{\xi}_p$ via simulating the SDE in Eq.~\ref{eq:sde_MCNP}. {The Classical Euler–Maruyama method~\cite{kloeden2011numerical} has been adopted to approximate corresponding SDEs~\cite{han2018solving,nusken2021interpolating}, i.e., }

\begin{equation}\label{eq:bsde_time_inver}
\begin{aligned}
    {\boldsymbol{\xi}}_{p,t}^m &= {\boldsymbol{\xi}}_{p,t+\Delta t} + \underbrace{\boldsymbol{\beta}[u](\boldsymbol{\xi}_{p,t+\Delta t}, t+\Delta t)\Delta t}_{\text{convection}} + \underbrace{\sqrt{2\kappa\Delta t} \boldsymbol{b}^m}_{\text{diffusion}}, \\ \boldsymbol{b}^m & \sim\mathcal{N}(\boldsymbol{0}, \boldsymbol{I}),\ {\boldsymbol{\xi}}_{p,t+\Delta t} = \boldsymbol{x}_p,
\end{aligned}
\end{equation}
where the physical meaning of $\boldsymbol{\beta}[u](\boldsymbol{\xi}_{p,t+\Delta t}, t+\Delta t)\Delta t$ and $\sqrt{2\kappa\Delta t} \boldsymbol{b}^m$ denote the convection and diffusion processes, respectively. After sampling $M$ particles $\{\boldsymbol{\xi}^m_{p,t}\}_{m=1}^M$, the training target in Eq.~\ref{eq:sde_g} can be approximated via the Monte Carlo method, i.e., 

\begin{equation}\label{eq:mc}
    \hat{u}_{t+\Delta t}(\boldsymbol{x}_p) \approx \frac{1}{M}\sum_{m=1}^M \left[u_t(\boldsymbol{\xi}_{p,t}^m) + f(\boldsymbol{\xi}_{p, t+\Delta t}, t+\Delta t)\Delta t\right].
\end{equation}

The selection of the simulation algorithm significantly impacts the efficiency of the training process and the accuracy of the trained solver. This necessitates a careful design approach that considers both the approximation error and the training efficiency, such as computational time and memory. On the one hand, to reduce the discretization error when dealing with the convection term, we consider a higher-order approximation scheme. On the other hand, sampling {a large number of} particles to calculate the diffusion term will introduce severe computational issues, especially when the diffusive rate $\kappa$ is high. As a result, we explore a method that avoids sampling of the particles during the diffusion process. 

In the subsequent sections, we will introduce two methods, including:
1). Utilize Heun's method to obtain a more accurate approximation of the convection process; 2). Calculate the mathematical expectation via the probability density function (PDF) of neighbouring grid points when handling the diffusion process to eliminate the sampling in Monte Carlo methods, as shown in Fig.~\ref{frame}.

\subsubsection{The Simulation of Convection Process} \label{sec:heun} 

We aim to approximate the location of $\boldsymbol{\xi}_{p,s}$ moving back to time $t$ with the convection effect, i.e., $\boldsymbol{\xi}_{p, t}^{c} \triangleq \boldsymbol{\xi}_{p, t+\Delta t} + \int_{t}^{t+\Delta t} \boldsymbol{\beta}[u](\boldsymbol{\xi}_{p,s}, s) ds$. To obtain a more accurate result, we replace the classical Euler scheme with Heun's method, which calculates the intermediate location before the final approximation. The complete mathematical expression can be written as follows:

\begin{equation}\label{heun}
    \begin{aligned}
        \boldsymbol{\beta}_{p, t}^{c, 1}&=\boldsymbol{\beta}[u]\left(\boldsymbol{\xi}_{p, t+\Delta t}, t+\Delta t\right) \Delta t;\\
        \boldsymbol{\beta}_{p, t}^{c, 2}&=\boldsymbol{\beta}[u]\left(\boldsymbol{\xi}_{p, t+\Delta t}+\boldsymbol{\beta}_{p, t}^{c, 1}, t\right) \Delta t;\\
        \boldsymbol{\xi}_{p, t}^c&=\boldsymbol{\xi}_{p, t+\Delta t}+\frac{1}{2}\left(\boldsymbol{\beta}_{p, t}^{c, 1}+\boldsymbol{\beta}_{p, t}^{c, 2}\right).
    \end{aligned}
\end{equation}
{For the cases in which the drift term $\boldsymbol{\beta}$ depends on solution $u$, we utilize the output of MCNP Solver to approximate $\boldsymbol{\beta}[u]$ accordingly.}

\subsubsection{The Simulation of Diffusion Process}\label{sec:diffusion}
\begin{figure*}[!t]
\begin{center}
\centerline{\includegraphics[width=15cm]{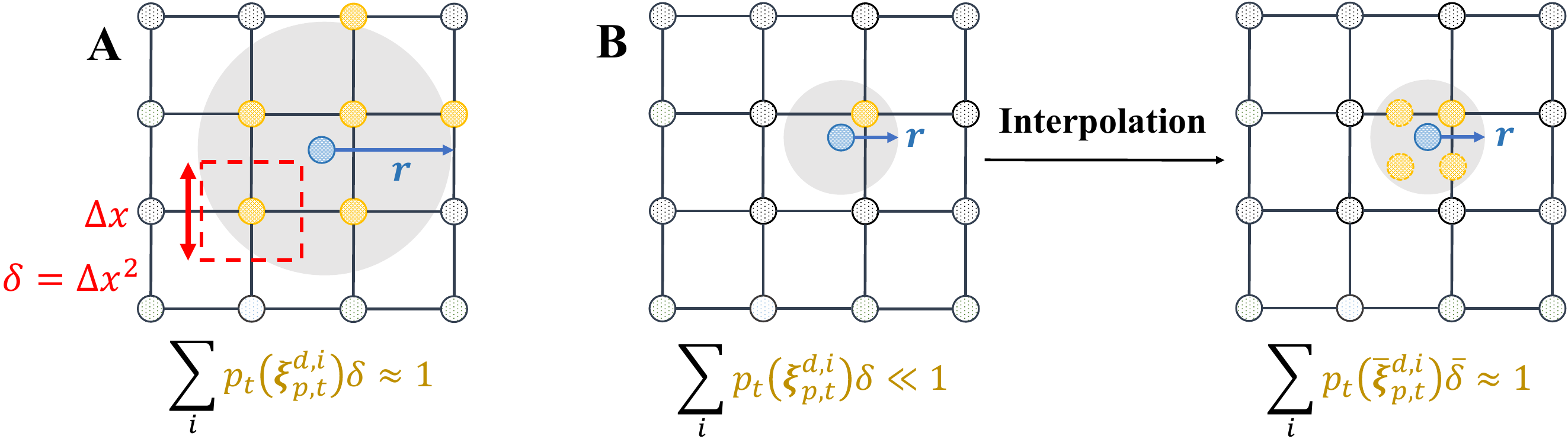}}
\vspace{-5pt}
\caption{{\textbf{The Choice of Neighbourhood Radius $r$ }\textbf{A: }We first choose the smallest possible value of $r$ that satisfies Eq.~\ref{eq:r}. If we have $\sum_i p_t(\boldsymbol{\xi}_{p,t}^{d,i}) \delta \approx 1$, using Eq.~\ref{eq:expectation_nfk} directly to approximate the corresponding mathematical expectation can reduce computational cost while ensuring accuracy. $\delta$ denotes the volume of each cell (as shown in the red box) in the coordinate system. \textbf{B: }When the grid size is close to the radius $r$, we may encounter the scenario that $\sum_i p_t(\boldsymbol{\xi}_{p,t}^{d,i}) \delta \ll 1$. To address this issue, we interpolate the coordinate system to a high-resolution one to satisfy the normalization condition in Eq.~\ref{eq:r}. $\Bar{\delta}$ denotes the volume of each cell in the high-resolution coordinate system.}}
\label{fig:r}
\end{center}
\vspace{-10pt}
\end{figure*}

Unlike the Euler-Maruyama method, which samples $M$ particles to simulate the diffusion process, we calculate the mathematical expectation via the PDF of neighbouring grid points to replace the sampling in Monte Carlo methods. Given $\boldsymbol{\xi}_{p,t}^c$, which represents the location of particles after the convection process (Eq.~\ref{heun}), we first find its neighbourhood $N(\boldsymbol{\xi}_{p,t}^c, r)$ from the coordinates as follows:
\begin{equation}
    N(\boldsymbol{\xi}_{p,t}^c, r) \triangleq \{\boldsymbol{\xi}_{p,t}^{d,i}:\|\boldsymbol{\xi}_{p,t}^{d,i}-\boldsymbol{\xi}_{p,t}^c\|_2\leq r,\boldsymbol{\xi}_{p,t}^{d,i} \in \{\boldsymbol{x}_p\}_{p=1}^P\},
\end{equation}
where $r>0$ represents the radius of the neighbourhood. Considering the transition probability for $\boldsymbol{\xi}_{p,t}^c$ moving to its neighbour $\boldsymbol{\xi}_{p,t}^{d,i}\in N(\boldsymbol{\xi}_{p,t}^c, r)$, we can calculate the analytical form of the corresponding PDF as follows:
\begin{equation}\label{eq:pdf}
    p_t(\boldsymbol{\xi}_{p,t}^{d,i}) = \frac{1}{({2\pi\sigma^2})^{d/2}} \exp\left({-\frac{\|\boldsymbol{\xi}_{p,t}^{d,i} - \boldsymbol{\xi}_{p,t}^{c}\|_2^2}{2\sigma^d}}\right),
\end{equation}
where $\sigma=\sqrt{2\kappa\Delta t}$ denotes the standard deviation of the Brownian motion in Eq.~\ref{eq:sde_MCNP}. Please note in Eq.~\ref{eq:pdf}, we assume that the particles will not hit the boundary by default. When the particle $\boldsymbol{\xi}_{p,t}^c$ has the possibility of hitting the boundary, we need to modify the PDF in Eq.~\ref{eq:pdf} to the corresponding form. For instance, the PDF corresponds to the Brownian motion with absorption/reflection when we handle the Dirichlet/Neumann boundary conditions, respectively. We use the grid points inside the neighbourhood to approximate the mathematical expectation in Eq.~\ref{eq:sde_g} as follows:
\begin{equation}\label{eq:expectation_nfk}
\begin{aligned}
        &\mathbb{E}_{\boldsymbol{\xi}_p}\left[\mathcal{G}_{\theta}(u_{0}, t)(\boldsymbol{\xi}_{p,t}) \right] \\
        \approx & \sum_{\boldsymbol{\xi}_{p,t}^{d,i}\in N(\boldsymbol{\xi}_{p,t}^c, r)} \mathcal{G}_{\theta}(u_{0}, t)(\boldsymbol{\xi}_{p,t}^{d,i})\cdot p_t(\boldsymbol{\xi}_{p,t}^{d,i})\cdot \delta,
\end{aligned}
\end{equation}
where we use the density value at point $\boldsymbol{\xi}_{p,t}^{d,i}$ to approximate the density of the 
cells centered on $\boldsymbol{\xi}_{p,t}^{d,i}$ (the red box in Fig.~\ref{fig:r}.A) and $\delta$ denotes the volume of each cell. We utilize a similar way to calculate the integration of the external forcing $f$ in Eq.~\ref{eq:sde_g} as follows:
\begin{equation}
\begin{aligned}
   & \mathbb{E}_{\boldsymbol{\xi}_p}\left[ \int_{t}^{t+\Delta t}f(\boldsymbol{\xi}_{p,s},s) ds\right] \\
   \approx &\frac{\Delta t}{2}\left[f(\boldsymbol{\xi}_{p, t+\Delta t}, t+\Delta t)+\sum_{\boldsymbol{\xi}_{p,t}^{d,i}\in N(\boldsymbol{\xi}_{p,t}^c, r)} f(\boldsymbol{\xi}_{p,t}^{d,i}, t)p_t(\boldsymbol{\xi}_{p,t}^{d,i}) \right].
\end{aligned}
\end{equation}

\subsubsection{The Choice of Neighbourhood Radius $r$} \label{sec:r}
In Eq.~\ref{eq:expectation_nfk}, we utilize the grid points located in the neighbourhood $N(\boldsymbol{\xi}_{p,t}^c, r)$ to approximate the mathematical expectation. To ensure the approximation error is under control, we hope the accumulated probability of the points in the range will approach 1, i.e., given a sufficiently small $\epsilon$, we hope the radius $r$ satisfies the following conditions:
\begin{equation}\label{eq:r}
\begin{aligned}
        &r=\arg \min {\gamma},\\
        \text{s.t.}, &\int_{\|\boldsymbol{\xi}-\boldsymbol{\xi}^c_{p,t}\|_2\leq \gamma} p_t(\boldsymbol{\xi})d\boldsymbol{\xi} \geq 1-\epsilon.
\end{aligned}
\end{equation}

The radius $r$ in Eq.~\ref{eq:r} represents the minimum scope {within which} particles can be covered with a probability close to $1$ during the diffusion process. After choosing the radius $r$, we use the neighbouring grid points to calculate the corresponding mathematical expectation via Eq.~\ref{eq:expectation_nfk}, which can be accurately approximated with minimal computational cost when $\sum_i p_t(\boldsymbol{\xi}_{p,t}^{d,i}) \delta \approx 1$. However, when the grid size is close to the radius $r$, we may encounter $\sum_i p_t(\boldsymbol{\xi}_{p,t}^{d,i}) \delta \ll 1$ because the random walks of particles are concentrated around $\boldsymbol{\xi}_{p,t}^c$ while the grid points are not dense enough. To address this issue, we interpolate the coordinate system $\{\boldsymbol{x}_p\}_{p=1}^P$ to a high-resolution one $\{\boldsymbol{\Bar{x}}_p\}_{p=1}^{\Bar{P}}$ {(yellow points in Fig.~\ref{fig:r}.B)}, such that $\sum_i p_t(\boldsymbol{\Bar{\xi}}_{p,t}^{d,i})\Bar{\delta} \approx 1$, where $\Bar{\delta}$ is the volume of each cell in the high-resolution coordinate system and $\boldsymbol{\Bar{\xi}}_{p,t}^{d,i}$ located in the neighbourhood $\Bar{N}(\boldsymbol{\xi}_{p,t}^c, r)$ defined as:
\begin{equation}
    \Bar{N}(\boldsymbol{\xi}_{p,t}^c, r) \triangleq \{\boldsymbol{\Bar{\xi}}_{p,t}^{d,i}:\|\boldsymbol{\Bar{\xi}}_{p,t}^{d,i}-\boldsymbol{{\xi}}_{p,t}^c\|_2\leq r,\boldsymbol{\Bar{\xi}}_{p,t}^{d,i} \in \{\boldsymbol{\Bar{x}}_p\}_{p=1}^{\Bar{P}}\}.
\end{equation}

After that, we approximate the expectation in Eq.~\ref{eq:sde_g} in the high-resolution coordinate system as in Eq.~\ref{eq:expectation_nfk}. Please note that we only have access to the value of $\mathcal{G}_{\theta}(u_{0}, t)$ in the low-resolution coordinate system directly, and thus we need to interpolate the physical field $\mathcal{G}_{\theta}(u_{0}, t)$ to the corresponding resolution. In practice, we utilize the Fourier transform to map the low-resolution PDE fields to the frequency domain, and use the inverse Fourier transform to remap it to the high-resolution space. We illustrate the choice of $r$ and the interpolation trick of the MCNP Solver in Fig.~\ref{fig:r}.

\section{Experiments}\label{sec:exp}
In this section, we conduct numerical experiments to evaluate the proposed MCNP Solver on four tasks: 1D convection-diffusion equation, 1D Allen-Cahn equation, 2D Navier-Stokes equation, and 2D fractional equation on a disk. We introduce the implementation details of each baseline method and the generalization scheme of the test data in Appendix~\uppercase\expandafter{\romannumeral1}. We utilize the FNO~\cite{DBLP:conf/iclr/LiKALBSA21} as the backbone network for MCNP Solver. For each task, we divide the time interval into $10$ uniform frames, and evaluate the model performance via the average relative $\ell_2$ error ($E_{\ell_2}$) and relative $\ell_{\infty}$ error ($E_{\ell_{\infty}}$) over $10$ frames on 200 test PDE samples. We repeat each experiment with three random seeds in $\{0,1,2\}$ and report the mean value and variance. {All experiments are implemented on an NVIDIA RTX 3090 GPU. The source code is publicly available at: \url{https://github.com/optray/MCNP}.}

\subsection{1D Convection-Diffusion Equation}\label{sec:cde}

\begin{figure*}[!t]
\begin{center}
\centerline{\includegraphics[width=18cm]{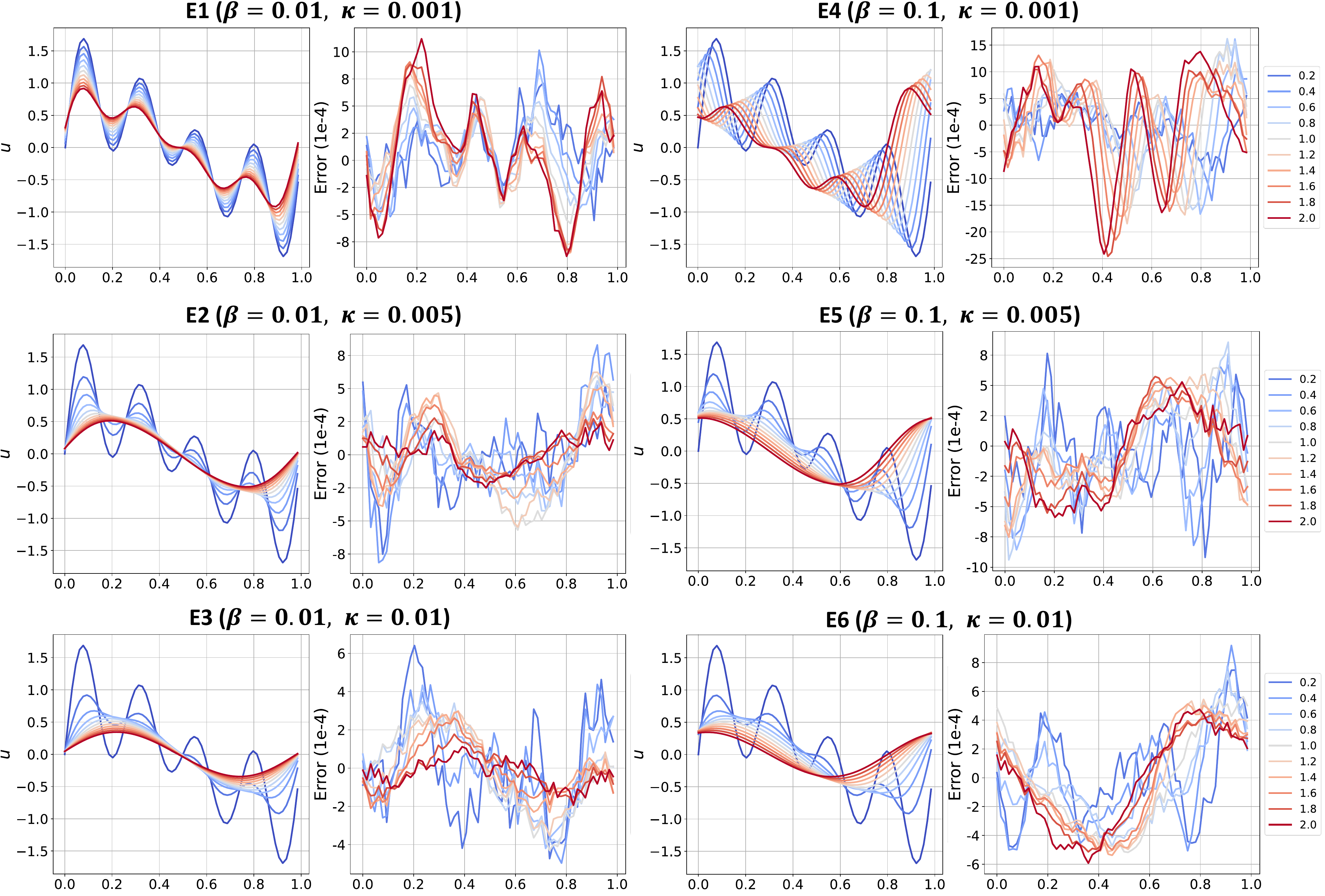}}
\vspace{-5pt}
\caption{{{\textbf{Simulation of 1D convection-diffusion equation.} The prediction result (Left) and point-wise error (Right) of MCNP-10 for an example in E1-E6. The x-axis and y-axis represent spatial coordinates and the predicted values (point-wise error).}}}
\label{cde_fig}
\end{center}
\vspace{-10pt}
\end{figure*}

\begin{table}[!t]
\centering
\caption{{{\textbf{1D convection-diffusion equation with varying $\beta$ and $\kappa$.} Relative $\ell_2$ errors ($E_{\ell_2}$), relative $\ell_{\infty}$ errors ($E_{\ell_{\infty}}$) and computational costs for baseline methods and MCNP Solver. The best results are marked in bold, and the second best results are underlined.}}}\label{tab:diffusion}
\vspace{-5pt}
\resizebox{8.8cm}{!}{\begin{tabular}{cccccc}
\toprule
Task                                     & Model & $E_{\ell_2}$ (\%) & $E_{\ell_{\infty}}$ (\%)& \shortstack{Train \\Time (H)} & \shortstack{Param. \\(M)}\\ \midrule
\multicolumn{1}{l|}{\multirow{6}{*}{{\shortstack{E1 \\$\beta=0.01$\\ $\kappa=0.001$}}}} &  {PINO-10}      &  {0.155 \footnotesize{$\pm$0.005}}    &   {0.157 \footnotesize{$\pm$0.010}}   &    {0.034}  &   {0.140}    \\
\multicolumn{1}{l|}{}                    &  {PINO-20}     & {\textbf{0.075 \footnotesize{$\pm$0.001}}}     &    {\textbf{0.083 \footnotesize{$\pm$0.001}}}  &  {0.054}    &   {0.140}     \\
\multicolumn{1}{l|}{}                    &  {PINO-100}     & {0.118 \footnotesize{$\pm$0.005}}     &    {0.129 \footnotesize{$\pm$0.003}}    & {0.257}     &   {0.140}     \\
\multicolumn{1}{l|}{}                    &    {PI-DeepONet}   &   {0.495 \footnotesize{$\pm$0.196}}     &    {0.705 \footnotesize{$\pm$0.312}}   &   {0.086}   &   {0.153}     \\
\multicolumn{1}{l|}{}                    &    {PI-DeepONet-M}   &   {0.367 \footnotesize{$\pm$0.087}}     &    {0.594 \footnotesize{$\pm$0.092}}   &   {0.280}   &   {0.153}     \\
\multicolumn{1}{l|}{}                    &   {MCNP-10}    &  {\underline{0.090 \footnotesize{$\pm$0.002}}}    &   {\underline{0.103 \footnotesize{$\pm$0.002}}} &  {0.032}    &    {0.140}    \\
\multicolumn{1}{l|}{}                    &  {MCNP-20}     &  {{0.119 \footnotesize{$\pm$0.010}}}    &   {{0.134 \footnotesize{$\pm$0.002}}}  &   {0.054}   &   {0.140}     \\ \midrule
\multicolumn{1}{l|}{\multirow{6}{*}{{\shortstack{E2 \\$\beta=0.01$\\ $\kappa=0.005$}}}} &  {PINO-10}      &  {1.717 \footnotesize{$\pm$0.042}}    &   {1.886 \footnotesize{$\pm$0.058}}   &    {0.034}  &   {0.140}    \\
\multicolumn{1}{l|}{}                    &  {PINO-20}     & {{0.484 \footnotesize{$\pm$0.003}}}     &    {{0.536 \footnotesize{$\pm$0.002}}}  &  {0.054}    &   {0.140}     \\
\multicolumn{1}{l|}{}                    &  {PINO-100}     & {0.179 \footnotesize{$\pm$0.027}}     &    {0.209 \footnotesize{$\pm$0.034}}    & {0.257}     &   {0.140}     \\
\multicolumn{1}{l|}{}                    &    {PI-DeepONet}   &   {0.686 \footnotesize{$\pm$0.143}}     &    {0.929 \footnotesize{$\pm$0.195}}   &   {0.086}   &   {0.153}     \\
\multicolumn{1}{l|}{}                    &    {PI-DeepONet-M}   &   {0.483 \footnotesize{$\pm$0.122}}     &    {0.637 \footnotesize{$\pm$0.177}}   &   {0.280}   &   {0.153}     \\
\multicolumn{1}{l|}{}                    &   {MCNP-10}    &  {\textbf{0.128 \footnotesize{$\pm$0.014}}}    &   {\textbf{0.160 \footnotesize{$\pm$0.012}}} &  {0.032}    &    {0.140}    \\
\multicolumn{1}{l|}{}                    &  {MCNP-20}     &  {\underline{0.150 \footnotesize{$\pm$0.012}}}    &   {\underline{0.180 \footnotesize{$\pm$0.014}}}  &   {0.054}   &   {0.140}     \\ \midrule
\multicolumn{1}{l|}{\multirow{6}{*}{{\shortstack{E3 \\$\beta=0.01$\\ $\kappa=0.01$}}}} &  {PINO-10}      &  {3.387 \footnotesize{$\pm$0.031}}    &   {3.937 \footnotesize{$\pm$0.046}}   &    {0.034}  &   {0.140}    \\
\multicolumn{1}{l|}{}                    &  {PINO-20}     & {{1.023 \footnotesize{$\pm$0.002}}}     &    {{1.176 \footnotesize{$\pm$0.013}}}  &  {0.054}    &   {0.140}     \\
\multicolumn{1}{l|}{}                    &  {PINO-100}     & {0.235 \footnotesize{$\pm$0.036}}     &    {0.273 \footnotesize{$\pm$0.036}}    & {0.257}     &   {0.140}     \\
\multicolumn{1}{l|}{}                    &    {PI-DeepONet}   &   {1.268 \footnotesize{$\pm$0.140}}     &    {1.756 \footnotesize{$\pm$0.192}}   &   {0.086}   &   {0.153}     \\
\multicolumn{1}{l|}{}                    &    {PI-DeepONet-M}   &   {0.659 \footnotesize{$\pm$0.069}}     &    {0.885 \footnotesize{$\pm$0.088}}   &   {0.280}   &   {0.153}     \\
\multicolumn{1}{l|}{}                    &   {MCNP-10}    &  {\textbf{0.170 \footnotesize{$\pm$0.012}}}    &   {\textbf{0.210 \footnotesize{$\pm$0.009}}} &  {0.032}    &    {0.140}    \\
\multicolumn{1}{l|}{}                    &  {MCNP-20}     &  {\underline{0.228 \footnotesize{$\pm$0.029}}}    &   {\underline{0.272 \footnotesize{$\pm$0.035}}}  &   {0.054}   &   {0.140}     \\ \midrule
\multicolumn{1}{l|}{\multirow{6}{*}{{\shortstack{E4 \\$\beta=0.1$\\ $\kappa=0.001$}}}} &  {PINO-10}      &  {6.287 \footnotesize{$\pm$0.155}}    &   {6.783 \footnotesize{$\pm$0.354}}   &    {0.034}  &   {0.140}    \\
\multicolumn{1}{l|}{}                    &  {PINO-20}     & {{1.567 \footnotesize{$\pm$0.120}}}     &    {{1.664 \footnotesize{$\pm$0.121}}}  &  {0.054}    &   {0.140}     \\
\multicolumn{1}{l|}{}                    &  {PINO-100}     & {0.257 \footnotesize{$\pm$0.017}}     &    {0.295 \footnotesize{$\pm$0.031}}    & {0.257}     &   {0.140}     \\
\multicolumn{1}{l|}{}                    &    {PI-DeepONet}   &   {0.615 \footnotesize{$\pm$0.048}}     &    {0.889 \footnotesize{$\pm$0.078}}   &   {0.086}   &   {0.153}     \\
\multicolumn{1}{l|}{}                    &    {PI-DeepONet-M}   &   {0.300 \footnotesize{$\pm$0.042}}     &    {0.376 \footnotesize{$\pm$0.046}}   &   {0.280}   &   {0.153}     \\
\multicolumn{1}{l|}{}                    &   {MCNP-10}    &  {\textbf{0.187 \footnotesize{$\pm$0.007}}}    &   {\textbf{0.228 \footnotesize{$\pm$0.012}}} &  {0.032}    &    {0.140}    \\
\multicolumn{1}{l|}{}                    &  {MCNP-20}     &  {\underline{0.201 \footnotesize{$\pm$0.017}}}    &   {\underline{0.235 \footnotesize{$\pm$0.023}}}  &   {0.054}   &   {0.140}     \\ \midrule
\multicolumn{1}{l|}{\multirow{6}{*}{\shortstack{E5 \\$\beta=0.1$\\ $\kappa=0.005$}}} &  PINO-10     &  3.919 \footnotesize{$\pm$0.025}    &   4.526 \footnotesize{$\pm$0.155}   &    0.034  &   0.140     \\
\multicolumn{1}{l|}{}                    &  PINO-20     & 1.106 \footnotesize{$\pm$0.010}     &    1.318 \footnotesize{$\pm$0.023}  &  0.054    &   0.140     \\
\multicolumn{1}{l|}{}                    &  PINO-100     & 0.222 \footnotesize{$\pm$0.015}     &  0.268 \footnotesize{$\pm$0.017}    & 0.257     &   0.140     \\
\multicolumn{1}{l|}{}                    &    PI-DeepONet   &   0.706 \footnotesize{$\pm$0.113} &   0.968 \footnotesize{$\pm$0.114}   &   0.086   &   0.153     \\
\multicolumn{1}{l|}{}                    &    {PI-DeepONet-M}   &   {0.588 \footnotesize{$\pm$0.038}}     &    {0.718 \footnotesize{$\pm$0.043}}   &   {0.280}   &   {0.153}     \\
\multicolumn{1}{l|}{}                    &   MCNP-10    &  \textbf{0.172 \footnotesize{$\pm$0.018}}    &   \textbf{ 0.222 \footnotesize{$\pm$0.018} } &  0.032    &    0.140    \\
\multicolumn{1}{l|}{}                    &  MCNP-20     &  \underline{0.185 \footnotesize{$\pm$0.010}}    &   \underline{0.235 \footnotesize{$\pm$0.013}}  &   0.054   &   0.140     \\ \midrule
\multicolumn{1}{l|}{\multirow{6}{*}{\shortstack{E6 \\$\beta=0.1$\\ $\kappa=0.01$}}} &  PINO-10     &  4.784 \footnotesize{$\pm$0.031}    &   5.868 \footnotesize{$\pm$0.072}   &  0.034    &   0.140     \\
\multicolumn{1}{l|}{}                    &  PINO-20     & 1.435 \footnotesize{$\pm$0.005}     &    1.720 \footnotesize{$\pm$0.008}  &   0.054   & 0.140       \\
\multicolumn{1}{l|}{}                    &  PINO-100     & 0.328 \footnotesize{$\pm$0.065}     &  0.385 \footnotesize{$\pm$0.063}    &   0.257   &    0.140    \\
\multicolumn{1}{l|}{}                    &    PI-DeepONet   &   1.440 \footnotesize{$\pm$0.410} &   1.921 \footnotesize{$\pm$0.424}   &   0.086   &   0.153  \\
\multicolumn{1}{l|}{}                    &    {PI-DeepONet-M}   &   {0.827 \footnotesize{$\pm$0.051}}     &    {1.129 \footnotesize{$\pm$0.042}}   &   {0.280}   &   {0.153}     \\
\multicolumn{1}{l|}{}                    &   MCNP-10    & \textbf{0.200 \footnotesize{$\pm$0.012}}    &    \textbf{0.259 \footnotesize{$\pm$0.012}}  &   0.032   &   0.140     \\
\multicolumn{1}{l|}{}                    &  MCNP-20     &  \underline{0.248 \footnotesize{$\pm$0.025}}    &   \underline{0.307 \footnotesize{$\pm$0.024}}   &   0.054   &   0.140     \\ \bottomrule
\end{tabular}}
\vspace{-10pt}
\end{table}

In this section, we conduct experiments on periodical 1D convection-diffusion equation defined as follows:
\begin{equation}\label{eq:heat}
        \frac{\partial u(x,t)}{\partial t} = \beta \nabla u(x, t)  + \kappa \Delta u(x,t),\ x \in [0, 1], t\in [0, 2].
\end{equation}
The initial states $u(x, 0)$ are generated from the functional space $\mathcal{F}_N \triangleq \{\sum_{n=1}^N a_n \sin(2\pi n x):a_n \sim \mathbb{U}(0, 1)\},$
where $\mathbb{U}(0, 1)$ denotes the uniform distribution over $(0, 1)$, and $N$ represents the maximum frequency of the functional space. 

\subsubsection{Experimental Settings} 
{In this setting, $\beta$ and $\kappa$ respectively represent the convection and diffusion rate, which dominate the inherent property of corresponding physical revolution for Eq.~\ref{eq:heat}. To systematically evaluate the performance of the methods employed, we select two different $\beta$ in $\{0.01, 0.1\}$ and three different $\kappa$ in $\{0.001, 0.005, 0.01\}$. These six experimental settings are denoted E1-E6 with $(\beta,\kappa)=(0.01, 0.001), (0.01, 0.005), (0.01, 0.01), (0.1, 0.001),\\ (0.1,0.005), (0.1, 0.01)$, respectively. In this section, we set the maximum frequency $N$ as 5. We divide the spatial domain $[0, 1)$ into $64$ grid elements for all experiments.}

\subsubsection{Baselines} We introduce the neural PDE solvers performed on 1D convection-diffusion equations, including: \romannumeral1). \textbf{PINO}~\cite{li2021physics}: an unsupervised neural operator. To evaluate the performance of PINO with varying step sizes, we divide the time interval into 10/20/100 uniform frames, denoted as PINO-10/20/100, respectively. The loss function is constructed through the pseudo-spectral method due to the periodic boundary condition; thus, the boundary conditions can be naturally satisfied. \romannumeral2). \textbf{PI-DeepONet}~\cite{wang2021learning}: an unsupervised neural operator based on PINN loss and DeepONet, which utilizes the residual of PDE to construct the loss function. {\romannumeral3). \textbf{PI-DeepONet-M}~\cite{wang2022improved}: a modified version of PI-DeepONet, which utilizes an adaptive re-weighting scheme to balance training samples and loss functions.} \romannumeral4). \textbf{MCNP Solver}, we divide the time interval into 10/20 uniform frames, which are denoted as MCNP-10/20, respectively. 

\subsubsection{Results} 
Fig.~\ref{cde_fig} illustrates the predicted $u$ for MCNP-10 from $t = 0.2$ to $t = 2.0$ for E1-E6, respectively. When comparing the cases with $\beta=0.01$ and $0.1$, the effect of material advection becomes more remarkable with larger $\beta$, with peaks and troughs of the heat flows more noticeably shifting towards the negative $x$-axis direction. Keeping $\beta$ as a constant, as $\kappa$ increases, we observe that the diffusion effect gradually governs the motion of the material, causing the peaks and troughs to decay rapidly. In particular, E4 exhibits the most significant convection effect due to the P\'eclet number ($\operatorname{Pe}\propto \beta/\kappa $) reaching maximum in this example. When $\operatorname{Pe}$ number is high, the convection effect is dominant, and the diffusion effect can be disregarded, which means that the macroscopic motion of the heat flow has a much greater impact than the random motion of particles. Table~\ref{tab:diffusion} presents each method's performance and computational cost in the 1D convection-diffusion equation. Among all unsupervised neural PDE solvers, including PI-DeepONet-(M) and PINO-10/20/100, the MCNP-10 performs the best on most tasks, particularly for cases with large convection or diffusion rates. For PINO, its accuracy is not robust with step size, indicating that the pseudo-spectral method cannot provide sufficiently accurate training targets with a coarser partition of the time interval. Additionally, the accuracy of PINO in E4 declines most rapidly with increasing temporal step size. Such phenomenons happen because the $\operatorname{Pe}$ number in this experiment is significantly higher than in other experiments; hence, the conventional Eulerian method requires a finer time step to maintain simulation accuracy. For PINO-100, despite achieving comparable results on E1 and E5, its training cost is eight times more than that of MCNP-10. It is also interesting to note that MCNP-10 performs better than MCNP-20 in Table~\ref{tab:diffusion}. The underlying reason is that the drift coefficient $\beta$ and the forcing $f$ are constants in Eq.~\ref{eq:heat}, and thus, the discrete errors can be eliminated when simulating the trajectories of particles for Feynman-Kac law. As a result, the setting $\Delta t=2/10$ is already sufficient to provide accurate training signals for the MCNP Solver, while further refining the step size would only add additional fitting and generalization burdens.

\subsection{1D Allen-Cahn Equation}
In this section, we conduct experiments on the 1D Allen-Cahn equation with Dirichlet/Neumann boundary conditions as follows:

\begin{equation}\label{eq:ac}
\begin{aligned}
        \frac{\partial u(x,t)}{\partial t} = & \epsilon \Delta u(x,t) + u(x, t) - u(x, t)^3,\\ 
         &x \in [0, 1], t\in [0, 1];\\
        \operatorname{Dirichlet:}&\ u(0, t) = u(1, t) = 0;\\
        \operatorname{Neuman:}&\ \frac{\partial u(x, t)}{\partial x}\bigg|_{x=0} = \frac{\partial u(x, t)}{\partial x}\bigg|_{x=1} = 0;
\end{aligned}
\end{equation}
The initial states $u(x, 0)$ are generated from the functional space $\mathcal{F}_N \triangleq \{\sum_{n=1}^N a_n \sin(2\pi n x):a_n \sim \mathbb{U}(0, 1)\}$, and $N$ represents the maximum frequency of the functional space. 
\subsubsection{Experimental Settings} {Firstly, we fix $\epsilon$ as 0.01, and select two different $N$ in $\{5, 10\}$ and two kinds of boundary conditions to evaluate the performance of different methods in handling spatial variations and varying boundary conditions. Secondly, when $\epsilon$ tends to zero, the interface layers become thinner and sharper, leading to long-lived metastable structures~\cite{folino2020exponentially}. To observe such phenomena, we choose $\epsilon$ as 0.0001 with $N$ in $\{5, 10\}$, respectively. Therefore, these six experimental settings are denoted E1-E6 with $(\epsilon, N, \operatorname{Boundary})=(0.01, 5, \operatorname{D}), (0.01, 10, \operatorname{D}), (0.01, 5, \operatorname{N}),\\ (0.01, 10, \operatorname{N}), (0.0001, 5, \operatorname{D}), (0.0001, 10, \operatorname{D})$\footnote{"D" represents the Dirichlet boundary condition and "N" represents the Neumann boundary condition.}, respectively.} We divide the spatial domain $[0, 1]$ into $65$ uniform grid elements for all experiments. 

\subsubsection{Baselines} We introduce the neural PDE solvers performed on the 1D Allen-Cahn equations, including: \romannumeral1). \textbf{PINO}~\cite{li2021physics}: an unsupervised neural operator. We divide the time interval into 20/100 uniform frames, denoted as PINO-20/100. The loss function is constructed through the finite difference method, and an additional loss term is involved in enforcing the neural PDE solver that satisfies the boundary conditions. \romannumeral2). \textbf{PI-DeepONet}~\cite{wang2021learning}: an unsupervised neural operator based on PINN loss and DeepONets. {\romannumeral3). \textbf{PI-DeepONet-M}~\cite{wang2022improved}: a modified version of PI-DeepONet.} \romannumeral4). \textbf{MCNP Solver}, we divide the time interval into 20/100 uniform frames, denoted as MCNP-20/100, respectively. {When applying Feynman-Kac law to Allen-Cahn equation, the term $u - u^3$ in Eq.~\ref{eq:ac} is regarded as the external forcing in Eq.~\ref{eq:sde_g}.}

\subsubsection{Results} 
\begin{figure*}[!t]
\begin{center}
\centerline{\includegraphics[width=18cm]{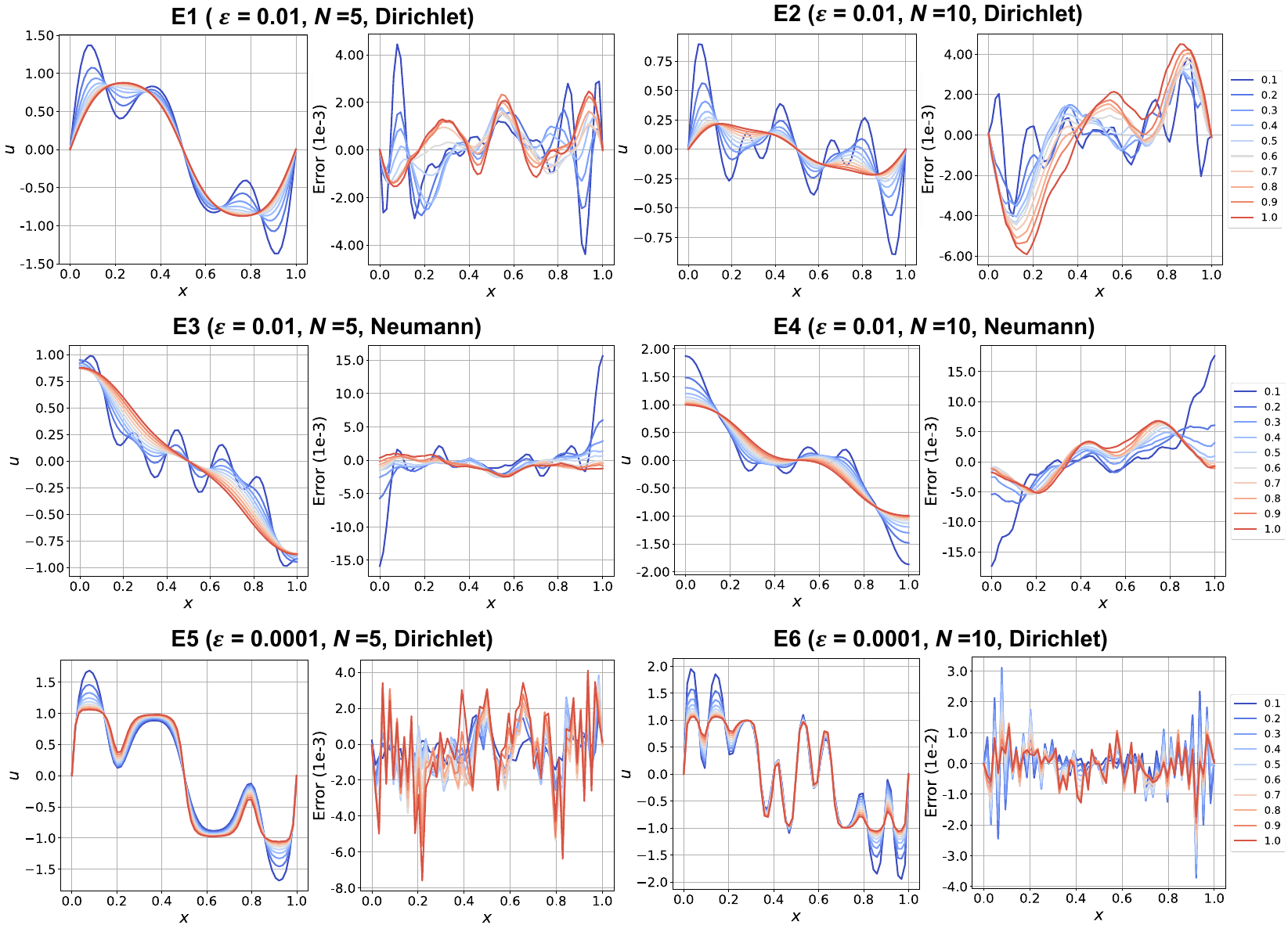}}
\vspace{-5pt}
\caption{{{\textbf{Simulation of 1D Allen-Cahn equation.} The prediction result (Left) and point-wise error (Right) of MCNP-100 for an example in E1-E6. The x-axis and y-axis represent spatial coordinates and the predicted values (point-wise error).}}}
\label{ace_fig}
\end{center}
\vspace{-10pt}
\end{figure*}

{Fig.~\ref{ace_fig} illustrates the predicted $u$ for MCNP-100 from $t = 0.1$ to $t = 1.0$ for E1 through E6, respectively.} The simulation of the Allen-Cahn equation is more challenging than the convection-diffusion equation due to its nonlinearity. We can observe that higher $N$ values can result in more obvious spatial variations, particularly near the initial time. In the experiments with Dirichlet boundary conditions, the system values decay rapidly near the boundary, whereas in the case of Neumann conditions, they maintain a relatively stable variation. From a microscopic perspective, this occurs because the particles are absorbed after hitting the Dirichlet boundary, while they are reflected after colliding with the Neumann boundary. Unlike other baseline methods, MCNP Solver does not rely on additional loss functions to encode the boundary conditions. Instead, the MCNP Solver can naturally satisfy these boundary conditions by choosing the corresponding transition probability during its random walk. The simulation results in Fig.~\ref{ace_fig} demonstrate that the MCNP Solver effectively meets the Dirichlet and Neumann boundary conditions when solving the Allen-Cahn equations. {It can be seen in Fig.~\ref{ace_fig} that when $\epsilon = 0.0001$, the phase field $u(x)$ oscillates rapidly between positive and negative values, highlighting the sharpness of the interface layers due to the very small $\epsilon$. These oscillations are caused by the equation striving to minimize the energy, leading to high-frequency changes in the solution near the interface to accommodate the small $\epsilon$~\cite{folino2020exponentially}.} Table~\ref{tab:ac} presents each method's performance and computational cost on the 1D Allen-Cahn equation. The results of PI-DeepONet indicate that the PINN loss cannot efficiently handle high-frequency components, which has also been observed in previous literature~\cite{krishnapriyan2021characterizing, wang2022and}. Although PI-DeepONet-M mitigates this issue via adaptive sampling, there is still a gap in precision when solving high-frequency problems compared to other methods. Among all unsupervised methods, only MCNP-100 achieves a relative error lower than 1\% on all experiments and metrics. Moreover, the performance of MCNP-20 is generally comparable to that of PINO-100 while taking only around 21\% of the training time, demonstrating the advantages of neural Monte Carlo loss to learn PDEs unsupervised.

\begin{table}[!t]
\centering
\caption{{{\textbf{1D Allen-Cahn equation with varying $\epsilon$, $N$ and boundary conditions.} Relative $\ell_2$ errors ($E_{\ell_2}$), relative $\ell_{\infty}$ errors ($E_{\ell_{\infty}}$) and computational costs for baseline methods and MCNP Solver. The best results are marked in bold, and the second best results are underlined.}}}\label{tab:ac}
\vspace{-5pt}
\resizebox{8.8cm}{!}{\begin{tabular}{cccccc}
\toprule
Task                                     & Model & $E_{\ell_2}$ (\%) & $E_{\ell_{\infty}}$ (\%)& \shortstack{Train\\Time (H)} & \shortstack{Param.\\ (M)}\\ \midrule
\multicolumn{1}{l|}{\multirow{5}{*}{\shortstack{E1\\$\epsilon=0.01$\\$N=5$\\Dirichlet}}} &  PINO-20     & 1.707 \footnotesize{$\pm$0.262}     &    2.082 \footnotesize{$\pm$0.271}  &   0.135  &   0.140     \\
\multicolumn{1}{l|}{}                    &  PINO-100     & \underline{0.639 \footnotesize{$\pm$0.020}}     &  \underline{0.736 \footnotesize{$\pm$0.009}}    &  0.627      &   0.140     \\
\multicolumn{1}{l|}{}                    &    PI-DeepONet   &   2.444 \footnotesize{$\pm$1.022} &  3.551 \footnotesize{$\pm$1.579}   &    0.431   &   0.219     \\
\multicolumn{1}{l|}{}                    &    {PI-DeepONet-M}   &   {1.071\footnotesize{$\pm$ 0.113}}     &    {1.431 \footnotesize{$\pm$0.125}}   &   {1.004}   &   {0.219}     \\
\multicolumn{1}{l|}{}                    &   MCNP-20    &  0.918 \footnotesize{$\pm$0.037}    &  1.108 \footnotesize{$\pm$0.034}  &    0.133   &    0.140    \\
\multicolumn{1}{l|}{}                    &  MCNP-100     &   \textbf{0.547 \footnotesize{$\pm$0.053}}    &   \textbf{0.636 \footnotesize{$\pm$0.061}}  &     0.621   &   0.140     \\ \midrule
\multicolumn{1}{l|}{\multirow{5}{*}{\shortstack{E2\\$\epsilon=0.01$\\$N=10$\\Dirichlet}}}   &  PINO-20     & 3.180 \footnotesize{$\pm$0.182}     &    4.534 \footnotesize{$\pm$0.232}  &   0.135          & 0.140       \\
\multicolumn{1}{l|}{}                    &  PINO-100     & \underline{1.165 \footnotesize{$\pm$0.192}}     &  \underline{1.430 \footnotesize{$\pm$0.193}}    &    0.627  &    0.140    \\
\multicolumn{1}{l|}{}                    &    PI-DeepONet   &  5.615 \footnotesize{$\pm$1.191} &   7.163 \footnotesize{$\pm$1.455}   &  0.431    &   0.219  \\
\multicolumn{1}{l|}{}                    &    {PI-DeepONet-M}   &   {2.659 \footnotesize{$\pm$0.259}}     &    {3.241 \footnotesize{$\pm$0.240}}   &   {1.004}   &   {0.219}     \\
\multicolumn{1}{l|}{}                    &   MCNP-20    & 1.290 \footnotesize{$\pm$0.062}   &     1.592 \footnotesize{$\pm$0.079}  &     0.133     &   0.140     \\
\multicolumn{1}{l|}{}                    &  MCNP-100     &   \textbf{0.831 \footnotesize{$\pm$0.200}}    & \textbf{0.955 \footnotesize{$\pm$0.212}}  &    0.621  &   0.140     \\ \midrule
\multicolumn{1}{l|}{\multirow{5}{*}{\shortstack{E3\\$\epsilon=0.01$\\$N=5$\\Neumann}}} &  PINO-20     & 1.727 \footnotesize{$\pm$0.368}     &    2.885 \footnotesize{$\pm$0.043}  &  0.136   &   0.140     \\
\multicolumn{1}{l|}{}                    &  PINO-100     & 1.705 \footnotesize{$\pm$0.019}     &  2.768 \footnotesize{$\pm$0.043}    &     0.627   &   0.140     \\
\multicolumn{1}{l|}{}                    &    PI-DeepONet   &   1.773 \footnotesize{$\pm$0.890} &   2.353 \footnotesize{$\pm$1.128}   &   0.433     &   0.219     \\
\multicolumn{1}{l|}{}                    &    {PI-DeepONet-M}   &   {\underline{0.438 \footnotesize{$\pm$0.047}}}     &    {\underline{0.617 \footnotesize{$\pm$0.062}}}   &   {1.012}   &   {0.219}     \\
\multicolumn{1}{l|}{}                    &   MCNP-20    &  {0.598 \footnotesize{$\pm$0.011}}    & {0.801 \footnotesize{$\pm$0.018}}  &   0.133    &    0.140    \\
\multicolumn{1}{l|}{}                    &  MCNP-100     &  \textbf{ 0.394 \footnotesize{$\pm$0.037} }   &   \textbf{0.518 \footnotesize{$\pm$0.040}}  &   0.622     &   0.140     \\ \midrule
\multicolumn{1}{l|}{\multirow{5}{*}{\shortstack{E4\\$\epsilon=0.01$\\$N=10$\\Neumann}}}   &  PINO-20     & 2.178 \footnotesize{$\pm$0.368}     &   2.885 \footnotesize{$\pm$0.368}  &     0.136        & 0.140       \\
\multicolumn{1}{l|}{}                    &  PINO-100     & \underline{1.176 \footnotesize{$\pm$0.025}}     &  1.805 \footnotesize{$\pm$0.043}    &   0.627   &    0.140    \\
\multicolumn{1}{l|}{}                    &    PI-DeepONet   &   2.587 \footnotesize{$\pm$0.247} &  3.111 \footnotesize{$\pm$0.300}   &   0.433    &   0.219  \\
\multicolumn{1}{l|}{}                    &    {PI-DeepONet-M}   &   {2.169 \footnotesize{$\pm$0.072}}     &    {2.489 \footnotesize{$\pm$0.091}}   &   {1.012}   &   {0.219}     \\
\multicolumn{1}{l|}{}                    &   MCNP-20    & 1.218 \footnotesize{$\pm$0.016}   &     \underline{1.470 \footnotesize{$\pm$0.026}}  &     0.133     &   0.140     \\
\multicolumn{1}{l|}{}                    &  MCNP-100     &   \textbf{0.582 \footnotesize{$\pm$0.078}}    & \textbf{0.671 \footnotesize{$\pm$0.083}}  &   0.622   &   0.140     \\ 
\midrule
\multicolumn{1}{l|}{\multirow{5}{*}{{\shortstack{E5\\$\epsilon=0.0001$\\$N=5$\\Dirichlet}}}} &  {PINO-20}     & {0.735 \footnotesize{$\pm$0.162}}     &    {1.445 \footnotesize{$\pm$0.275}}  &   {0.135}  &   {0.140}     \\
\multicolumn{1}{l|}{}                    &  {PINO-100}     & {{0.330 \footnotesize{$\pm$0.004}}}     &  {{0.842 \footnotesize{$\pm$0.003}}}    &  {0.627}      &   {0.140}     \\
\multicolumn{1}{l|}{}                    &    {PI-DeepONet}   &   {8.251 \footnotesize{$\pm$2.665}} &  {15.853 \footnotesize{$\pm$4.519}}   &    {0.431}   &   {0.219}     \\
\multicolumn{1}{l|}{}                    &    {PI-DeepONet-M}   &   {3.781\footnotesize{$\pm$ 1.943}}     &    {4.724 \footnotesize{$\pm$2.568}}   &   {1.004}   &   {0.219}     \\
\multicolumn{1}{l|}{}                    &   {MCNP-20}    &  {\textbf{0.285 \footnotesize{$\pm$0.010}}}    &  {\textbf{0.593 \footnotesize{$\pm$0.014}}}  &    {0.156}   &    {0.140}    \\
\multicolumn{1}{l|}{}                    &  {MCNP-100}     &   {\underline{0.303 \footnotesize{$\pm$0.018}}}    &   {\underline{0.594 \footnotesize{$\pm$0.047}}}  &     {0.651}   &   {0.140}     \\ \midrule
\multicolumn{1}{l|}{\multirow{5}{*}{{\shortstack{E6\\$\epsilon=0.0001$\\$N=10$\\Dirichlet}}}} &  {PINO-20}     & {3.605 \footnotesize{$\pm$1.030}}     &    {7.181 \footnotesize{$\pm$1.636}}  &   {0.135}  &   {0.140}     \\
\multicolumn{1}{l|}{}                    &  {PINO-100}     & {{1.817 \footnotesize{$\pm$0.008}}}     &  {{4.114 \footnotesize{$\pm$0.042}}}    &  {0.627}      &   {0.140}     \\
\multicolumn{1}{l|}{}                    &    {PI-DeepONet}   &   {15.009 \footnotesize{$\pm$2.398}} &  {25.870 \footnotesize{$\pm$3.776}}   &    {0.431}   &   {0.219}     \\
\multicolumn{1}{l|}{}                    &    {PI-DeepONet-M}   &   {9.542\footnotesize{$\pm$ 1.058}}     &    {12.357 \footnotesize{$\pm$1.569}}   &   {1.004}   &   {0.219}     \\
\multicolumn{1}{l|}{}                    &   {MCNP-20}    &  {\textbf{1.459 \footnotesize{$\pm$0.007}}}    &  {\textbf{2.957 \footnotesize{$\pm$0.011}}}  &    {0.156}   &    {0.140}    \\
\multicolumn{1}{l|}{}                    &  {MCNP-100}     &   {\underline{1.487 \footnotesize{$\pm$0.090}}}    &   {\underline{3.083 \footnotesize{$\pm$0.230}}}  &     {0.651}   &   {0.140}     \\   
\bottomrule
\end{tabular}}
\vspace{-10pt}
\end{table}

\subsection{2D Navier-Stokes Equation}\label{sec:nse}
In this experiment, we simulate the vorticity field for 2D  incompressible flows in a periodic domain $\Omega=[0,1] \times[0,1]$, whose vortex equation is given as follows:
\begin{equation}\label{eq:nse}
\begin{aligned}
\frac{\partial \omega}{\partial t}  &=-(\boldsymbol{u} \cdot \nabla) \omega+\nu \Delta \omega+f(\boldsymbol{x}), \ \boldsymbol{x}\in \Omega, t\in [0, 10],\\
\omega  & =\nabla \times \boldsymbol{u},
\end{aligned}
\end{equation}
where $f(\boldsymbol{x})$ is the external forcing, $\boldsymbol{u}\in \mathbb{R}^2$ denotes the velocity, and $\nu \in \mathbb{R}^+$ represents the viscosity coefficient. The initial vorticity is generated from the Gaussian random field $\mathcal{N}\left(0,7^{3 / 2}(-\Delta+49 \boldsymbol{I})^{-2.5}\right)$ with periodic boundaries. {In this section, we consider three kinds of external forcing, including: (1) zero forcing $f_1(\boldsymbol{x})\equiv0$; (2) Li forcing $f_2(\boldsymbol{x}) \triangleq 0.1 \sin \left(2 \pi\left(\boldsymbol{x}_1+\boldsymbol{x}_2\right)\right)+0.1 \cos \left(2 \pi\left(\boldsymbol{x}_1+\boldsymbol{x}_2\right)\right)$, which is a classical forcing in the paper of FNO~\cite{DBLP:conf/iclr/LiKALBSA21}; (3) Kolmogorov forcing $f_3(\boldsymbol{x}) \triangleq 0.1\cos(8\pi \boldsymbol{x}_1)$, which can result in a much wider range of trajectories due to the chaotic~\cite{smaoui2021control}}.

\subsubsection{Experimental Settings} The viscosity coefficient $\nu$ can be regarded as a measure of the spatiotemporal complexity of the Navier-Stokes equation. As $\nu$ decreases, the nonlinear term $(\boldsymbol{u} \cdot \nabla)\omega$ gradually governs the motion of fluids, increasing the difficulty of simulation. To evaluate the performance of handling different degrees of turbulence, we conduct the experiments with $\nu$ in $\{10^{-4}, 10^{-5}\}$. {Therefore, six experimental settings are denoted E1-E6 with $(f(\boldsymbol{x}), \nu)=(f_1(\boldsymbol{x}), 10^{-4}), (f_1(\boldsymbol{x}), 10^{-5}), (f_2(\boldsymbol{x}), 10^{-4}), (f_2(\boldsymbol{x}), 10^{-5}), \\(f_3(\boldsymbol{x}), 10^{-4}), (f_3(\boldsymbol{x}), 10^{-5})$, respectively.} We divide the domain $\Omega$ into $64\times 64$ uniform grid elements. 

\begin{figure*}[!t]
\begin{center}
\vspace{-5pt}
\centerline{\includegraphics[width=18.5cm]{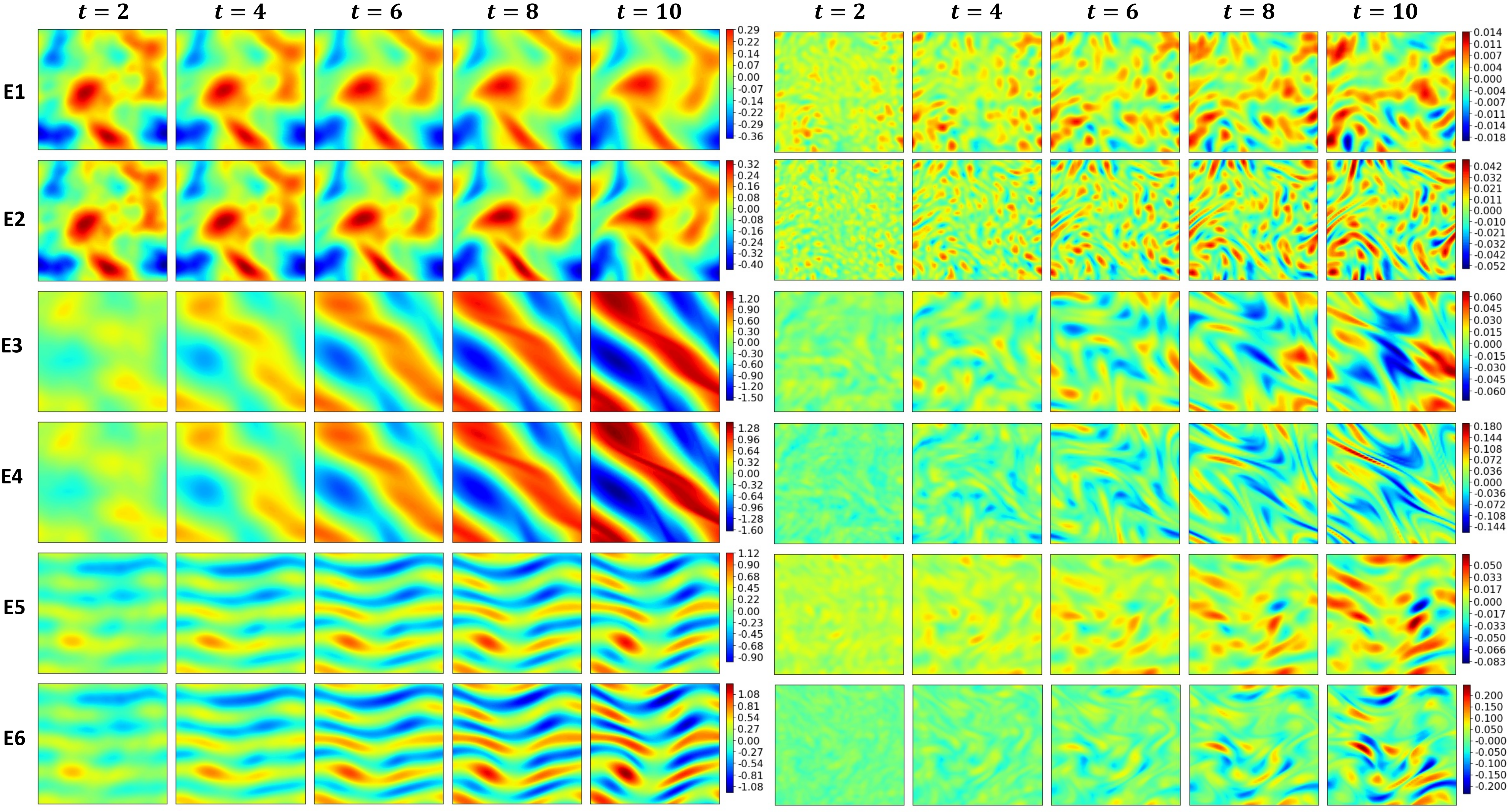}}
\caption{{{\textbf{Simulation of 2D Navier-Stokes equation.} The prediction result (Left) and point-wise error (Right) of MCNP-20 for an example in E1-E6.}}}
\label{ns_fig}
\end{center}
\vspace{-10pt}
\end{figure*}

\begin{table}[!t]
\centering
\caption{{\textbf{2D Navier-Stokes equation with varying $\nu$ and forcing.} Relative $\ell_2$ errors ($E_{\ell_2}$), relative $\ell_{\infty}$ errors ($E_{\ell_{\infty}}$) and computational costs for baseline methods and MCNP Solver. The best results are marked in bold, and the second best results are underlined. The training time for MCNP varies among the six experiments due to the interpolation trick.}}\label{tab:nse}
\vspace{-5pt}
\resizebox{8.8cm}{!}{\begin{tabular}{cccccc}
\toprule
Task                                     & Model & $E_{\ell_2}$ (\%) & $E_{\ell_{\infty}}$ (\%)& \shortstack{Train\\Time (H)} & \shortstack{Param.\\(M)}\\ \midrule
\multicolumn{1}{c|}{\multirow{5}{*}{\shortstack{{E1} \\{$\nu=10^{-4}$}\\{Zero}}}} &  {PINO-10}     & {3.693 \footnotesize{$\pm$0.081}}     &    {6.195 \footnotesize{$\pm$0.084}}  & {0.232}    &      {5.319} \\
\multicolumn{1}{l|}{}                    &  {PINO-20}     & {2.749 \footnotesize{$\pm$0.091}}     &  {4.654 \footnotesize{$\pm$0.129}}    &  {0.413}     &    {5.319}   \\
\multicolumn{1}{l|}{}                    &    {PINO-100}   &   {3.027 \footnotesize{$\pm$0.134}} &  {5.155 \footnotesize{$\pm$0.188}}   &  {1.919}     &       {5.319} \\
\multicolumn{1}{l|}{}                    &   {SP-FNO-10}    &   {4.348 \footnotesize{$\pm$}0.077}    &  {7.597 \footnotesize{$\pm$}0.084} &    {$>12$}   &     {5.319}  \\
\multicolumn{1}{l|}{}                    &   {SP-FNO-20}    &  {3.821 \footnotesize{$\pm$0.264}}    &  {6.855 \footnotesize{$\pm$0.337}}  &    {$>24$}   &     {5.319}  \\
\multicolumn{1}{l|}{}                    &   {MCNP-10}    &  {\textbf{2.397 \footnotesize{$\pm$0.101}}}    &  {\textbf{4.102 \footnotesize{$\pm$0.139}}}  &    {0.256}   &     {5.319}  \\
\multicolumn{1}{l|}{}                    &  {MCNP-20}     &   {\underline{2.680 \footnotesize{$\pm$0.130}}}    &   {\underline{4.565 \footnotesize{$\pm$0.202}}}  &     {0.488}   &    {5.319}    \\ \midrule
\multicolumn{1}{c|}{\multirow{5}{*}{\shortstack{{E2}\\{$\nu=10^{-5}$}\\ {Zero}}}} &  {PINO-10}     & {10.035 \footnotesize{$\pm$0.229}}     &   {14.685 \footnotesize{$\pm$0.250}}  &    {0.232} &     {5.319}  \\
\multicolumn{1}{l|}{}                    &  {PINO-20}     &  {8.464 \footnotesize{$\pm$0.402}}     &  {12.596 \footnotesize{$\pm$0.441}}    &    {0.413}    &   {5.319}    \\
\multicolumn{1}{l|}{}                    &    {PINO-100}   &   {\underline{7.554 \footnotesize{$\pm$0.206}}} &  {11.627 \footnotesize{$\pm$0.384}}   &  {1.919}     &    {5.319}    \\
\multicolumn{1}{l|}{}                    &   {SP-FNO-10}    &  {9.023 \footnotesize{$\pm$}0.154}    &  {14.487 \footnotesize{$\pm$}0.208}   &    {$>12$}   &     {5.319}  \\
\multicolumn{1}{l|}{}                    &   {SP-FNO-20}    &  {8.677 \footnotesize{$\pm$}0.090}    &  {13. 217\footnotesize{$\pm$}0.183}  &    {$>24$}   &     {5.319}  \\
\multicolumn{1}{l|}{}                    &   {MCNP-10}    &  {\textbf{7.055 \footnotesize{$\pm$0.123}}}    &  {\textbf{10.815 \footnotesize{$\pm$0.138}}}  &    {0.256}   &     {5.319}  \\
\multicolumn{1}{l|}{}                    &  {MCNP-20}     &   {{7.620 \footnotesize{$\pm$0.184}}}    &   {\underline{11.564 \footnotesize{$\pm$0.201}}}  &     {0.527}   &    {5.319}    \\ \midrule
\multicolumn{1}{c|}{\multirow{5}{*}{\shortstack{E3 \\$\nu=10^{-4}$\\Li}}} &  PINO-10     & 5.502 \footnotesize{$\pm$0.040}     &    9.002 \footnotesize{$\pm$0.056}  & 0.232    &      5.319 \\
\multicolumn{1}{l|}{}                    &  PINO-20     & 3.971 \footnotesize{$\pm$0.115}     &  6.819 \footnotesize{$\pm$0.184}    &  0.413     &    5.319   \\
\multicolumn{1}{l|}{}                    &    PINO-100   &   3.366 \footnotesize{$\pm$0.034} &  5.908 \footnotesize{$\pm$0.041}   &  1.919     &       5.319 \\
\multicolumn{1}{l|}{}                    &   {SP-FNO-10}    &  {3.794\footnotesize{$\pm$} 0.076}    &  {6.885 \footnotesize{$\pm$}0.109}  &    {$>12$}   &     {5.319}  \\
\multicolumn{1}{l|}{}                    &   {SP-FNO-20}    &  {3.638 \footnotesize{$\pm$0.056}}    &  {6.516 \footnotesize{$\pm$0.082}}  &    {$>24$}   &     {5.319}  \\
\multicolumn{1}{l|}{}                    &   MCNP-10    &  \underline{3.101 \footnotesize{$\pm$0.043}}    &  \underline{5.815 \footnotesize{$\pm$0.075}}  &    0.261   &     5.319  \\
\multicolumn{1}{l|}{}                    &  MCNP-20     &   \textbf{2.999 \footnotesize{$\pm$0.082}}    &   \textbf{5.336 \footnotesize{$\pm$0.131}}  &     0.491   &    5.319    \\ \midrule
\multicolumn{1}{c|}{\multirow{5}{*}{\shortstack{E4\\$\nu=10^{-5}$\\Li}}} &  PINO-10     & 10.617 \footnotesize{$\pm$0.166}     &   16.799 \footnotesize{$\pm$0.255}  &    0.232 &     5.319  \\
\multicolumn{1}{l|}{}                    &  PINO-20     & 8.124 \footnotesize{$\pm$0.261}     &  13.636 \footnotesize{$\pm$0.351}    &    0.413    &   5.319    \\
\multicolumn{1}{l|}{}                    &    PINO-100   &   6.439 \footnotesize{$\pm$0.105} &  11.123 \footnotesize{$\pm$0.139}   &  1.919     &    5.319    \\
\multicolumn{1}{l|}{}                    &   {SP-FNO-10}    &  {6.561 \footnotesize{$\pm$0.098}}    &  {11.529\footnotesize{$\pm$} 0.134}  &    {$>12$}   &     {5.319}  \\
\multicolumn{1}{l|}{}                    &   {SP-FNO-20}    &  {6.320 \footnotesize{$\pm$}0.109}    &  {11.214 \footnotesize{$\pm$}0.144}  &    {$>24$}   &     {5.319}  \\
\multicolumn{1}{l|}{}                    &   MCNP-10    &  \underline{5.825 \footnotesize{$\pm$0.068}}    &  \underline{10.452 \footnotesize{$\pm$0.122}}  &   0.260    &    5.319   \\
\multicolumn{1}{l|}{}                    &  MCNP-20     &   \textbf{5.776 \footnotesize{$\pm$0.085}}    &   \textbf{10.137 \footnotesize{$\pm$0.146}}  &     0.531   &   5.319     \\ \midrule
\multicolumn{1}{c|}{\multirow{5}{*}{\shortstack{E5\\$\nu=10^{-4}$\\Kolmogorov}}} &  PINO-10     & 6.652 \footnotesize{$\pm$0.141}     &    8.922 \footnotesize{$\pm$0.205}  &  0.232   &     5.319  \\
\multicolumn{1}{l|}{}                    &  PINO-20     & 5.164 \footnotesize{$\pm$0.138}     &  7.195 \footnotesize{$\pm$0.196}    &     0.413   &     5.319  \\
\multicolumn{1}{l|}{}                    &    PINO-100   &   4.925 \footnotesize{$\pm$0.037} &  6.972 \footnotesize{$\pm$0.042}   &    1.919   &  5.319      \\
\multicolumn{1}{l|}{}                    &   {SP-FNO-10}    &  {6.801 \footnotesize{$\pm$}0.124}    &  {9.581 \footnotesize{$\pm$}0.272}  &    {$>12$}   &     {5.319}  \\
\multicolumn{1}{l|}{}                    &   {SP-FNO-20}    &  {6.682 \footnotesize{$\pm$}0.040}    &  {9.525 \footnotesize{$\pm$}0.049}  &    {$>24$}   &     {5.319}  \\
\multicolumn{1}{l|}{}                    &   MCNP-10    &  \underline{4.799 \footnotesize{$\pm$0.114}}    &  \textbf{6.443 \footnotesize{$\pm$0.086}}  &   0.278   &    5.319   \\
\multicolumn{1}{l|}{}                    &  MCNP-20     &   \textbf{4.609 \footnotesize{$\pm$0.048}}    &   \underline{6.488 \footnotesize{$\pm$0.053}}  &     0.518   &    5.319    \\ \midrule
\multicolumn{1}{c|}{\multirow{5}{*}{\shortstack{E6\\$\nu=10^{-5}$\\Kolmogorov}}} &  PINO-10     & 16.342 \footnotesize{$\pm$0.345}     &   22.497 \footnotesize{$\pm$0.266}  &   0.232  &   5.319    \\
\multicolumn{1}{l|}{}                    &  PINO-20     & 13.501 \footnotesize{$\pm$0.446}     &  19.200 \footnotesize{$\pm$0.748}    &  0.413      &   5.319      \\
\multicolumn{1}{l|}{}                    &    PINO-100   &   11.874 \footnotesize{$\pm$0.235} &  16.973 \footnotesize{$\pm$0.204}   & 
 1.919      &  5.319      \\
 \multicolumn{1}{l|}{}                    &   {SP-FNO-10}    &  {13.789 \footnotesize{$\pm$}0.429}    &  {19.004 \footnotesize{$\pm$}0.517}  &    {$>12$}   &     {5.319}  \\
\multicolumn{1}{l|}{}                    &   {SP-FNO-20}    &  {13.601 \footnotesize{$\pm$}0.390}    &  {19.214 \footnotesize{$\pm$}0.445}  &    {$>24$}   &     {5.319}  \\
\multicolumn{1}{l|}{}                    &   MCNP-10    &  \textbf{10.161 \footnotesize{$\pm$0.150}}    &  \underline{15.053 \footnotesize{$\pm$0.227}}  &    0.252   &   5.319    \\
\multicolumn{1}{l|}{}                    &  MCNP-20     &   \underline{10.829 \footnotesize{$\pm$0.154}}    &   \textbf{15.764 \footnotesize{$\pm$0.196}}  &    0.528    &   5.319     \\ \bottomrule
\end{tabular}}
\vspace{-10pt}
\end{table}

\subsubsection{Baselines} We introduce the baselines conducted on 2D Navier-Stokes equation, including:\footnote{For PI-DeepONets~\cite{wang2021learning, wang2022improved}, they only conduct experiments on time-independent PDE in 2D situations in their paper.} \romannumeral1). \textbf{PINO}~\cite{li2021physics}: we divide the time interval $[0, 10]$ into 10/20/100 uniform frames, denoted as PINO-10/20/100, respectively. The loss function is constructed through the pseudo-spectral method due to the periodic boundary conditions. {\romannumeral2). \textbf{SP-FNO} ~\cite{navaneeth2023stochastic, navaneeth2024physics}, an unsupervised operator learning method that employs stochastic projection (SP) to enhance the precision of spatial gradient computation. We use FNO as the backbone model for a fair comparison and divide the time interval into uniform 10/20 frames, denoted as SP-FNO-10/20.} \romannumeral3). \textbf{MCNP Solver}, we divide the time interval into uniform 10/20 frames, denoted as MCNP-10/20, respectively. {When applying Feynman-Kac law to Navier-Stokes equation, the velocity $\boldsymbol{u}$ in Eq.~\ref{eq:nse} is regarded as the drift term $\boldsymbol{\beta}$ in Eq.~\ref{eq:sde_g}.}

\subsubsection{Results} 
{Fig.~\ref{ns_fig} shows the predicted vorticity field $\omega$ of a learned MCNP Solver from $t = 2$ to $t = 10$ in E1-E6, respectively. Compared to the case with $\nu=10^{-4}$, we observe that the vorticity field has more remarkable peak and trough values with more intricate details when $\nu=10^{-5}$, which implies that the fluid elements are rotating at higher speeds. This instability is caused by the nonlinear convection terms gradually taking control of the motion of fluids. Additionally, the external forcing $f$ significantly influences the trend of fluid motion. When the external forcing is zero, the fluid motion primarily arises from internal dynamical processes within the system, such as the interaction between viscous and inertial forces, resulting in a relatively slow change in the vorticity field over time. When an external forcing drives the system, the corresponding vorticity field changes markedly in response to the forcing, leading to a faster evolution of the vorticity field. In particular, when driven by a Kolmogorov forcing, numerous new vortices of different scales are generated during the flow, making the fluid simulation more challenging.} More simulation results of other baseline methods can be seen in the Appendix~~\uppercase\expandafter{\romannumeral5}. Table~\ref{tab:nse} presents each method's performance and computational cost on the 2D Navier-Stokes equations. {Compared to other unsupervised training methods, the MCNP Solver achieved the lowest relative error for all tasks and metrics.} Unlike PINO, whose performance is highly affected by step size, especially for cases with $\nu=10^{-5}$, the MCNP Solver can maintain stability with respect to step size. It is worth mentioning that MCNP-10 outperforms PINO-100 on all metrics while only taking 13.1-14.5\% of the training time. Such an advantage stems from the fact that the Lagrangian method can be better adapted to coarse step size, as discussed in~\cite{irisov2011numerical, kramer2001review}.

\subsection{{2D Fractional Diffusion Equation on a Disk}}\label{sec:fractional}
{In this section, we simulate the fractional diffusion equation on a 2D unit circle as follows:}
\begin{equation}\label{eq:fractional}
\begin{aligned}
     {\frac{\partial u}{\partial t}} & {= -\kappa (-\Delta)^{\alpha/2} u(\boldsymbol{x},t),\quad  \|\boldsymbol{x}\|_2\leq 1 , t\in [0,1],}   \\
    & {u(\boldsymbol{x},t)}  {=0\quad \text{for}\  \|\boldsymbol{x}\|_2=1,}
\end{aligned}
\end{equation}
{where $(-\Delta)^{\alpha/2}$ denotes the fractional Laplacian operator, defined via the following hyper-singular integral~\cite{lischke2020fractional}:
}
\begin{equation}
{(-\Delta)^{\alpha / 2} u(\boldsymbol{x}) \triangleq C_{\alpha} \text { P.V. } \int_{\mathbb{R}^2} \frac{u(\boldsymbol{x})-u(\boldsymbol{y})}{\|\boldsymbol{x}-\boldsymbol{y}\|_2^{2+\alpha}} \mathrm{d} \boldsymbol{y}, \  0<\alpha<2,}
\end{equation}
{where P.V. denotes the principle value and $C_{\alpha}$ is defined as:}
\begin{equation}
    {C_{\alpha}=\frac{2^\alpha \Gamma\left(\frac{\alpha+2}{2}\right)}{\pi|\Gamma(-\alpha / 2)|}.}
\end{equation}
{As shown in Eq.~\ref{eq:fractional}, calculating the fractional Laplacian operator is challenging due to its singularity and non-local properties. However, we can efficiently simulate it from the probability perspective, where we only need to replace the Brownian motion in Eq.~\ref{eq:bsde} with the $\alpha$-stable L\'evy process~\cite{kozubowski2006fractional, zhang2012stochastic, zhang2012stochastic2}. The initial states $u(\boldsymbol{x}, 0)$ are generated from the functional space $\mathcal{H}_N \triangleq \{\sum_{n=1}^N a_n (1- \|\boldsymbol{x}\|_2^2)^{n+1}:a_n \sim \mathbb{U}(0, 1)\}$, and $N$ represents the maximum degree of the functional space.}

\subsubsection{{Experimental Settings}}
{We select four different $\alpha$ in $\{0.5, 1.0, 1.5, 2.0\}$ to evaluate the performance of MCNP Solver in handling varying fractional coefficients. These four experiments are denoted as E1-E4. We divide the time interval $[0, 1]$ into $100$ uniform intervals. We set the maximum degree $N$ as 10. 
Because Eq.~\ref{eq:fractional} is defined on a disk, we use DeepONets as backbone models and train MCNP Solver in a mesh-free regime.}

\begin{figure*}[!t]
\begin{center}
\centerline{\includegraphics[width=17cm]{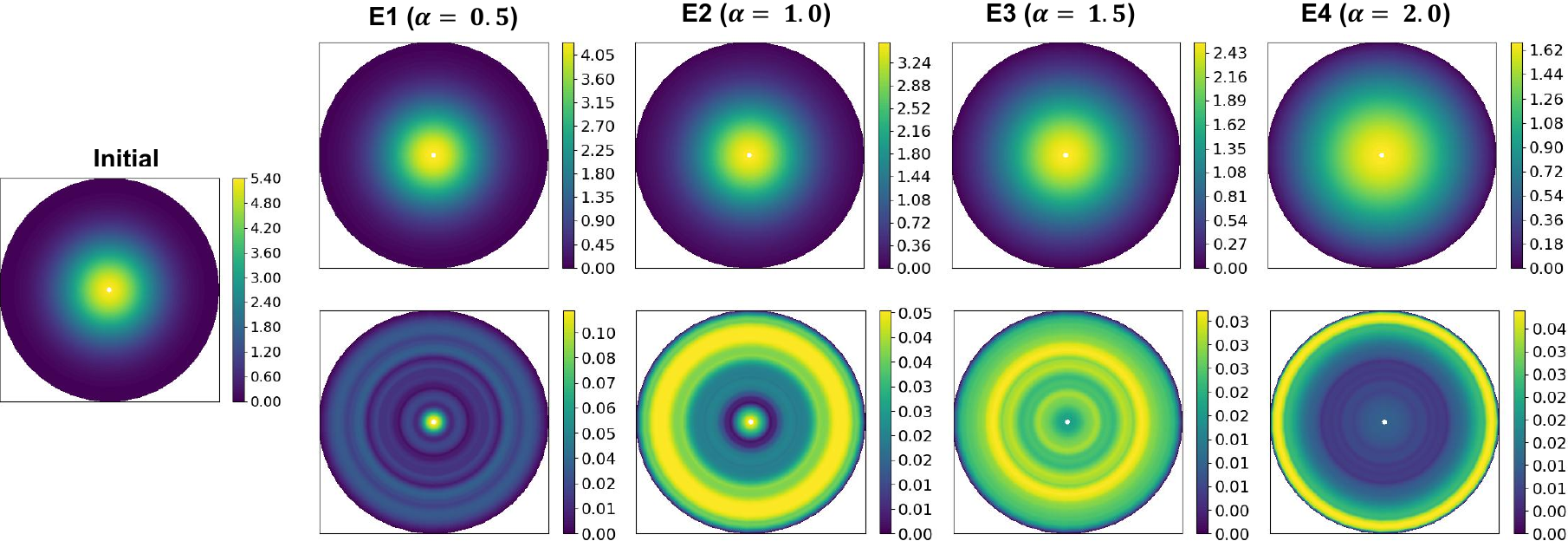}}
\caption{{{\textbf{Simulation of 2D fractional diffusion equation on a disk.} The prediction result (Above) and point-wise error (Below) of MCNP Solver for an example in E1-E4 at $t=1.0$.}}}
\label{fractional_fig}
\end{center}
\vspace{-10pt}
\end{figure*}

\begin{table}[!t]
\centering
\caption{{\textbf{2D fractional diffusion equation with varying $\alpha$.} Relative $\ell_2$ errors ($E_{\ell_2}$), relative $\ell_{\infty}$ errors ($E_{\ell_{\infty}}$) and computational costs.}}\label{tab:fractional}
\vspace{-5pt}
\resizebox{8cm}{!}{
\begin{tabular}{ccccc}
\toprule
{Task}     & {$E_{\ell_2}$ (\%)} & {$E_{\ell_{\infty}}$ (\%)} & {\shortstack{Train\\Time (H)}} & {\shortstack{Param.\\(M)}} \\ \midrule
\multicolumn{1}{l|}{{E1 ($\alpha=0.5$)}} &   {1.139 \footnotesize{$\pm$}0.145}  &  {1.689 \footnotesize{$\pm$}0.320}   &   0.139             &    {0.325}          \\
\multicolumn{1}{l|}{{E2 ($\alpha=1.0$)}} &  {1.538 \footnotesize{$\pm$}0.616}  & {1.846 \footnotesize{$\pm$}1.015}   &    0.139               &     {0.325}         \\
\multicolumn{1}{l|}{{E3 ($\alpha=1.5$)}} &    {1.424 \footnotesize{$\pm$}0.266}  & {1.386 \footnotesize{$\pm$}0.315}  &   0.139                &     {0.325}         \\
\multicolumn{1}{l|}{{E4 ($\alpha=2.0$)}} &  {0.977 \footnotesize{$\pm$}0.045}  & {1.300 \footnotesize{$\pm$}0.132}    &      0.139              &   {0.325}          \\ \bottomrule
\end{tabular}}
\vspace{-5pt}
\end{table}

\subsubsection{{Results}}
Fig.~\ref{fractional_fig} shows the predicted physical field $u$ of a learned MCNP Solver at $t = 1.0$ in E1-E4, respectively. Comparing different $\alpha$ values, we observe that the diffusion effect becomes more pronounced as the fractional coefficient $\alpha$ increases. Table~\ref{tab:fractional} presents each method's performance and computational cost on the 2D fractional diffusion equation. The results reveal that the MCNP Solver has consistently maintained an error rate under 2\% for all tasks and performance indicators. It is interesting to note that the precision of the simulation does not vary monotonically with changes in the fractional coefficient $\alpha$. This phenomenon happens because excessively large or small values of $\alpha$ present challenges to the simulation tasks. For smaller $\alpha$ values, the singularities of the fractional operators would become more remarkable. From a microscopic perspective, particles are more likely to undergo significant jumps following a Lévy process. Conversely, the diffusion rate accelerates with larger $\alpha$ values, leading to a more rapid evolution of the entire physical field. It is worth mentioning that despite the non-rectangular geometry in this section, the MCNP Solver can be extended to a mesh-free regime and preserve the boundary conditions automatically through the random walk of particles.

\section{Additional Experimental Results}\label{sec:addexp}

In this section, we conduct several additional numerical experiments to comprehensively compare the scopes and advantages of the proposed MCNP Solver with other widely-used PDE solvers, including the Monte Carlo methods, traditional Eulerian methods and supervised solver learning methods. Furthermore, we conduct ablation studies on MCNP Solver to study the effects of the proposed tricks.

\begin{table*}[!t]
\caption{{\textbf{In comparing the relative $\ell_2$ errors (\%) between MCNP and other numerical solvers across four types of equations.} In column 9, we report the inference time (S), denoted as $T_{\operatorname{GPU}}/T_{\operatorname{CPU}}$, for a single test sample on both GPU and CPU. The inference time for the fractional PDE on the GPU is not included, as its implementation is based on a CPU package. {In column 10, we report the training time of MCNP for each experiment.}}}\label{tab:solver}
\vspace{-5pt}
\centering
\resizebox{18cm}{!}{
\begin{tabular}{cccccccccc}
\toprule
\multicolumn{1}{c|}{{Equation}}          & \multicolumn{1}{c|}{{Solver}} &    {E1}                   &      {E2}                &           {E3}          &  {E4}                   &          {E5}            &          {E6}           &       {$T_{\operatorname{GPU}}/T_{\operatorname{CPU}}$ (S)}       &      {Train Time (H)}             \\ \midrule
\multicolumn{1}{c|}{\multirow{8}{*}{{\shortstack{Convection-\\Diffusion}}}} & \multicolumn{1}{c|}{{MCM}}       &  {3.374 \footnotesize{$\pm$0.047}}                     &      {4.648 \footnotesize{$\pm$0.129}}                 &          {7.269 \footnotesize{$\pm$0.382}}             &         {7.269 \footnotesize{$\pm$0.382}}             &            {4.612 \footnotesize{$\pm$0.098}}           &       {7.248 \footnotesize{$\pm$0.301}}               &           {0.163 / 0.236}        &           {--}      \\
\multicolumn{1}{c|}{}                  & \multicolumn{1}{c|}{{MCM+}}       &   {1.932 \footnotesize{$\pm$0.008}}                  &      {1.440 \footnotesize{$\pm$0.050}}                 &          {2.144 \footnotesize{$\pm$0.045}}             &          {6.952 \footnotesize{$\pm$0.010}}             &            {1.427 \footnotesize{$\pm$0.031}}           &       {2.097 \footnotesize{$\pm$0.065}}                &           {0.188 / 0.518}       &           {--}       \\
\multicolumn{1}{c|}{}                  & \multicolumn{1}{c|}{{MCM++}}       &   {1.718 \footnotesize{$\pm$0.003}}                    &      {0.476 \footnotesize{$\pm$0.013}}                 &          {0.718 \footnotesize{$\pm$0.036}}             &          {6.896 \footnotesize{$\pm$0.010}}              &            {0.489 \footnotesize{$\pm$0.011}}           &       {0.737 \footnotesize{$\pm$0.022}}                &           {0.279 / 3.045}     &           {--}         \\
\multicolumn{1}{c|}{}                  & \multicolumn{1}{c|}{{PSM}}       &  {0.115}                     &      {1.121}                 &          {2.640}             &         {3.278}              &            {2.428}           &       {3.686}                &           {0.005 / 0.003}      &           {--}        \\       
\multicolumn{1}{c|}{}                  & \multicolumn{1}{c|}{{PSM+}}       &  {0.056}                     &      {0.312}                 &          {0.644}             &         {0.733}              &            {0.588}           &       {0.843}                &           {0.007 / 0.005}         &           {--}   \\                         
\multicolumn{1}{c|}{}                 & \multicolumn{1}{c|}{{PSM++}}       &  {0.043}                     &      {0.101}                 &          {0.165}             &         {0.184}              &            {0.160}           &       {0.209}                &           {0.014 / 0.010}      &           {--}    \\      
\multicolumn{1}{c|}{}                  & \multicolumn{1}{c|}{{MCNP}}       &  {0.090 \footnotesize{$\pm$0.002}}                     &      {0.128 \footnotesize{$\pm$0.014}}                 &          {0.170 \footnotesize{$\pm$0.012}}             &         {0.187 \footnotesize{$\pm$0.007}}              &            {0.172 \footnotesize{$\pm$0.018}}           &       {0.200 \footnotesize{$\pm$0.012}}                &           {0.002 / 0.009}    &           {0.032-0.054}        \\          \midrule

\multicolumn{1}{c|}{\multirow{8}{*}{{\shortstack{Allen-\\Cahn}}}} & \multicolumn{1}{c|}{{MCM}}       &  {7.430 \footnotesize{$\pm$0.102}}                     &      {7.458 \footnotesize{$\pm$0.096}}                 &          {16.155 \footnotesize{$\pm$0.343}}             &         {18.133 \footnotesize{$\pm$0.459}}              &            {7.368 \footnotesize{$\pm$0.203}}           &       {12.068 \footnotesize{$\pm$0.023}}                &           {0.048 / 0.257}      &           {--}      \\
\multicolumn{1}{c|}{}                  & \multicolumn{1}{c|}{{MCM+}}       &  {2.902 \footnotesize{$\pm$0.059}}                       &      {2.992 \footnotesize{$\pm$0.071}}                 &          {8.772 \footnotesize{$\pm$0.570}}             &         {10.079 \footnotesize{$\pm$0.521}}           &       {3.449 \footnotesize{$\pm$0.048}}     &            {10.821 \footnotesize{$\pm$0.068}}                          &           {0.063 / 2.692}      &           {--}      \\
\multicolumn{1}{c|}{}                  & \multicolumn{1}{c|}{{MCM++}}       &  {2.103 \footnotesize{$\pm$0.014}}                       &      {2.129 \footnotesize{$\pm$0.028}}                 &         {1.265 \footnotesize{$\pm$0.002}}             &         {1.848 \footnotesize{$\pm$0.005}}                    &       {2.590 \footnotesize{$\pm$0.022}}            &            {7.110 \footnotesize{$\pm$0.040}}         &           {0.551 / 14.568}        &           {--}    \\
\multicolumn{1}{c|}{}                  & \multicolumn{1}{c|}{{FDM}}       &  {NAN}                     &      {NAN}                 &          {NAN}             &         {NAN}              &            {1.812}           &       {12.441}                &           {0.005 / 0.002}       &           {--}     \\            

\multicolumn{1}{c|}{}                  & \multicolumn{1}{c|}{{FDM+}}       &  {0.983}                     &      {1.468}                 &          {1.256}             &         {2.166}              &            {0.480}           &       {1.987}                &       {0.021 / 0.009}        &           {--}     \\    

\multicolumn{1}{c|}{}                 & \multicolumn{1}{c|}{{FDM++}}       &  {0.625}                     &      {0.788}              &            {1.424}    &          {1.140}             &                         {0.366}           &       {1.886}                &           {0.043 / 0.018}     &           {--}       \\      
\multicolumn{1}{c|}{}                & \multicolumn{1}{c|}{{MCNP}}     &  {0.547 \footnotesize{$\pm$0.053}}                     &      {0.831 \footnotesize{$\pm$0.200}}                 &          {0.394 \footnotesize{$\pm$0.037}}             &         {0.582 \footnotesize{$\pm$0.078}}              &         {0.285 \footnotesize{$\pm$0.010}}           &       {1.487 \footnotesize{$\pm$0.090}}      &           {0.002 / 0.009}     &           {0.133-0.651}       \\        \midrule

\multicolumn{1}{c|}{\multirow{8}{*}{{\shortstack{Navier-\\Stokes}}}} & \multicolumn{1}{c|}{{MCM}}       &  {7.326 \footnotesize{$\pm$0.035}}                     &      {7.452 \footnotesize{$\pm$0.045}}                 &  {5.864 \footnotesize{$\pm$0.032}}                     &      {4.884 \footnotesize{$\pm$0.012}}            &            {6.890 \footnotesize{$\pm$0.026}}           &       {6.994 \footnotesize{$\pm$0.011}}                &           {0.180 / 4.552}       &           {--}     \\
\multicolumn{1}{c|}{}                  & \multicolumn{1}{c|}{{MCM+}}       &  {5.396 \footnotesize{$\pm$0.043}}                     &      {7.055 \footnotesize{$\pm$0.010}}                 &          {4.523 \footnotesize{$\pm$0.026}}             &         {4.648 \footnotesize{$\pm$0.007}}              &            {5.316 \footnotesize{$\pm$0.035}}           &       {6.700 \footnotesize{$\pm$0.014}}                &           {0.225 / 7.968}    &           {--}        \\
\multicolumn{1}{c|}{}                  & \multicolumn{1}{c|}{{MCM++}}        &  {3.744 \footnotesize{$\pm$0.006}}                     &      {6.814 \footnotesize{$\pm$0.005}}                 &          {3.453 \footnotesize{$\pm$0.008}}             &         {4.491 \footnotesize{$\pm$0.003}}              &            {4.023 \footnotesize{$\pm$0.025}}           &       {6.519 \footnotesize{$\pm$0.011}}                &           {0.469 / 16.647}       &           {--}     \\
\multicolumn{1}{c|}{}                  & \multicolumn{1}{c|}{{PSM}}       &  {4.583}                     &      {12.336}                 &          {6.116}             &         {11.518}              &            {4.500}           &      {9.920}          &           {0.069 / 0.107}        &           {--}    \\       
\multicolumn{1}{c|}{}                  & \multicolumn{1}{c|}{{PSM+}}       &  {4.227}                     &      {11.427}                 &          {3.508}             &         {8.443}              &            {4.122}           &       {9.335}                &           {0.139 / 0.204}      &           {--}      \\            
\multicolumn{1}{c|}{}                  & \multicolumn{1}{c|}{{PSM++}}       &  {4.102}                     &      {11.124}                 &          {2.946}             &         {6.789}              &            {4.007}           &      {9.147}          &           {0.344 / 0.513}     &           {--}       \\                         
\multicolumn{1}{c|}{}                  & \multicolumn{1}{c|}{{MCNP}}       &  {2.397 \footnotesize{$\pm$0.101}}                     &      {7.055 \footnotesize{$\pm$0.123}}                 &          {2.999 \footnotesize{$\pm$0.082}}             &        {5.776 \footnotesize{$\pm$0.085}}             &            {4.609 \footnotesize{$\pm$0.048}}            &       {10.161 \footnotesize{$\pm$0.150}}                &           {0.007 / 0.065}      &           {0.252-0.531}      \\           \midrule

\multicolumn{1}{c|}{\multirow{8}{*}{{Fractional}}} & \multicolumn{1}{c|}{{MCM}}       &  {5.668 \footnotesize{$\pm$0.629}}                     &      {5.710 \footnotesize{$\pm$0.886}}                 &          {5.483 \footnotesize{$\pm$0.866}}             &         {6.511 \footnotesize{$\pm$0.198}}              &            {--}           &       {--}                &           {-- / 
 0.058}        &           {--}    \\
\multicolumn{1}{c|}{}                  & \multicolumn{1}{c|}{{MCM+}}       &  {5.139 \footnotesize{$\pm$0.219}}                     &      {5.330 \footnotesize{$\pm$0.236}}                 &          {4.074 \footnotesize{$\pm$0.106}}             &         {2.821 \footnotesize{$\pm$0.168}}              &            {--}           &       {--}                &           {-- / 0.518}       &           {--}     \\
\multicolumn{1}{c|}{}                  & \multicolumn{1}{c|}{{MCM++}}       &  {5.057 \footnotesize{$\pm$0.066}}                     &      {5.129 \footnotesize{$\pm$0.056}}                 &          {3.799 \footnotesize{$\pm$0.088}}             &         {2.015 \footnotesize{$\pm$0.010}}              &            {--}           &       {--}                &           {-- / 6.625}      &           {--}      \\
\multicolumn{1}{c|}{}                  & \multicolumn{1}{c|}{{PSM}}       &  {0.114}                     &      {0.499}                 &          {1.908}             &         {4.644}              &            {--}           &       {--}                &           {-- / 0.011}      &           {--}      \\            
\multicolumn{1}{c|}{}                  & \multicolumn{1}{c|}{{PSM+}}       &  {0.061}                     &      {0.252}                 &          {0.928}             &         {2.076}              &            {--}           &       {--}                &           {-- / 0.018}       &           {--}     \\                         
\multicolumn{1}{c|}{}                 & \multicolumn{1}{c|}{{PSM++}}       &  {0.032}                     &      {0.114}                 &          {0.378}             &         {0.820}              &            {--}           &       {--}                &           {-- / 0.039}       &           {--}     \\      
\multicolumn{1}{c|}{}                & \multicolumn{1}{c|}{{MCNP}}       &  {1.139 \footnotesize{$\pm$0.145}}                     &      {1.538 \footnotesize{$\pm$0.616}}                 &          {1.424 \footnotesize{$\pm$0.266}}             &         {0.977 \footnotesize{$\pm$0.045}}              &          {--}           &    {--}             &           {-- / 0.010}       &           {0.139} 
\\        \bottomrule
\end{tabular}}
\vspace{-2pt}
\end{table*}

\subsection{{Comparison with Numerical Methods}}

{This section presents experiments comparing the MCNP with various numerical solvers, including the traditional Monte Carlo method and Eulerian methods, such as PSM and FDM, for all equations in Section~\ref{sec:exp}. For Monte Carlo methods, we directly simulate the corresponding SDEs for convection-diffusion and fractional equations, using the branching diffusion method~\cite{henry2014numerical} for Allen-Cahn and the random vortex method~\cite{qian2022random} for Navier-Stokes. To evaluate the effect of particle count, we vary the number of particles, with more plus signs (MCM/MCM+/MCM++) indicating a higher number of particles. For Eulerian solvers, we employ FDM for Allen-Cahn equations and PSM for the other three types of PDEs. In this case, we vary the step size, with more plus signs (FDM/FDM+/FDM++ or PSM/PSM+/PSM++) representing a smaller step size. To ensure a fair comparison, we test the solvers across multiple spatial resolutions, selecting the lowest resolution that yields comparable results. The implementation details are provided in Appendix~\uppercase\expandafter{\romannumeral1}.5. Table~\ref{tab:solver} summarizes the comparison between MCNP and other numerical solvers.}

{According to the results in Table~\ref{tab:solver}, the performance and efficiency of Monte Carlo methods are usually heavily constrained by the number of particles, particularly at high diffusion rates. For instance, when $\kappa$ is large (as in E3 and E6), there is a significant error difference between MCM and MCM++ in the convection-diffusion equation. While increasing the number of particles can reduce errors, it introduces serious computational challenges, including memory and inference time. In contrast, the MCNP solver approximates the expectation in Eq.~\ref{eq:sde_g} using the PDF of neighboring grid points, avoiding the need to sample a large number of particles, as discussed in Section~\ref{sec:diffusion}.}

{In comparing MCNP with traditional Eulerian methods, we observe that conventional solvers often require refining the time step, especially in highly nonlinear cases, leading to increased computational costs. In contrast, MCNP offers significant speed advantages, particularly in solving equations like Navier-Stokes equations, achieving 10-50x speedup on GPU and 2-8x on CPU. Nevertheless, the speedup is limited in some 1D experiments, such as E1 for convection-diffusion, with MCNP sometimes underperforming on the CPU. This suggests that traditional methods can still be more efficient in some simple cases, where the computational complexity is reduced. Another interesting observation is that the GPU provides a more significant speedup for MCNP than for traditional solvers. Such greater GPU speedup is due to MCNP's ability to leverage spatiotemporal parallelism and output physical fields directly at any time step without sequential iteration. On the other hand, traditional methods often struggle to take full advantage of GPU acceleration, as they are typically designed with sequential computations in the temporal dimension.}

{Furthermore, like other neural operators, the MCNP Solver requires hyperparameter tuning and training, while traditional numerical methods do not. However, once training is complete, the MCNP Solver can perform inference on new initial value problems without additional training or tuning. This feature makes neural operators particularly promising for applications in inverse design and real-time physical simulations, such as weather forecasting and fluid control~\cite{tanyu2022deep, kurth2023fourcastnet}. In this section, we report the inference time for the MCNP Solver, while the hyperparameter tuning costs are discussed in Appendix~\uppercase\expandafter{\romannumeral4}.}

\begin{table*}[!t]
\caption{{\textbf{Compared to the FNO on 2D Navier-Stokes equation with varying $\nu$.} Relative $\ell_2$ errors ($E_{\ell_2}$), relative $\ell_{\infty}$ errors ($E_{\ell_{\infty}}$) and the training costs for baseline methods and MCNP Solver. The training time of MCNP Solver reported in this table is the average over {E3 and E4}.}}\label{tab:fno}
\vspace{-2pt}
\centering
\begin{tabular}{cccccccc}
\toprule
    & \multicolumn{2}{c}{{E3 ($\nu = 10^{-4}$)}} & \multicolumn{2}{c}{{E4 ($\nu = 10^{-5}$)}} &  \multicolumn{3}{c}{Time (H)}  \\ \cmidrule(r){2-3} \cmidrule(r){4-5} \cmidrule(r){6-8}
              {Method}  &      $E_{\ell_2}$ (\%)     &     $E_{\ell_{\infty}}$ (\%)     &     $E_{\ell_2}$ (\%)    &$E_{\ell_{\infty}}$ (\%)      &       Data   &  Train   & Total      \\ \midrule
\multicolumn{1}{c|}{FNO} &      4.754 \footnotesize{$\pm$0.139}    &     8.934 \footnotesize{$\pm$0.248}   &    8.003 \footnotesize{$\pm$0.161}      &   15.184 \footnotesize{$\pm$0.239}      &     0.346    &  0.207 &    0.553       \\
\multicolumn{1}{c|}{FNO+} &   3.460 \footnotesize{$\pm$0.163}       &   6.630 \footnotesize{$\pm$0.358}    &    6.115 \footnotesize{$\pm$0.085}    &   11.743 \footnotesize{$\pm$0.143}    &   0.692  &   0.207 &     0.899          \\
\multicolumn{1}{c|}{FNO++} &     2.619 \footnotesize{$\pm$0.124}   &    4.906 \footnotesize{$\pm$0.195}     &   4.640 \footnotesize{$\pm$0.032}   &   8.839 \footnotesize{$\pm$0.148}     &   1.384  & 0.207  &         1.591     \\
\multicolumn{1}{c|}{MCNP-10} &   3.101 \footnotesize{$\pm$0.043}   &  5.815 \footnotesize{$\pm$0.075}     &     5.825 \footnotesize{$\pm$0.068}   &  {10.452 \footnotesize{$\pm$0.122}}   &  0  &  0.261  &  0.261  \\
\multicolumn{1}{c|}{MCNP-20} &     2.999 \footnotesize{$\pm$0.082}   &   5.336 \footnotesize{$\pm$0.131}     &      5.776 \footnotesize{$\pm$0.085}    &  10.137 \footnotesize{$\pm$0.146}      &  0  &  0.511  &  0.511  \\ \bottomrule
\end{tabular}
\vspace{-5pt}
\end{table*}

\subsection{Comparison with Supervised Solver Learning}\label{sec:supervised}

In this section, we conduct experiments to compare MCNP Solver with supervised neural operator learning methods FNO~\cite{DBLP:conf/iclr/LiKALBSA21} on the Navier-Stokes equations {(E3 and E4)} in Section~\ref{sec:nse}. To evaluate the performance of FNO with the amounts of datasets, we utilize 500/1000/2000 PDE trajectories to train FNO and denote the corresponding methods as FNO/FNO+/FNO++, respectively. 

Table~\ref{tab:fno} presents a comparison between MCNP-10/20 and three versions of FNO. Compared to the supervised methods, which use pre-simulated fixed data for training, MCNP Solver can sample new initial fields per epoch, thereby increasing the diversity of training data. As a result, MCNP-10/20 can outperform FNO and FNO+. With the increase of training data, FNO++ achieves better results. This is because the training data is generated from high-precision numerical methods, while the unsupervised methods construct the loss function with low spatiotemporal resolution. However, as a trade-off, FNO++ spends 211-510\% more total time than MCNP-20/10.

\begin{table}[!t]
\caption{\textbf{Ablation Studies of each component in MCNP Solver. }Relative error (\%) and training time for each method on the Navier-Stokes equation tasks with {$\nu=10^{-4}$ (E3) and $\nu=10^{-5}$ (E4)}. The training time of each method reported in this table is the average over {E3 and E4}.}\label{tab:ab}
\centering
\vspace{-5pt}
\resizebox{8.8cm}{!}{
\begin{tabular}{cccccc}
\toprule
    & \multicolumn{2}{c}{{E3 ($\nu = 10^{-4}$)}} & \multicolumn{2}{c}{{E4 ($\nu = 10^{-5}$)}} &     \\ \cmidrule(r){2-3} \cmidrule(r){4-5}
              {Method}  &      $E_{\ell_2}$ (\%)     &     $E_{\ell_{\infty}}$ (\%)     &     $E_{\ell_2}$ (\%)      &   $E_{\ell_{\infty}}$ (\%)      &       {Time (H)}             \\ \midrule
\multicolumn{1}{c|}{MCNP-\sout{H}-10} &      11.572 \footnotesize{$\pm$0.009}     &        19.482 \footnotesize{$\pm$0.013}    &       13.374 \footnotesize{$\pm$0.055}     &        23.280 \footnotesize{$\pm$0.076}   &     0.253            \\
\multicolumn{1}{c|}{MCNP-\sout{I}-10} &   3.168 \footnotesize{$\pm$0.026}     &        5.931 \footnotesize{$\pm$0.048}       &   61.719 \footnotesize{$\pm$3.475}     &        78.465 \footnotesize{$\pm$9.320}   &       0.242        \\
\multicolumn{1}{c|}{MCNP-10} &     3.101 \footnotesize{$\pm$0.043}   &  5.815 \footnotesize{$\pm$0.075}     &     5.825 \footnotesize{$\pm$0.068}   &  {10.452 \footnotesize{$\pm$0.122}}     &     0.261     \\ \midrule    
\multicolumn{1}{c|}{MCNP-\sout{H}-20} &   7.104 \footnotesize{$\pm$0.048}   &  12.403 \footnotesize{$\pm$0.080}     &      9.249 \footnotesize{$\pm$0.041}   &  16.613 \footnotesize{$\pm$0.101} &           0.509        \\
\multicolumn{1}{c|}{MCNP-\sout{I}-20} &  8.970 \footnotesize{$\pm$0.025}     &        15.138 \footnotesize{$\pm$0.026}    &     69.067 \footnotesize{$\pm$5.975}     &    72.800 \footnotesize{$\pm$4.340}      &     0.408              \\
\multicolumn{1}{c|}{MCNP-20} &     2.999 \footnotesize{$\pm$0.082}   &   5.336 \footnotesize{$\pm$0.131}     &      5.776 \footnotesize{$\pm$0.085}    &  10.137 \footnotesize{$\pm$0.146}   &    0.511           \\ \bottomrule
\end{tabular}}
\vspace{-10pt}
\end{table}

\subsection{Ablation Studies}

In this section, we conduct several ablation studies on the MCNP Solver applied to the Navier-Stokes equation {(E3 and E4)}. Our goal is to evaluate the individual contribution of each method component. MCNP-\sout{H} replaces Heun's method (Section~\ref{sec:heun}) with the traditional Euler method when simulating the SDEs. MCNP-\sout{I} represents the MCNP Solver without the interpolation trick introduced in Section~\ref{sec:r}. 

Table~\ref{tab:ab} reports the results and training costs. Compared to MCNP Solver with MCNP-\sout{H}, the results show that using Heun's method to simulate the SDEs significantly improves the accuracy of MCNP Solver, while incurring only minimal additional computational cost, especially for MCNP-10. Compared to MCNP Solver with MCNP-\sout{I}, the interpolation trick plays a crucial role in {E4} when $\nu=10^{-5}$. As discussed in Section~\ref{sec:r}, extremely low diffusion rates can lead to very short-distance diffusion effects, resulting in $\sum_i p_t(\boldsymbol{\xi}_{p,t}^{d, i}) \delta \ll 1$. Therefore, interpolating the original vorticity field becomes essential to ensure that the PDF of grid points satisfies normalization conditions in Eq.~\ref{eq:r}. Similarly, the interpolation trick can reduce more relative error for MCNP-20 than MCNP-10 in {E3} because fine step size can also introduce localized random walks per step.

\section{Conclusion and Discussion}\label{sec:future}

In this paper, we propose the MCNP Solver, which leverages the Feynman-Kac formula to train neural PDE solvers in an unsupervised manner. {Compared to other unsupervised neural PDE solvers, the MCNP Solver can be more robust to complex spatiotemporal variations due to the advantages of Lagrangian methods. Moreover, the experiments on 2D fractional diffusion equations demonstrate the applicability of the MCNP Solver in mesh-free scenarios and its capability to handle fractional order Laplacian operators.}

This paper has several limitations: (1) Some PDEs are not suitable for the Feynman-Kac formula and therefore do not fall within the scope of the MCNP Solver, such as third or higher-order PDEs (involving high-order operators like $u_{xxx}$). (2) The accuracy of the MCNP Solver cannot outperform traditional numerical solvers when disregarding inference time, which is also a major drawback for other existing neural solvers~\cite{tanyu2022deep,grossmann2023physicsinformed}, as we discussed in Section~\ref{sec:diffusion}. {(3) Like other neural operators, the MCNP Solver requires hyperparameter tuning and training, while traditional numerical methods do not. However, once trained, neural operators can be applied directly to new initial value problems for rapid inference.}

Furthermore, we suggest several directions for future research: (1) Extend the proposed MCNP Solver to broader scenarios, such as high-dimensional PDEs and optimal control problems; (2) Utilize techniques from out-of-distribution generalization~\cite{shen2021towards} to improve the generalization ability of MCNP Solver; {(3) There are some mathematical works have extended the probabilistic representation of PDEs to the higher-order cases~\cite{allouba2001brownian, orsingher2015higher}, and extending the MCNP Solver to such scenario is also a feasible and promising direction.}

\section*{Acknowledgments}
This work was in part supported by NSFC [No. 9247010235], National Key Research and Development Program of China [2019YFA0709\linebreak501] and CAS Project for Young Scientists in Basic Research [No. YSBR-034].

\section*{Appendix ~\uppercase\expandafter{\romannumeral1}: Implementation Details}
\subsection*{\uppercase\expandafter{\romannumeral1}.1: Baselines}\label{app:baselines}
In this paper, we adopt Pytorch~\cite{paszke2019pytorch} to implement MCNP Solver, FNO, SP-FNO, and PINO, and JAX~\cite{jax2018github} for PI-DeepONet-(M), respectively. Here, we introduce PINO and PI-DeepONet as follows.
\subsubsection*{PI-DeepONet~\cite{wang2021learning}} 
PI-DeepONet utilizes the PDE residuals to train DeepONets in an unsupervised way. The loss function in PI-DeepONet can be formulated as follows:
\begin{equation}\label{eq:pi-deeponet}
\begin{aligned}
        \mathcal{L}_{\operatorname{PI-DeepONet}} &= \lambda_1  \mathcal{L}_{\operatorname{initial}} + \lambda_2 \mathcal{L}_{\operatorname{boundary}} + \mathcal{L}_{\operatorname{physics}},\\
        \text{where} \quad \mathcal{L}_{\operatorname{initial}} &= \operatorname{RMSE}[\mathcal{G}_{\theta}(u^b_0,t=0)(\boldsymbol{x}_p)-u_0^b(\boldsymbol{x}_p)],\\
        \mathcal{L}_{\operatorname{boundary}} &=\operatorname{RMSE}[\mathcal{B}(\mathcal{G}_{\theta}, \boldsymbol{x}, t)],\\
        \mathcal{L}_{\operatorname{physics}} &= \operatorname{RMSE}[\mathcal{R}(\mathcal{G}_{\theta}(u^b_0,t)(\boldsymbol{x}_p), \boldsymbol{x}_p,t)],
\end{aligned}
\end{equation}
where $\operatorname{RMSE}$ represents the rooted mean square error, $\mathcal{G}_{\theta}$ represents a neural operator, $\mathcal{G}$ and $\mathcal{R}$ denote the ground truth and the residual of the PDE operator, respectively. As shown in Eq.~\ref{eq:pi-deeponet}, $\mathcal{L}_{\operatorname{inital}}$, $\mathcal{L}_{\operatorname{boundary}}$ and $\mathcal{L}_{\operatorname{physics}}$ enforce
$\mathcal{G}_{\theta}$ to satisfy the initial conditions, boundary conditions and the PDE constraints, respectively. Like PINNs~\cite{raissi2019physics}, the PDE residuals in Eq.~\ref{eq:pi-deeponet} are calculated via the auto-differentiation.

\subsubsection*{PINO~\cite{li2021physics}} PINO utilizes the pseudo-spectral or finite difference methods to construct the loss function between $\mathcal{G}_{\theta}(u^b_t)$ and $\mathcal{G}_{\theta}(u^b_{t+\Delta t})$. PINO utilized the FNO~\cite{li2021neural} as the backbone network. The loss function in PINO can be formulated as follows:
\begin{equation}\label{eq:pino}
\begin{aligned}
        \mathcal{L}_{\operatorname{PINO}} &= \lambda \mathcal{L}_{\operatorname{boundary}} + \mathcal{L}_{\operatorname{physics}},\\
        \text{where} \quad \mathcal{L}_{\operatorname{boundary}} &= \operatorname{RMSE}[\mathcal{B}(\mathcal{G}_{\theta}, \boldsymbol{x}, t)],\\
        \mathcal{L}_{\operatorname{physics}} &=\sum_{t=0}^{T-\Delta t} \operatorname{RMSE}[\mathcal{G}_{\theta}(u^b_0,t+\Delta t)(\boldsymbol{x}_p)  - \mathcal{P}(\mathcal{G}_{\theta},\boldsymbol{x}_p,t)],
\end{aligned}
\end{equation}
where $\mathcal{B}$ denotes the constraints on the boundary conditions, and $\mathcal{P}$ denotes the update regime of numerical PDE solvers. 

\subsection*{\uppercase\expandafter{\romannumeral1}.2: 1D Convection-Diffusion Equation}\label{app:heat}
\subsubsection*{Data}
The initial states $u(x, 0)$ are generated from the functional space $\mathcal{F}_N \triangleq \{\sum_{n=1}^N a_n \sin(2\pi n x):a_n \sim \mathbb{U}(0, 1)\},$
where $\mathbb{U}(0, 1)$ denotes the uniform distribution over $(0, 1)$, and $N$ represents the maximum frequency of the functional space. We generate the ground truth with the pseudo-spectral methods with the Crank–Nicolson regime. All PDE instances are generated on the spatial grid $1024$, then down-sampled to $64$. The step size is fixed as $10^{-5}$. We generate 200 test data with seed 1 and 200 validation data with seed 2.

\subsubsection*{Hyperparameters} 
The PINO and MCNP Solver use the 1D FNO as the backbone models. We fix the number of layers and widths as 4 and 32, respectively. We choose the best $modes$ (the number of frequency components) in $\{12, 16, 20\}$ for FNO, respectively. For PINO and MCNP Solver, we utilize Adam to optimize the neural network for 10000 epochs with an initial learning rate $lr$ and decay the learning rate by a factor of 0.5 every 1000 epochs. The batch size is fixed as 200. The learning rate is chosen from the set $\{0.02, 0.01, 0.005, 0.001\}$. Because the pseudo-spectral methods can naturally stratify the boundary conditions, we fix $\lambda$ as $0$ in this section. For PI-DeepONet, we choose the network structure in line with the 1D case in~\cite{wang2021learning} and extend the training iterations to 50000 to ensure the convergence of the model. Moreover, we search the hyperparameters $\lambda_1$, $\lambda_2$ and the width of the neural networks in $\{1, 5, 10\}$, $\{1, 5, 10, 25, 50\}$ and $\{80,100, 120 \}$, respectively. {Because PI-DeepONet-M utilizes an adaptive re-weighting scheme to balance training loss, we do not need to tune the hyper-parameters in PI-DeepONet.} All hyperparameters are chosen via the validation set with seed 0.

\subsection*{\uppercase\expandafter{\romannumeral1}.3: 1D Allen-Cahn Equation}
\subsubsection*{Data}
The initial states $u(x, 0)$ are generated from the functional space $\mathcal{F}_N \triangleq \{\sum_{n=1}^N a_n \sin(2\pi n x):a_n \sim \mathbb{U}(0, 1)\},$
where $\mathbb{U}(0, 1)$ denotes the uniform distribution over $(0, 1)$, and $N$ represents the maximum frequency of the functional space. We generate the ground truth with the Python package `py-pde'~\cite{py-pde}. The ground truth solution is generated via the finite-difference method with the Runge-Kutta 4 method. All PDE instances are generated on the spatial grid $1024$, then down-sampled to $65$. The step size is fixed as $10^{-6}$. We generate 200 test data with seed 1 and 200 validation data with seed 2.

\subsubsection*{Hyperparameters} 
The PINO and MCNP Solver use the 1D FNO as the backbone models. We fix the number of layers and widths as 4 and 32, respectively. We choose the best $modes$ in $\{12, 16, 20\}$ for FNO, respectively. For PINO and MCNP Solver, we utilize Adam to optimize the neural network for 20000 epochs with an initial learning rate $lr$ and decay the learning rate by a factor of 0.5 every 2000 epochs. The batch size is fixed as 200. The learning rate is chosen from the set $\{0.02, 0.01, 0.005, 0.001\}$. The hyperparameter $\lambda$ in the PINO loss is chosen from the set $\{1, 2, 5, 10, 25, 50\}$. For PI-DeepONet, we choose the network structure in line with the 1D case in~\cite{wang2021learning} and extend the training iterations to 200000 to ensure the convergence of the model. Moreover, we search the hyperparameters $\lambda_1$, $\lambda_2$ and the width of the neural networks in $\{1, 5, 10\}$, $\{1, 5, 10, 25, 50\}$ and $\{80,100, 120 \}$, respectively. {Because PI-DeepONet-M utilizes an adaptive re-weighting scheme to balance training loss, we do not need to tune the hyper-parameters in PI-DeepONet.} All hyperparameters are chosen via the validation set with seed 0.

\subsection*{\uppercase\expandafter{\romannumeral1}.4: 2D Navier-Stokes Equation}
\subsubsection*{Data} 
We utilize the pseudo-spectral methods to generate the ground truth test data with the step size of $10^{-4}$ for the Crank–Nicolson scheme. Furthermore, all PDE instances are generated on the grid $256 \times 256$, then down-sampled to $64 \times 64$, which is in line with the setting in~\cite{li2021neural}. We generate 200 test data with seed 0, 200 validation data with seed 1 and 2000 training data with seed 2.

\subsubsection*{Hyperparameters} 
The PINO, SP-FNO and MCNP Solver use the 2D FNO as the backbone models. We fix the number of layers and widths as 4 and 36, respectively. We choose the best $modes$ in $\{12, 16, 20\}$ for FNO, respectively. {For PINO, SP-FNO and MCNP Solver, we utilize Adam to optimize the neural network for 20000 epochs with an initial learning rate of $lr$ and decay the learning rate by a factor of 0.8 every 2000 epochs.} The batch size is fixed as 10. The learning rate $lr$ is chosen from the set $\{0.02, 0.01, 0.005, 0.001\}$. Because the pseudo-spectral methods can naturally stratify the boundary conditions, we fix the $\lambda$ as $0$ in this section. For FNO, we find that a cosine annealing schedule can obtain the best result when training with the supervised regime. We utilize Adam to optimize the neural network for 400/200/100 epochs with the initial learning rate of $lr$ for FNO/FNO+/FNO++, respectively. The learning rate $lr$ is chosen from the set $\{0.02, 0.01, 0.005, 0.001\}$. All hyperparameters are chosen via the validation set with seed 0.

\subsection*{\uppercase\expandafter{\romannumeral1}.4: 2D Fractional Diffusion Equation on a Disk}
\subsubsection*{Data} 
{The test data is generated from a fine-grained pseudo-spectral method, using the Bessel function as the basis function. The radius $[0, 1]$ is divided into 400 grid points, with a time step of $10^{-4}$. Afterward, we downsample the spatiotemporal resolution to $40 \times 10$. Fifty test datasets are generated using seed 1, and fifty validation datasets are generated using seed 2.}

\subsubsection*{Hyperparameters} 
We utilize the DeepONet as a backbone model and train the MCNP Solver in a mesh-free regime. The batch size is fixed as 20. For each training epoch, we uniformly sample 2000 spatiotemporal points in the target domain and construct the loss function accordingly. We fix the number of layers and widths as 5 and 200, respectively. We use ReLU as an activation function. We utilize Adam to optimize the neural network for 10000 epochs with an initial learning rate of $lr$ and decay the learning rate by 0.8 every 2000 epochs. The learning rate $lr$ is chosen from the set $\{0.02, 0.01, 0.005, 0.001\}$. All hyperparameters are chosen via the validation set with seed 0.

{\subsection*{\uppercase\expandafter{\romannumeral1}.5: Implementation Details of Numerical Solvers}
In this section, we introduce the implementation details of traditional numerical solvers, including both Eulerian and Monte Carlo methods. To ensure a fair comparison, we tested the solvers across multiple spatial resolutions, selecting the lowest resolution that produced comparable results. The implementation of traditional Eulerian methods follows the same procedure as the data generation process; however, to accelerate inference, we employed a coarser spatiotemporal grid. The specific spatiotemporal resolutions used are provided in Table~\ref{tab:solver_}.
}

\begin{table*}[!t]
\caption{{\textbf{Spatiotemporal resolutions and number of particles used for testing numerical solvers.}}}\label{tab:solver_}
\vspace{-5pt}
\centering
\resizebox{15.8cm}{!}{
\begin{tabular}{ccccc}
\toprule
\multicolumn{1}{c|}{{Equation}}          & \multicolumn{1}{c|}{{Solvers}} &    {Temporal Resolution}                   &      {Spatial Resolution}                    &  {Number of Particles}                        \\ \midrule
\multicolumn{1}{c|}{\multirow{2}{*}{{\shortstack{Convection-\\Diffusion}}}} & \multicolumn{1}{c|}{{MCM / MCM+ / MCM++}}       &  {10}                     &      {64}                   &         {200 / 2000 / 20000}              \\
\multicolumn{1}{c|}{}                  & \multicolumn{1}{c|}{{PSM / PSM+ / PSM++}}       &  {10 / 20 / 50}                     &      {16}                      &         {--}            \\       \midrule     

\multicolumn{1}{c|}{\multirow{2}{*}{{\shortstack{Allen-\\Cahn}}}} &  \multicolumn{1}{c|}{{MCM / MCM+ / MCM++}}         &  {20}                     &      {64}             &         {200 / 2000 / 20000}             \\
\multicolumn{1}{c|}{}                  & \multicolumn{1}{c|}{{FDM / FDM+ / FDM++}}      &  {20 / 100 / 200}                     &      {64}                       &         {--}            \\      \midrule      

\multicolumn{1}{c|}{\multirow{2}{*}{{\shortstack{Navier-\\Stokes}}}} &  \multicolumn{1}{c|}{{MCM / MCM+ / MCM++}}       &  {10}                     &      { $64 \times 64$}            &         {10 / 20 / 50}             \\
\multicolumn{1}{c|}{}                  & \multicolumn{1}{c|}{{PSM / PSM+ / PSM++}}      &  {100 / 200 / 500}                     &      { $16\times 16$}             &         {--}            \\      \midrule     

\multicolumn{1}{c|}{\multirow{2}{*}{{\shortstack{Fractional}}}} &  \multicolumn{1}{c|}{{MCM / MCM+ / MCM++}}        &  {20}                     &      {40}                  &         {200 / 2000 / 20000}             \\
\multicolumn{1}{c|}{}                  &  \multicolumn{1}{c|}{{PSM / PSM+ / PSM++}}        &  {10 / 20 / 50}                     &      {20}                  &         {--}            \\          \bottomrule
\end{tabular}}
\vspace{-2pt}
\end{table*}

\section*{{Appendix ~\uppercase\expandafter{\romannumeral2}: The Validation Loss During Training}}\label{app:loss}

\begin{figure*}[!t]
\begin{center}
\centerline{\includegraphics[width=19cm]{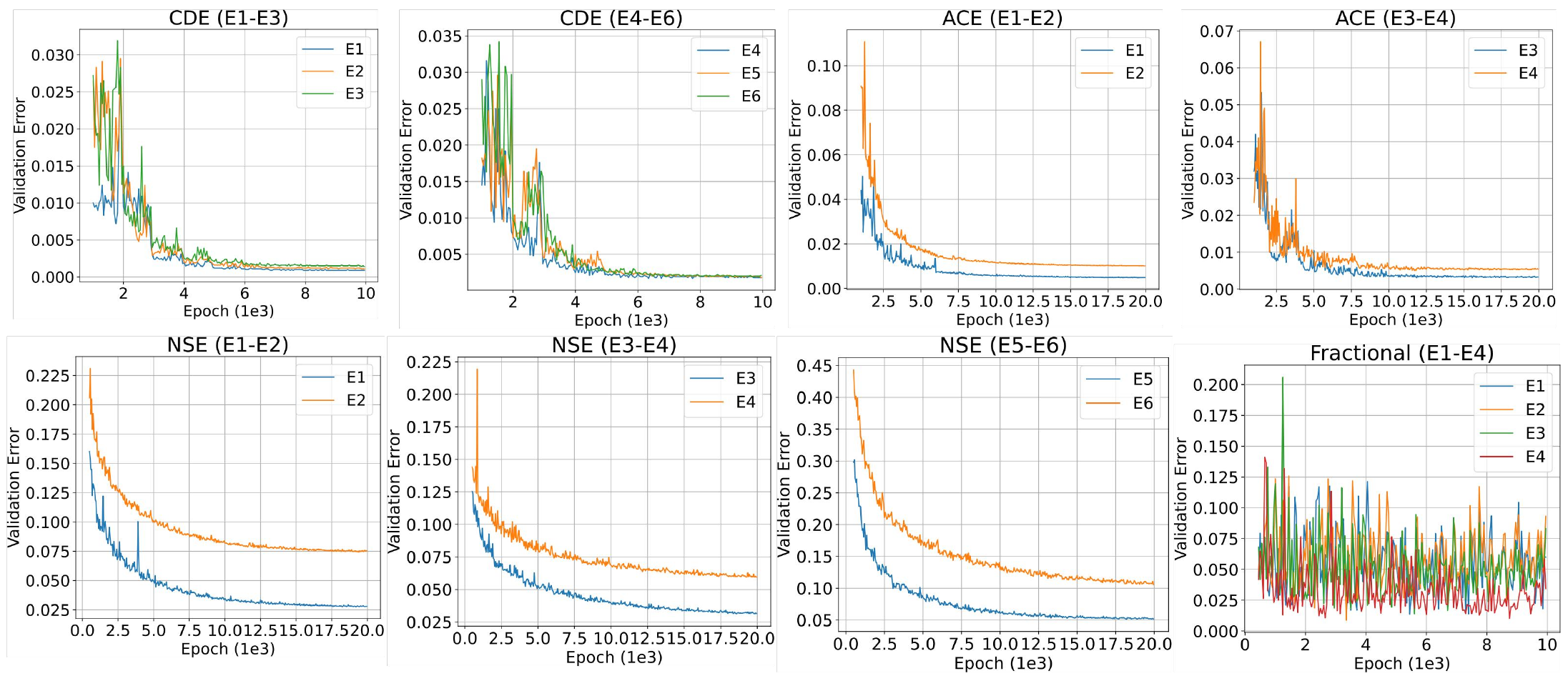}}
\caption{{{\textbf{The validation loss of MCNP Solver during training.} CDE, ACE, and NSE are the abbreviations of the convection-diffusion equation, Allen-Cahn equation and Navier-Stokes equation, respectively. We use the results of MCNP-10, MCNP-100 and MCNP-20 to plot the training curves for CDE, ACE, and NSE, respectively.}}}
\label{loss}
\end{center}
\end{figure*}

{In this section, we present the validation loss of all experiments during training process in Fig.~\ref{loss}.}

\section*{{Appendix ~\uppercase\expandafter{\romannumeral3}: Choice of the Backbone Network}}\label{app:backbone}
{In this section, we discuss the choice of the backbone network of MCNP Solver. Firstly, we test three network structures on the 2D Navier-Stokes equation ($\nu=10^{-4}$ with Li forcing), including FNO~\cite{li2021neural}, UNet~\cite{ronneberger2015u} and MultiWaveleT- (MWT) based model~\cite{gupta2021multiwavelet}. We divide the time interval $[0, 10]$ into ten uniform lattices for each method. Table~\ref{tab:backbone} presents each backbone method's performance and computational cost on the 2D Navier-Stokes equations. In this paper, our primary contribution lies in the design of the loss function for unsupervised neural PDE-solving training. Consequently, we choose one of the most widely used backbone models, i.e., FNO, for our main experiments.}

\begin{table}
\centering
\caption{{\textbf{Choice of the backbone network.} Relative $\ell_2$ errors ($E_{\ell_2}$), relative $\ell_{\infty}$ errors ($E_{\ell_{\infty}}$) and computational costs for varying backbone models.}}
\label{tab:backbone}
\begin{tabular}{ccccc}
\toprule
{Model}& {$E_{\ell_2}$ (\%)} & {$E_{\ell_{\infty}}$ (\%)} & {Train Time (H)} & {Param.(M)}  \\ \midrule
{FNO} & {3.101 \footnotesize{$\pm$0.043}}    &  {5.815 \footnotesize{$\pm$0.075}} & {0.261}   &     {5.319} \\
{U-Net}&  {6.030 \footnotesize{$\pm$0.138}}    &  {11.107 \footnotesize{$\pm$0.189}} & {0.287} &  {6.823} \\
 {MWT}  & {3.929 \footnotesize{$\pm$0.057}}    &  {7.367 \footnotesize{$\pm$0.086}} & {0.621} & {6.361} \\
         \bottomrule
\end{tabular}
\end{table}

\section*{{Appendix ~\uppercase\expandafter{\romannumeral4}: Discussions on Hyperparameter Tuning}}\label{app:param}
{Compared to traditional solvers, the MCNP Solver and other neural operators require an optimization process that includes neural network training and hyperparameter tuning, while traditional numerical methods do not. We explain this issue from two perspectives. Firstly, neural operators are trained by sampling various initial fields as input, enabling them to generalize to new initial fields and predict evolving dynamics without retraining or fine-tuning after the training phase. Since neural networks can roll out quickly, neural operators can significantly accelerate the simulation process for new initial value problems. This makes them promising for inverse design and real-time physical simulations, such as weather forecasting and fluid control~\cite{tanyu2022deep, kurth2023fourcastnet}. Secondly, MCNP Solver is more efficient in hyperparameter tuning compared to other physics-driven neural operators~\cite{wang2021learning, li2021physics}. This efficiency stems from MCNP’s simple loss function, involving only Monte Carlo loss, with boundary and initial conditions naturally embedded in the random walks of the particles. For example, in models like PI-DeepONet~\cite{wang2021learning}, the loss includes physical, initial, and boundary terms, making hyperparameter tuning complicated. In contrast, for the MCNP Solver, the hyperparameter tuning only involves adjusting the optimizer’s learning rate and the neural operator’s $modes$ (the number of frequency components). Fig.~\ref{fig:param} shows the performance of the Navier-Stokes equation (E3) under seed 0 with varying initial learning rates and modes, and the results show that the performance of MCNP solver with varying learning rates and modes are relatively robust, indicating easier parameter tuning.}

\pgfplotsset{width=8cm, height=6cm}
\begin{figure}[!tb]
\centering
\resizebox{7cm}{!}{
\begin{tikzpicture}
	\begin{groupplot}[group style={group size=2 by 2, horizontal sep = 50pt, vertical sep = 60pt},width=6cm, height=5cm]
    \nextgroupplot[title=(a) The Effects of Learning Rates, 
    x tick label style={/pgf/number format/1000 sep={}},
    legend pos=south east,
    ymajorgrids=true,
    symbolic x coords={0.001, 0.005, 0.01, 0.02},
    xtick=data,
    grid style=dashed,
    axis lines = left,
    xlabel=Learning Rates, 
    ylabel=$E_{\ell_2}$ (\%), 
    tick align=outside, 
    ymin=0, ymax=10]

    \addplot[blue,very thick,mark=square] plot coordinates { 
    (0.001, 5.198) 
    (0.005, 3.145) 
    (0.01, 3.083)
    (0.02, 8.127)
    };
    \nextgroupplot[title=(b) The Effects of $Modes$, 
    x tick label style={/pgf/number format/1000 sep={}},
    legend pos=south east,
    ymajorgrids=true,
    symbolic x coords={12, 16, 20},
    xtick=data,
    grid style=dashed,
    axis lines = left,
    xlabel=$Modes$, 
    ylabel=$E_{\ell_2}$ (\%), 
    tick align=outside, 
    ymin=0, ymax=6]

    \addplot[blue,very thick,mark=square] plot coordinates { 
    (12,3.243) 
    (16,3.083) 
    (20, 3.020)
    };
	\end{groupplot}

\end{tikzpicture}}
\caption{{\textbf{The effects of hyperparameter tuning.} Relative $\ell_2$ errors ($E_{\ell_2}$) for varying learning rates (a) and $modes$ (b) on Navier-Stokes experiments (E3, seed 0).} 
}
\label{fig:param}
\end{figure}

\section*{Appendix ~\uppercase\expandafter{\romannumeral5}: Additional Simulation Results}
\begin{figure*}[!t]
\begin{center}
\centerline{\includegraphics[width=18cm]{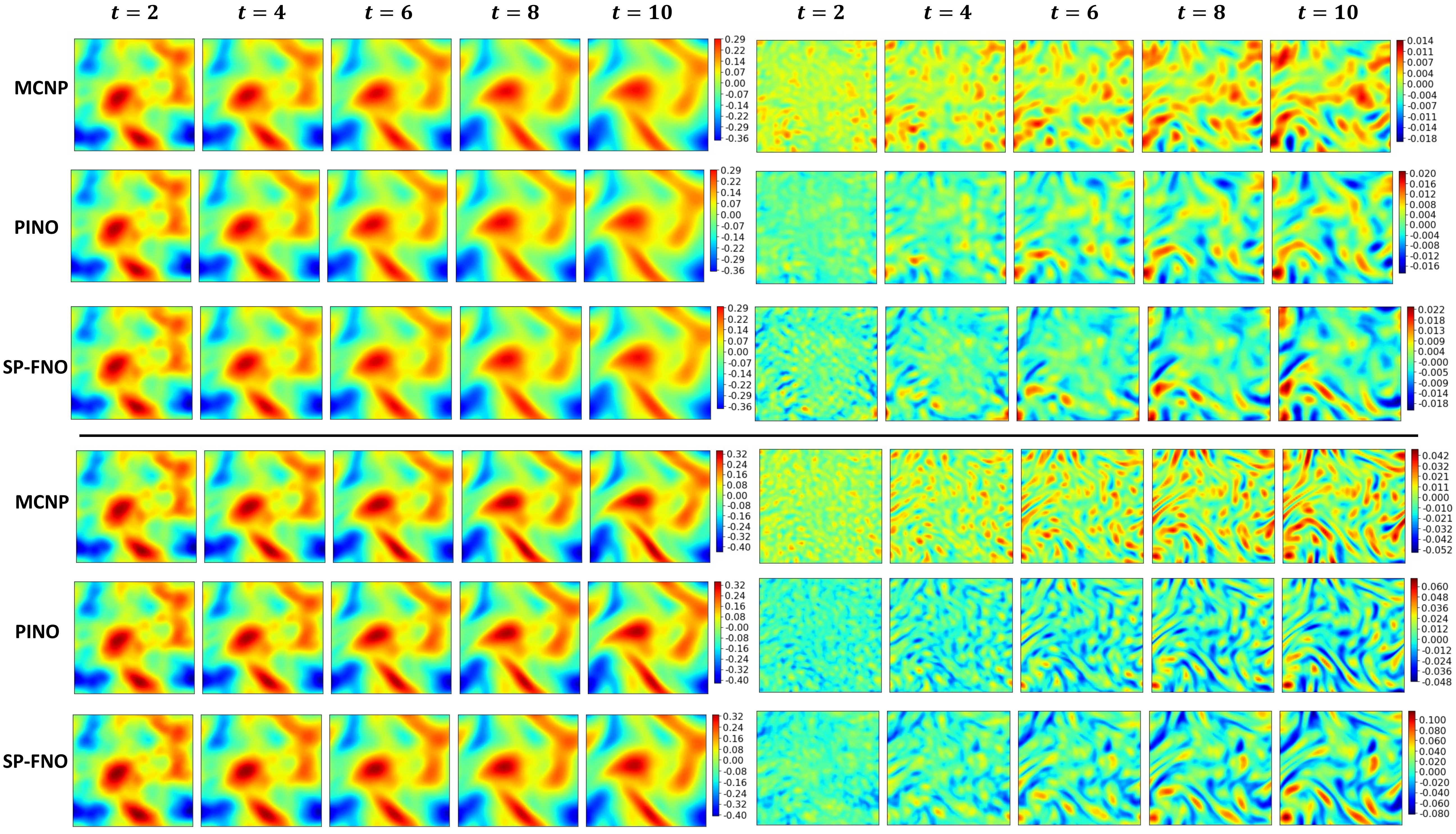}}
\caption{{\textbf{Simulation of 2D Navier-Stokes equations.} The ground truth vorticity versus the prediction of learned neural PDE solvers for an example in the test set from $t=2$ to $t=10$, with Zero forcing. (Above: $\nu=10^{-4}$; Below: $\nu=10^{-5}$).}}
\label{ns1}
\end{center}
\end{figure*}

\begin{figure*}[!t]
\begin{center}
\centerline{\includegraphics[width=18cm]{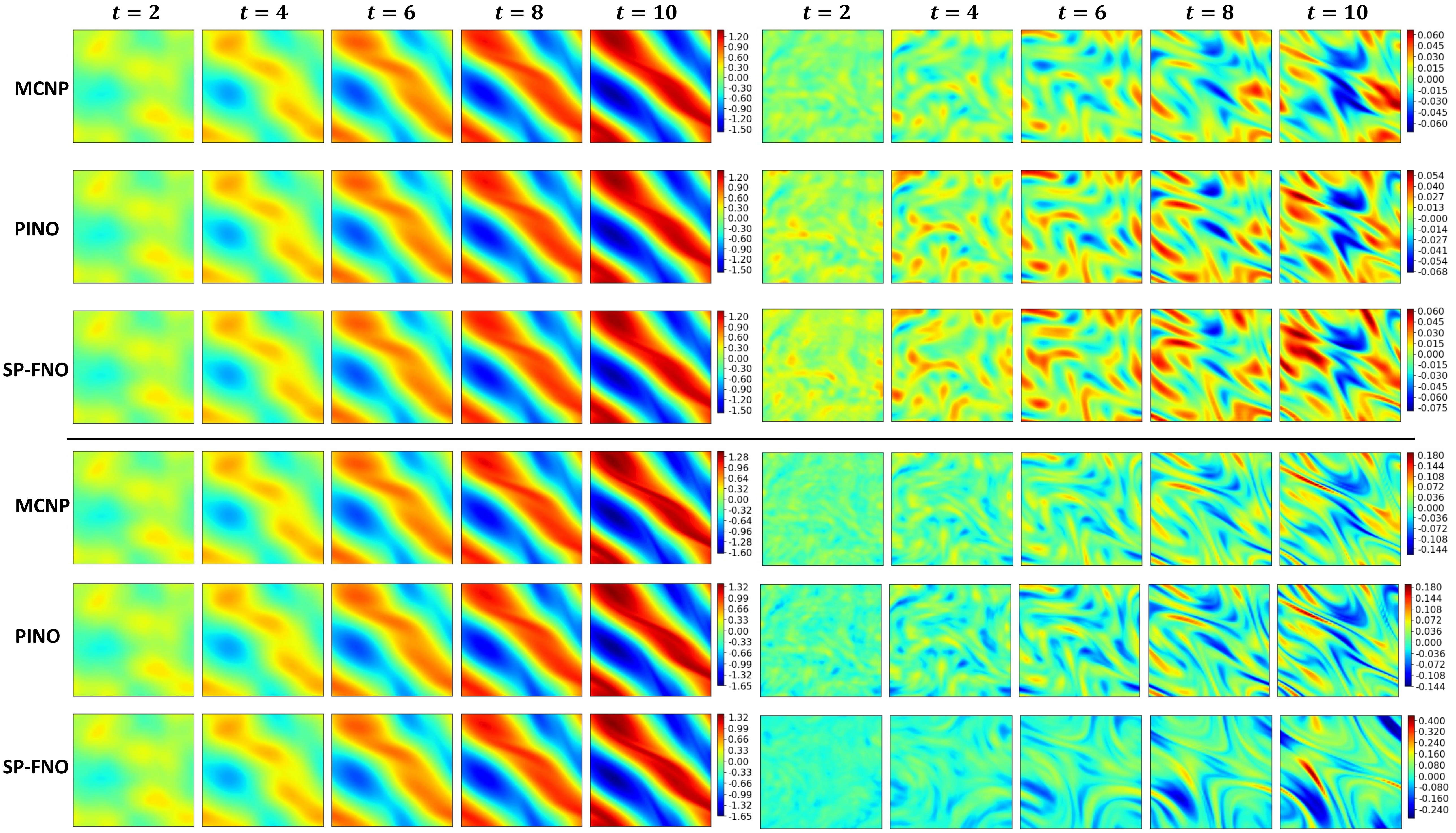}}
\caption{{\textbf{Simulation of 2D Navier-Stokes equations.} The ground truth vorticity versus the prediction of learned neural PDE solvers for an example in the test set from $t=2$ to $t=10$, with Li forcing. (Above: $\nu=10^{-4}$; Below: $\nu=10^{-5}$).}}
\label{ns2}
\end{center}
\end{figure*}

\begin{figure*}[!t]
\begin{center}
\centerline{\includegraphics[width=18cm]{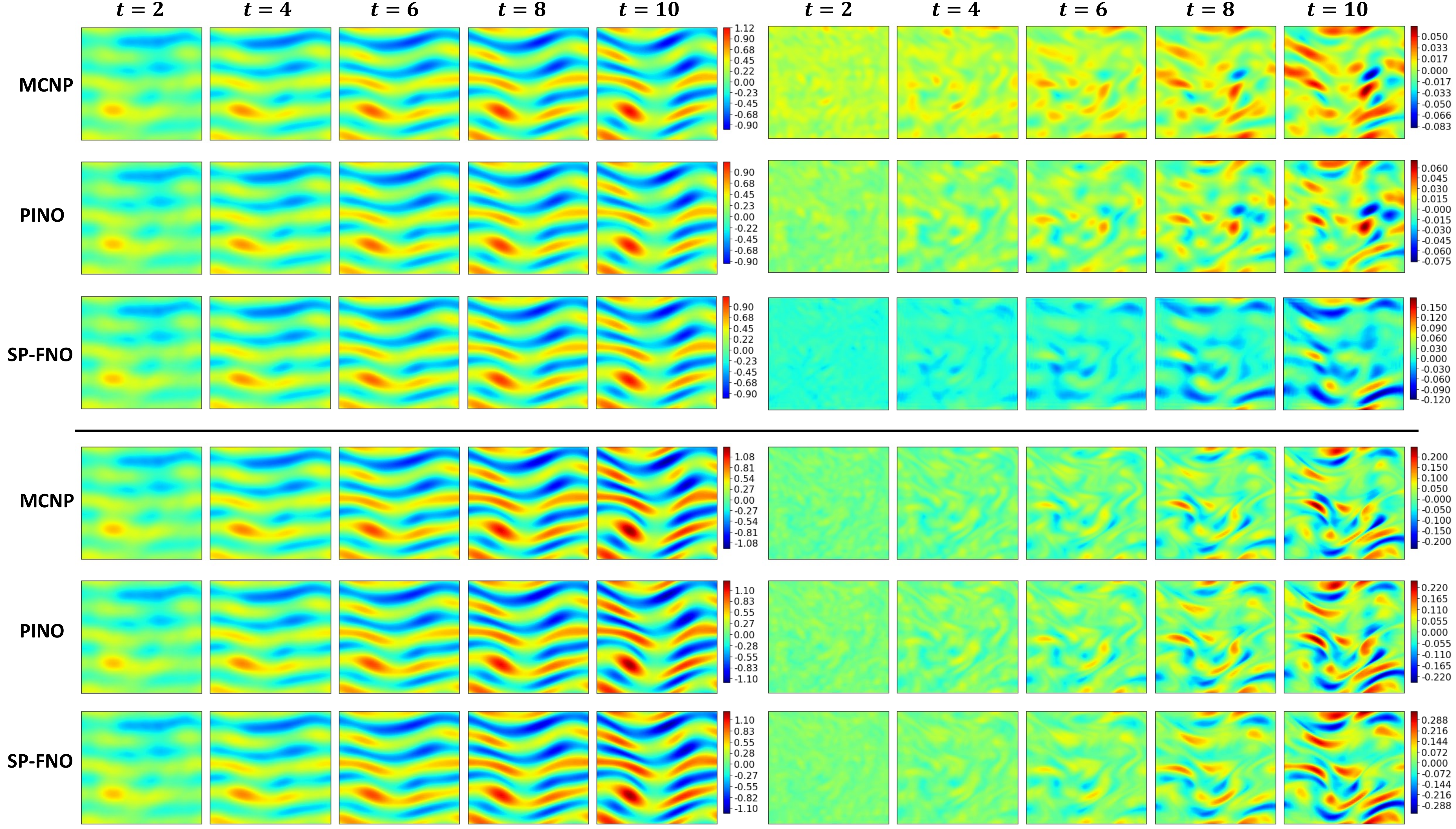}}
\caption{{\textbf{Simulation of 2D Navier-Stokes equations.} The ground truth vorticity versus the prediction of learned neural PDE solvers for an example in the test set from $t=2$ to $t=10$, with Kolmogorov forcing. (Above: $\nu=10^{-4}$; Below: $\nu=10^{-5}$).}}
\label{ns3}
\end{center}
\end{figure*}

In this section, we present a comparison between the ground truth and the predicted vorticity fields $\omega$ of all unsupervised methods from $t = 2$ to $t=10$ with Zero, Li and Kolmogorov forcing, respectively (Fig.~\ref{ns1}, Fig.~\ref{ns2} and \ref{ns3}).

\bibliographystyle{plainnat}
\bibliography{main}

\begin{thebibliography}{102}
\providecommand{\natexlab}[1]{#1}
\providecommand{\url}[1]{\texttt{#1}}
\expandafter\ifx\csname urlstyle\endcsname\relax
  \providecommand{\doi}[1]{doi: #1}\else
  \providecommand{\doi}{doi: \begingroup \urlstyle{rm}\Url}\fi

\bibitem[Allouba and Zheng(2001)]{allouba2001brownian}
Hassan Allouba and Weian Zheng.
\newblock Brownian-time processes: the pde connection and the half-derivative
  generator.
\newblock \emph{Annals of Probability}, pages 1780--1795, 2001.

\bibitem[Anderson and Wendt(1995)]{anderson1995computational}
John~David Anderson and John Wendt.
\newblock \emph{Computational fluid dynamics}, volume 206.
\newblock Springer, 1995.

\bibitem[Anitescu et~al.(2019)Anitescu, Atroshchenko, Alajlan, and
  Rabczuk]{anitescu2019artificial}
Cosmin Anitescu, Elena Atroshchenko, Naif Alajlan, and Timon Rabczuk.
\newblock Artificial neural network methods for the solution of second order
  boundary value problems.
\newblock \emph{Computers, Materials \& Continua}, 59\penalty0 (1):\penalty0
  345--359, 2019.

\bibitem[Bauer(1958)]{10.2307/2098715}
W.~F. Bauer.
\newblock The monte carlo method.
\newblock \emph{Journal of the Society for Industrial and Applied Mathematics},
  6\penalty0 (4):\penalty0 438--451, 1958.
\newblock ISSN 03684245.

\bibitem[Bezgin et~al.(2021)Bezgin, Schmidt, and Adams]{bezgin2021data}
Deniz~A Bezgin, Steffen~J Schmidt, and Nikolaus~A Adams.
\newblock A data-driven physics-informed finite-volume scheme for nonclassical
  undercompressive shocks.
\newblock \emph{Journal of Computational Physics}, 437:\penalty0 110324, 2021.

\bibitem[Bradbury et~al.(2018)Bradbury, Frostig, Hawkins, Johnson, Leary,
  Maclaurin, Necula, Paszke, Vander{P}las, Wanderman-{M}ilne, and
  Zhang]{jax2018github}
James Bradbury, Roy Frostig, Peter Hawkins, Matthew~James Johnson, Chris Leary,
  Dougal Maclaurin, George Necula, Adam Paszke, Jake Vander{P}las, Skye
  Wanderman-{M}ilne, and Qiao Zhang.
\newblock {JAX}: composable transformations of {P}ython+{N}um{P}y programs,
  2018.
\newblock URL \url{http://github.com/google/jax}.

\bibitem[Brandstetter et~al.(2022)Brandstetter, Worrall, and
  Welling]{brandstetter2022message}
Johannes Brandstetter, Daniel~E. Worrall, and Max Welling.
\newblock Message passing neural {PDE} solvers.
\newblock In \emph{International Conference on Learning Representations}, pages
  1--10, 2022.

\bibitem[Bu and Karpatne(2021)]{bu2021quadratic}
Jie Bu and Anuj Karpatne.
\newblock Quadratic residual networks: A new class of neural networks for
  solving forward and inverse problems in physics involving pdes.
\newblock In \emph{Proceedings of the SIAM International Conference on Data
  Mining}, pages 675--683. SIAM, 2021.

\bibitem[Cao(2021)]{cao2021choose}
Shuhao Cao.
\newblock Choose a transformer: Fourier or galerkin.
\newblock In \emph{Advances in Neural Information Processing Systems}, pages
  24924--24940, 2021.

\bibitem[Chen et~al.(2021)Chen, Liu, and Sun]{chen2021physics}
Zhao Chen, Yang Liu, and Hao Sun.
\newblock Physics-informed learning of governing equations from scarce data.
\newblock \emph{Nature Communications}, 12\penalty0 (1):\penalty0 1--13, 2021.

\bibitem[Chorin and Hald(2009)]{chorin2009stochastic}
Alexandre~Joel Chorin and Ole~H Hald.
\newblock \emph{Stochastic tools in mathematics and science}, volume~1.
\newblock Springer, 2009.

\bibitem[Cottet et~al.(2000)Cottet, Koumoutsakos, et~al.]{cottet2000vortex}
Georges-Henri Cottet, Petros~D Koumoutsakos, et~al.
\newblock \emph{Vortex methods: theory and practice}, volume~8.
\newblock Cambridge University Press, 2000.

\bibitem[Courant et~al.(1967)Courant, Friedrichs, and Lewy]{courant1967partial}
Richard Courant, Kurt Friedrichs, and Hans Lewy.
\newblock On the partial difference equations of mathematical physics.
\newblock \emph{IBM Journal of Research and Development}, 11\penalty0
  (2):\penalty0 215--234, 1967.

\bibitem[Dalang et~al.(2008)Dalang, Mueller, and Tribe]{dalang2008feynman}
Robert Dalang, Carl Mueller, and Roger Tribe.
\newblock A feynman-kac-type formula for the deterministic and stochastic wave
  equations and other pde’s.
\newblock \emph{Transactions of the American Mathematical Society},
  360\penalty0 (9):\penalty0 4681--4703, 2008.

\bibitem[Folino et~al.(2020)Folino, Hern{\'a}ndez~Melo, Lopez~Rios, and
  Plaza]{folino2020exponentially}
Raffaele Folino, C{\'e}sar~A Hern{\'a}ndez~Melo, Luis Lopez~Rios, and
  Ram{\'o}n~G Plaza.
\newblock Exponentially slow motion of interface layers for the one-dimensional
  allen--cahn equation with nonlinear phase-dependent diffusivity.
\newblock \emph{Zeitschrift F{\"u}r Angewandte Mathematik und Physik},
  71\penalty0 (4):\penalty0 132, 2020.

\bibitem[Fresca et~al.(2021)Fresca, Dede, and Manzoni]{fresca2021comprehensive}
Stefania Fresca, Luca Dede, and Andrea Manzoni.
\newblock A comprehensive deep learning-based approach to reduced order
  modeling of nonlinear time-dependent parametrized pdes.
\newblock \emph{Journal of Scientific Computing}, 87:\penalty0 1--36, 2021.

\bibitem[Giraldo and Neta(1997)]{giraldo1997comparison}
Francis~X Giraldo and Beny Neta.
\newblock A comparison of a family of eulerian and semi-lagrangian finite
  element methods for the advection-diffusion equation.
\newblock \emph{WIT Transactions on The Built Environment}, 30:\penalty0
  217--229, 1997.

\bibitem[Gnanasambandam et~al.(2023)Gnanasambandam, Shen, Chung, Yue, and
  Kong]{gnanasambandam2023self}
Raghav Gnanasambandam, Bo~Shen, Jihoon Chung, Xubo Yue, and Zhenyu Kong.
\newblock Self-scalable tanh (stan): Multi-scale solutions for physics-informed
  neural networks.
\newblock \emph{IEEE Transactions on Pattern Analysis and Machine
  Intelligence}, 45\penalty0 (12):\penalty0 15588--15603, 2023.
\newblock \doi{10.1109/TPAMI.2023.3307688}.

\bibitem[Grossmann et~al.(2023)Grossmann, Komorowska, Latz, and
  Sch{\"o}nlieb]{grossmann2023physicsinformed}
Tamara~G Grossmann, Urszula~Julia Komorowska, Jonas Latz, and Carola-Bibiane
  Sch{\"o}nlieb.
\newblock Can physics-informed neural networks beat the finite element method?
\newblock \emph{arXiv preprint arXiv:2302.04107}, 2023.

\bibitem[Guo et~al.(2022)Guo, Wu, Yu, and Zhou]{guo2022monte}
Ling Guo, Hao Wu, Xiaochen Yu, and Tao Zhou.
\newblock Monte carlo fpinns: Deep learning method for forward and inverse
  problems involving high dimensional fractional partial differential
  equations.
\newblock \emph{Computer Methods in Applied Mechanics and Engineering},
  400:\penalty0 115523, 2022.

\bibitem[Gupta et~al.(2021)Gupta, Xiao, and Bogdan]{gupta2021multiwavelet}
Gaurav Gupta, Xiongye Xiao, and Paul Bogdan.
\newblock Multiwavelet-based operator learning for differential equations.
\newblock In \emph{Advances in Neural Information Processing Systems},
  volume~34, pages 24048--24062, 2021.

\bibitem[Han et~al.(2018)Han, Jentzen, and E]{han2018solving}
Jiequn Han, Arnulf Jentzen, and Weinan E.
\newblock Solving high-dimensional partial differential equations using deep
  learning.
\newblock \emph{Proceedings of the National Academy of Sciences}, 115\penalty0
  (34):\penalty0 8505--8510, 2018.

\bibitem[Han et~al.(2020)Han, Nica, and Stinchcombe]{han2020derivative}
Jihun Han, Mihai Nica, and Adam~R Stinchcombe.
\newblock A derivative-free method for solving elliptic partial differential
  equations with deep neural networks.
\newblock \emph{Journal of Computational Physics}, 419:\penalty0 109672, 2020.

\bibitem[Henry-Labordere et~al.(2014)Henry-Labordere, Tan, and
  Touzi]{henry2014numerical}
Pierre Henry-Labordere, Xiaolu Tan, and Nizar Touzi.
\newblock A numerical algorithm for a class of bsdes via the branching process.
\newblock \emph{Stochastic Processes and Their Applications}, 124\penalty0
  (2):\penalty0 1112--1140, 2014.

\bibitem[Hermann et~al.(2020)Hermann, Sch{\"a}tzle, and
  No{\'e}]{hermann2020deep}
Jan Hermann, Zeno Sch{\"a}tzle, and Frank No{\'e}.
\newblock Deep-neural-network solution of the electronic schr{\"o}dinger
  equation.
\newblock \emph{Nature Chemistry}, 12\penalty0 (10):\penalty0 891--897, 2020.

\bibitem[Howe(1980)]{howe1980quantum}
Roger Howe.
\newblock Quantum mechanics and partial differential equations.
\newblock \emph{Journal of Functional Analysis}, 38\penalty0 (2):\penalty0
  188--254, 1980.

\bibitem[Huang and Russell(2010)]{huang2010adaptive}
Weizhang Huang and Robert~D Russell.
\newblock \emph{Adaptive moving mesh methods}, volume 174.
\newblock Springer Science \& Business Media, 2010.

\bibitem[Huang et~al.(2022)Huang, Ye, Liu, Ji, Wang, Yang, Li, Wang, CHU, Yu,
  Hua, Chen, and Dong]{huang2022metaautodecoder}
Xiang Huang, Zhanhong Ye, Hongsheng Liu, Shi~Bei Ji, Zidong Wang, Kang Yang,
  Yang Li, Min Wang, Haotian CHU, Fan Yu, Bei Hua, Lei Chen, and Bin Dong.
\newblock Meta-auto-decoder for solving parametric partial differential
  equations.
\newblock In \emph{Advances in Neural Information Processing Systems}, pages
  23426 -- 23438, 2022.

\bibitem[Irisov and Voronovich(2011)]{irisov2011numerical}
Vladimir Irisov and Alexander Voronovich.
\newblock Numerical simulation of wave breaking.
\newblock \emph{Journal of Physical Oceanography}, 41\penalty0 (2):\penalty0
  346--364, 2011.

\bibitem[Jin et~al.(2021)Jin, Cai, Li, and Karniadakis]{JIN2021109951}
Xiaowei Jin, Shengze Cai, Hui Li, and George~Em Karniadakis.
\newblock Nsfnets (navier-stokes flow nets): Physics-informed neural networks
  for the incompressible navier-stokes equations.
\newblock \emph{Journal of Computational Physics}, 426:\penalty0 109951, 2021.
\newblock ISSN 0021-9991.
\newblock \doi{https://doi.org/10.1016/j.jcp.2020.109951}.

\bibitem[Karlbauer et~al.(2022)Karlbauer, Praditia, Otte, Oladyshkin, Nowak,
  and Butz]{karlbauer2022composing}
Matthias Karlbauer, Timothy Praditia, Sebastian Otte, Sergey Oladyshkin,
  Wolfgang Nowak, and Martin~V Butz.
\newblock Composing partial differential equations with physics-aware neural
  networks.
\newblock In \emph{International Conference on Machine Learning}, pages
  10773--10801. PMLR, 2022.

\bibitem[Karniadakis et~al.(2021)Karniadakis, Kevrekidis, Lu, Perdikaris, Wang,
  and Yang]{karniadakis2021physics}
George~Em Karniadakis, Ioannis~G Kevrekidis, Lu~Lu, Paris Perdikaris, Sifan
  Wang, and Liu Yang.
\newblock Physics-informed machine learning.
\newblock \emph{Nature Reviews Physics}, 3\penalty0 (6):\penalty0 422--440,
  2021.

\bibitem[Kloeden and Platen(2011)]{kloeden2011numerical}
P.E. Kloeden and E.~Platen.
\newblock \emph{Numerical Solution of Stochastic Differential Equations}.
\newblock Stochastic Modelling and Applied Probability. Springer Berlin
  Heidelberg, 2011.
\newblock ISBN 9783540540625.

\bibitem[Kozubowski et~al.(2006)Kozubowski, Meerschaert, and
  Podgorski]{kozubowski2006fractional}
Tomasz~J Kozubowski, Mark~M Meerschaert, and Krzysztof Podgorski.
\newblock Fractional laplace motion.
\newblock \emph{Advances in Applied Probability}, 38\penalty0 (2):\penalty0
  451--464, 2006.

\bibitem[Kramer(2001)]{kramer2001review}
Peter~R. Kramer.
\newblock A review of some monte carlo simulation methods for turbulent
  systems.
\newblock \emph{Monte Carlo Methods and Applications}, 7\penalty0
  (3-4):\penalty0 229--244, 2001.

\bibitem[Krishnapriyan et~al.(2021)Krishnapriyan, Gholami, Zhe, Kirby, and
  Mahoney]{krishnapriyan2021characterizing}
Aditi Krishnapriyan, Amir Gholami, Shandian Zhe, Robert Kirby, and Michael~W
  Mahoney.
\newblock Characterizing possible failure modes in physics-informed neural
  networks.
\newblock In \emph{Advances in Neural Information Processing Systems},
  volume~34, pages 26548--26560, 2021.

\bibitem[Kurth et~al.(2023)Kurth, Subramanian, Harrington, Pathak, Mardani,
  Hall, Miele, Kashinath, and Anandkumar]{kurth2023fourcastnet}
Thorsten Kurth, Shashank Subramanian, Peter Harrington, Jaideep Pathak, Morteza
  Mardani, David Hall, Andrea Miele, Karthik Kashinath, and Anima Anandkumar.
\newblock Fourcastnet: Accelerating global high-resolution weather forecasting
  using adaptive fourier neural operators.
\newblock In \emph{Proceedings of the Platform for Advanced Scientific
  Computing Conference}, pages 1--11, 2023.

\bibitem[Larsson and Thom{\'e}e(2003)]{larsson2003partial}
Stig Larsson and Vidar Thom{\'e}e.
\newblock \emph{Partial differential equations with numerical methods},
  volume~45.
\newblock Springer, 2003.

\bibitem[Lee et~al.(2023)Lee, CHO, and Hwang]{lee2023hyperdeeponet}
Jae~Yong Lee, SungWoong CHO, and Hyung~Ju Hwang.
\newblock Hyperdeep{ON}et: learning operator with complex target function space
  using the limited resources via hypernetwork.
\newblock In \emph{International Conference on Learning Representations}, pages
  1--10, 2023.

\bibitem[Li et~al.(2021{\natexlab{a}})Li, Zhai, and Chen]{li2021neural}
Hong Li, Qilong Zhai, and Jeff~ZY Chen.
\newblock Neural-network-based multistate solver for a static schr{\"o}dinger
  equation.
\newblock \emph{Physical Review A}, 103\penalty0 (3):\penalty0 032405,
  2021{\natexlab{a}}.

\bibitem[Li et~al.(2023{\natexlab{a}})Li, Meidani, and
  Farimani]{li2023transformer}
Zijie Li, Kazem Meidani, and Amir~Barati Farimani.
\newblock Transformer for partial differential equations{\textquoteright}
  operator learning.
\newblock \emph{Transactions on Machine Learning Research}, pages 1--34,
  2023{\natexlab{a}}.

\bibitem[Li et~al.(2020)Li, Kovachki, Azizzadenesheli, Liu, Bhattacharya,
  Stuart, and Anandkumar]{li2020neural}
Zongyi Li, Nikola Kovachki, Kamyar Azizzadenesheli, Burigede Liu, Kaushik
  Bhattacharya, Andrew Stuart, and Anima Anandkumar.
\newblock Neural operator: Graph kernel network for partial differential
  equations.
\newblock \emph{arXiv preprint arXiv:2003.03485}, 2020.

\bibitem[Li et~al.(2021{\natexlab{b}})Li, Kovachki, Azizzadenesheli, Liu,
  Bhattacharya, Stuart, and Anandkumar]{DBLP:conf/iclr/LiKALBSA21}
Zongyi Li, Nikola~Borislavov Kovachki, Kamyar Azizzadenesheli, Burigede Liu,
  Kaushik Bhattacharya, Andrew~M. Stuart, and Anima Anandkumar.
\newblock Fourier neural operator for parametric partial differential
  equations.
\newblock In \emph{International Conference on Learning Representations}, pages
  1--16, 2021{\natexlab{b}}.

\bibitem[Li et~al.(2023{\natexlab{b}})Li, Huang, Liu, and
  Anandkumar]{li2022geofno}
Zongyi Li, Daniel~Zhengyu Huang, Burigede Liu, and Anima Anandkumar.
\newblock Fourier neural operator with learned deformations for pdes on general
  geometries.
\newblock \emph{Journal of Machine Learning Research}, 24\penalty0
  (388):\penalty0 1--26, 2023{\natexlab{b}}.

\bibitem[Li et~al.(2024)Li, Zheng, Kovachki, Jin, Chen, Liu, Azizzadenesheli,
  and Anandkumar]{li2021physics}
Zongyi Li, Hongkai Zheng, Nikola Kovachki, David Jin, Haoxuan Chen, Burigede
  Liu, Kamyar Azizzadenesheli, and Anima Anandkumar.
\newblock Physics-informed neural operator for learning partial differential
  equations.
\newblock \emph{ACM/IMS Journal of Data Science}, pages 1--27, feb 2024.

\bibitem[Lischke et~al.(2020)Lischke, Pang, Gulian, Song, Glusa, Zheng, Mao,
  Cai, Meerschaert, Ainsworth, et~al.]{lischke2020fractional}
Anna Lischke, Guofei Pang, Mamikon Gulian, Fangying Song, Christian Glusa,
  Xiaoning Zheng, Zhiping Mao, Wei Cai, Mark~M Meerschaert, Mark Ainsworth,
  et~al.
\newblock What is the fractional laplacian? a comparative review with new
  results.
\newblock \emph{Journal of Computational Physics}, 404:\penalty0 109009, 2020.

\bibitem[Liu et~al.(2020)Liu, Cai, and John~Xu]{CiCP-28-1970}
Ziqi Liu, Wei Cai, and Zhi-Qin John~Xu.
\newblock Multi-scale deep neural network (mscalednn) for solving
  poisson-boltzmann equation in complex domains.
\newblock \emph{Communications in Computational Physics}, 28\penalty0
  (5):\penalty0 1970--2001, 2020.
\newblock ISSN 1991-7120.

\bibitem[Lu et~al.(2021)Lu, Jin, Pang, Zhang, and Karniadakis]{lu2021learning}
Lu~Lu, Pengzhan Jin, Guofei Pang, Zhongqiang Zhang, and George~Em Karniadakis.
\newblock Learning nonlinear operators via deeponet based on the universal
  approximation theorem of operators.
\newblock \emph{Nature Machine Intelligence}, 3\penalty0 (3):\penalty0
  218--229, 2021.

\bibitem[Lu et~al.(2022)Lu, Blanchet, and Ying]{lu2022sobolev}
Yiping Lu, Jose Blanchet, and Lexing Ying.
\newblock Sobolev acceleration and statistical optimality for learning elliptic
  equations via gradient descent.
\newblock In \emph{Advances in Neural Information Processing Systems},
  volume~35, pages 33233--33247, 2022.

\bibitem[Maire and Tanré(2013)]{Maire2012MonteCA}
Sylvain Maire and Etienne Tanré.
\newblock Monte carlo approximations of the neumann problem.
\newblock \emph{Monte Carlo Methods and Applications}, 19\penalty0
  (3):\penalty0 201--236, 2013.

\bibitem[Mao(2007)]{mao2007stochastic}
Xuerong Mao.
\newblock \emph{Stochastic differential equations and applications}.
\newblock Elsevier, 2007.

\bibitem[Margenberg et~al.(2022)Margenberg, Hartmann, Lessig, and
  Richter]{margenberg2022neural}
Nils Margenberg, Dirk Hartmann, Christian Lessig, and Thomas Richter.
\newblock A neural network multigrid solver for the navier-stokes equations.
\newblock \emph{Journal of Computational Physics}, 460:\penalty0 110983, 2022.

\bibitem[Mattey and Ghosh(2022)]{mattey2022novel}
Revanth Mattey and Susanta Ghosh.
\newblock A novel sequential method to train physics informed neural networks
  for allen cahn and cahn hilliard equations.
\newblock \emph{Computer Methods in Applied Mechanics and Engineering},
  390:\penalty0 114474, 2022.

\bibitem[Mimeau and Mortazavi(2021)]{mimeau2021review}
Chlo{\'e} Mimeau and Iraj Mortazavi.
\newblock A review of vortex methods and their applications: From creation to
  recent advances.
\newblock \emph{Fluids}, 6\penalty0 (2):\penalty0 68, 2021.

\bibitem[Mitusch et~al.(2021)Mitusch, Funke, and Kuchta]{mitusch2021hybrid}
Sebastian~K Mitusch, Simon~W Funke, and Miroslav Kuchta.
\newblock Hybrid fem-nn models: Combining artificial neural networks with the
  finite element method.
\newblock \emph{Journal of Computational Physics}, 446:\penalty0 110651, 2021.

\bibitem[Navaneeth and Chakraborty(2023)]{navaneeth2023stochastic}
N~Navaneeth and Souvik Chakraborty.
\newblock Stochastic projection based approach for gradient free physics
  informed learning.
\newblock \emph{Computer Methods in Applied Mechanics and Engineering},
  406:\penalty0 115842, 2023.

\bibitem[Navaneeth et~al.(2024)Navaneeth, Tripura, and
  Chakraborty]{navaneeth2024physics}
N~Navaneeth, Tapas Tripura, and Souvik Chakraborty.
\newblock Physics informed wno.
\newblock \emph{Computer Methods in Applied Mechanics and Engineering},
  418:\penalty0 116546, 2024.

\bibitem[Nguyen-Thanh et~al.(2021)Nguyen-Thanh, Anitescu, Alajlan, Rabczuk, and
  Zhuang]{nguyen2021parametric}
Vien~Minh Nguyen-Thanh, Cosmin Anitescu, Naif Alajlan, Timon Rabczuk, and
  Xiaoying Zhuang.
\newblock Parametric deep energy approach for elasticity accounting for strain
  gradient effects.
\newblock \emph{Computer Methods in Applied Mechanics and Engineering},
  386:\penalty0 114096, 2021.

\bibitem[N{\"u}sken and Richter(2023)]{nusken2021interpolating}
Nikolas N{\"u}sken and Lorenz Richter.
\newblock Interpolating between bsdes and pinns--deep learning for elliptic and
  parabolic boundary value problems.
\newblock \emph{Journal of Machine Learning}, 2\penalty0 (1):\penalty0 31--64,
  2023.

\bibitem[Orsingher and DOvidio(2015)]{orsingher2015higher}
Enzo Orsingher and Mirko DOvidio.
\newblock Higher-order laplace equations and hyper-cauchy distributions.
\newblock \emph{Journal of Theoretical Probability}, 28:\penalty0 92--118,
  2015.

\bibitem[Pantidis and Mobasher(2023)]{pantidis2023integrated}
Panos Pantidis and Mostafa~E Mobasher.
\newblock Integrated finite element neural network (i-fenn) for non-local
  continuum damage mechanics.
\newblock \emph{Computer Methods in Applied Mechanics and Engineering},
  404:\penalty0 115766, 2023.

\bibitem[Pardoux and Peng(1992)]{pardoux1992backward}
Etienne Pardoux and Shige Peng.
\newblock Backward stochastic differential equations and quasilinear parabolic
  partial differential equations.
\newblock In \emph{Stochastic Partial Differential Equations and Their
  Applications}, pages 200--217. Springer, 1992.

\bibitem[Pardoux and Tang(1999)]{pardoux1999forward}
Etienne Pardoux and Shanjian Tang.
\newblock Forward-backward stochastic differential equations and quasilinear
  parabolic pdes.
\newblock \emph{Probability Theory and Related Fields}, 114\penalty0
  (2):\penalty0 123--150, 1999.

\bibitem[Paszke et~al.(2019)Paszke, Gross, Massa, Lerer, Bradbury, Chanan,
  Killeen, Lin, Gimelshein, Antiga, et~al.]{paszke2019pytorch}
Adam Paszke, Sam Gross, Francisco Massa, Adam Lerer, James Bradbury, Gregory
  Chanan, Trevor Killeen, Zeming Lin, Natalia Gimelshein, Luca Antiga, et~al.
\newblock Pytorch: An imperative style, high-performance deep learning library.
\newblock In \emph{Advances in Neural Information Processing Systems},
  volume~32, pages 8024--8035, 2019.

\bibitem[Pham(2015)]{pham2015feynman}
Huyen Pham.
\newblock Feynman-kac representation of fully nonlinear pdes and applications.
\newblock \emph{Acta Mathematica Vietnamica}, 40:\penalty0 255--269, 2015.

\bibitem[Qian et~al.(2022)Qian, S{\"u}li, and Zhang]{qian2022random}
Zhongmin Qian, Endre S{\"u}li, and Yihuang Zhang.
\newblock Random vortex dynamics via functional stochastic differential
  equations.
\newblock \emph{Proceedings of the Royal Society A}, 478\penalty0
  (2266):\penalty0 20220030, 2022.

\bibitem[Rahman et~al.(2023)Rahman, Ross, and Azizzadenesheli]{rahman2023uno}
Md~Ashiqur Rahman, Zachary~E Ross, and Kamyar Azizzadenesheli.
\newblock U-{NO}: U-shaped neural operators.
\newblock \emph{Transactions on Machine Learning Research}, pages 1--17, 2023.
\newblock ISSN 2835-8856.

\bibitem[Raissi et~al.(2019)Raissi, Perdikaris, and
  Karniadakis]{raissi2019physics}
Maziar Raissi, Paris Perdikaris, and George~E Karniadakis.
\newblock Physics-informed neural networks: A deep learning framework for
  solving forward and inverse problems involving nonlinear partial differential
  equations.
\newblock \emph{Journal of Computational Physics}, 378:\penalty0 686--707,
  2019.

\bibitem[Raissi et~al.(2020)Raissi, Yazdani, and
  Karniadakis]{doi:10.1126/science.aaw4741}
Maziar Raissi, Alireza Yazdani, and George~Em Karniadakis.
\newblock Hidden fluid mechanics: Learning velocity and pressure fields from
  flow visualizations.
\newblock \emph{Science}, 367\penalty0 (6481):\penalty0 1026--1030, 2020.

\bibitem[Richter and Berner(2022)]{richter2022robust}
Lorenz Richter and Julius Berner.
\newblock Robust sde-based variational formulations for solving linear pdes via
  deep learning.
\newblock In \emph{International Conference on Machine Learning}, pages
  18649--18666. PMLR, 2022.

\bibitem[Richter et~al.(2021)Richter, Sallandt, and
  N{\"u}sken]{richter2021solving}
Lorenz Richter, Leon Sallandt, and Nikolas N{\"u}sken.
\newblock Solving high-dimensional parabolic pdes using the tensor train
  format.
\newblock In \emph{International Conference on Machine Learning}, pages
  8998--9009. PMLR, 2021.

\bibitem[Rohsenow et~al.(1998)Rohsenow, Hartnett, Cho,
  et~al.]{rohsenow1998handbook}
Warren~M Rohsenow, James~P Hartnett, Young~I Cho, et~al.
\newblock \emph{Handbook of heat transfer}, volume~3.
\newblock Mcgraw-hill New York, 1998.

\bibitem[Ronneberger et~al.(2015)Ronneberger, Fischer, and
  Brox]{ronneberger2015u}
Olaf Ronneberger, Philipp Fischer, and Thomas Brox.
\newblock U-net: Convolutional networks for biomedical image segmentation.
\newblock In \emph{International Conference on Medical Image Computing and
  Computer-Assisted Intervention}, pages 234--241. Springer, 2015.

\bibitem[Rossi et~al.(2015)Rossi, Colagrossi, and Graziani]{rossi2015numerical}
Emanuele Rossi, Andrea Colagrossi, and Giorgio Graziani.
\newblock Numerical simulation of 2d-vorticity dynamics using particle methods.
\newblock \emph{Computers \& Mathematics with Applications}, 69\penalty0
  (12):\penalty0 1484--1503, 2015.

\bibitem[Samaniego et~al.(2020)Samaniego, Anitescu, Goswami, Nguyen-Thanh, Guo,
  Hamdia, Zhuang, and Rabczuk]{samaniego2020energy}
Esteban Samaniego, Cosmin Anitescu, Somdatta Goswami, Vien~Minh Nguyen-Thanh,
  Hongwei Guo, Khader Hamdia, Xiaoying Zhuang, and Timon Rabczuk.
\newblock An energy approach to the solution of partial differential equations
  in computational mechanics via machine learning: Concepts, implementation and
  applications.
\newblock \emph{Computer Methods in Applied Mechanics and Engineering},
  362:\penalty0 112790, 2020.

\bibitem[Sanchez-Gonzalez et~al.(2020)Sanchez-Gonzalez, Godwin, Pfaff, Ying,
  Leskovec, and Battaglia]{sanchez2020learning}
Alvaro Sanchez-Gonzalez, Jonathan Godwin, Tobias Pfaff, Rex Ying, Jure
  Leskovec, and Peter Battaglia.
\newblock Learning to simulate complex physics with graph networks.
\newblock In \emph{International Conference on Machine Learning}, pages
  8459--8468. PMLR, 2020.

\bibitem[Sawhney et~al.(2022)Sawhney, Seyb, Jarosz, and Crane]{sawhney2022grid}
Rohan Sawhney, Dario Seyb, Wojciech Jarosz, and Keenan Crane.
\newblock Grid-free monte carlo for pdes with spatially varying coefficients.
\newblock \emph{ACM Transactions on Graphics (TOG)}, 41\penalty0 (4):\penalty0
  1--17, 2022.

\bibitem[Seidman et~al.(2022)Seidman, Kissas, Perdikaris, and
  Pappas]{seidman2022nomad}
Jacob~H Seidman, Georgios Kissas, Paris Perdikaris, and George~J. Pappas.
\newblock {NOMAD}: Nonlinear manifold decoders for operator learning.
\newblock In \emph{Advances in Neural Information Processing Systems}, pages
  5601 -- 5613, 2022.

\bibitem[Shen et~al.(2021)Shen, Liu, He, Zhang, Xu, Yu, and
  Cui]{shen2021towards}
Zheyan Shen, Jiashuo Liu, Yue He, Xingxuan Zhang, Renzhe Xu, Han Yu, and Peng
  Cui.
\newblock Towards out-of-distribution generalization: A survey.
\newblock \emph{arXiv preprint arXiv:2108.13624}, 2021.

\bibitem[Shi et~al.(2022)Shi, Huang, Gao, Wei, Zhang, Bian, Yang, and
  Liu]{shi2022lordnet}
Wenlei Shi, Xinquan Huang, Xiaotian Gao, Xinran Wei, Jia Zhang, Jiang Bian, Mao
  Yang, and Tie-Yan Liu.
\newblock Lordnet: Learning to solve parametric partial differential equations
  without simulated data.
\newblock \emph{arXiv preprint arXiv:2206.09418}, 2022.

\bibitem[Smaoui et~al.(2021)Smaoui, El-Kadri, and Zribi]{smaoui2021control}
Nejib Smaoui, Alaa El-Kadri, and Mohamed Zribi.
\newblock On the control of the 2d navier--stokes equations with kolmogorov
  forcing.
\newblock \emph{Complexity}, 2021:\penalty0 1--18, 2021.

\bibitem[Sun et~al.(2022)Sun, Huang, Sun, and Wang]{sun2022bayesian}
Luning Sun, Daniel Huang, Hao Sun, and Jian-Xun Wang.
\newblock Bayesian spline learning for equation discovery of nonlinear dynamics
  with quantified uncertainty.
\newblock In \emph{Advances in Neural Information Processing Systems},
  volume~35, pages 6927--6940, 2022.

\bibitem[Tanyu et~al.(2023)Tanyu, Ning, Freudenberg, Heilenk{\"o}tter,
  Rademacher, Iben, and Maass]{tanyu2022deep}
Derick~Nganyu Tanyu, Jianfeng Ning, Tom Freudenberg, Nick Heilenk{\"o}tter,
  Andreas Rademacher, Uwe Iben, and Peter Maass.
\newblock Deep learning methods for partial differential equations and related
  parameter identification problems.
\newblock \emph{Inverse Problems}, 39\penalty0 (10):\penalty0 103001, 2023.

\bibitem[Tran et~al.(2023)Tran, Mathews, Xie, and Ong]{tran2023factorized}
Alasdair Tran, Alexander Mathews, Lexing Xie, and Cheng~Soon Ong.
\newblock Factorized fourier neural operators.
\newblock In \emph{International Conference on Learning Representations}, pages
  1--17, 2023.

\bibitem[Ummenhofer et~al.(2020)Ummenhofer, Prantl, Thuerey, and
  Koltun]{ummenhofer2019lagrangian}
Benjamin Ummenhofer, Lukas Prantl, Nils Thuerey, and Vladlen Koltun.
\newblock Lagrangian fluid simulation with continuous convolutions.
\newblock In \emph{International Conference on Learning Representations}, pages
  1--15, 2020.

\bibitem[Venturi and Casey(2023)]{venturi2023svd}
Simone Venturi and Tiernan Casey.
\newblock Svd perspectives for augmenting deeponet flexibility and
  interpretability.
\newblock \emph{Computer Methods in Applied Mechanics and Engineering},
  403:\penalty0 115718, 2023.

\bibitem[Wandel et~al.(2021{\natexlab{a}})Wandel, Weinmann, and
  Klein]{wandel2021learning}
Nils Wandel, Michael Weinmann, and Reinhard Klein.
\newblock Learning incompressible fluid dynamics from scratch - towards fast,
  differentiable fluid models that generalize.
\newblock In \emph{International Conference on Learning Representations}, pages
  1--15, 2021{\natexlab{a}}.

\bibitem[Wandel et~al.(2021{\natexlab{b}})Wandel, Weinmann, and
  Klein]{wandel2021teaching}
Nils Wandel, Michael Weinmann, and Reinhard Klein.
\newblock Teaching the incompressible navier--stokes equations to fast neural
  surrogate models in three dimensions.
\newblock \emph{Physics of Fluids}, 33\penalty0 (4):\penalty0 047117,
  2021{\natexlab{b}}.

\bibitem[Wandel et~al.(2022)Wandel, Weinmann, Neidlin, and
  Klein]{wandel2022spline}
Nils Wandel, Michael Weinmann, Michael Neidlin, and Reinhard Klein.
\newblock Spline-pinn: Approaching pdes without data using fast,
  physics-informed hermite-spline cnns.
\newblock In \emph{Proceedings of the AAAI Conference on Artificial
  Intelligence}, volume~36, pages 8529--8538, 2022.

\bibitem[Wang et~al.(2021)Wang, Wang, and Perdikaris]{wang2021learning}
Sifan Wang, Hanwen Wang, and Paris Perdikaris.
\newblock Learning the solution operator of parametric partial differential
  equations with physics-informed deeponets.
\newblock \emph{Science Advances}, 7\penalty0 (40):\penalty0 eabi8605, 2021.

\bibitem[Wang et~al.(2022{\natexlab{a}})Wang, Wang, and
  Perdikaris]{wang2022improved}
Sifan Wang, Hanwen Wang, and Paris Perdikaris.
\newblock Improved architectures and training algorithms for deep operator
  networks.
\newblock \emph{Journal of Scientific Computing}, 92\penalty0 (2):\penalty0 35,
  2022{\natexlab{a}}.

\bibitem[Wang et~al.(2022{\natexlab{b}})Wang, Yu, and Perdikaris]{wang2022and}
Sifan Wang, Xinling Yu, and Paris Perdikaris.
\newblock When and why pinns fail to train: A neural tangent kernel
  perspective.
\newblock \emph{Journal of Computational Physics}, 449:\penalty0 110768,
  2022{\natexlab{b}}.

\bibitem[Wang et~al.(2022{\natexlab{c}})Wang, Sun, Li, Lu, and
  Liu]{wang2022cenn}
Yizheng Wang, Jia Sun, Wei Li, Zaiyuan Lu, and Yinghua Liu.
\newblock Cenn: Conservative energy method based on neural networks with
  subdomains for solving variational problems involving heterogeneous and
  complex geometries.
\newblock \emph{Computer Methods in Applied Mechanics and Engineering},
  400:\penalty0 115491, 2022{\natexlab{c}}.

\bibitem[Weinan(2011)]{weinan2011principles}
E~Weinan.
\newblock \emph{Principles of multiscale modeling}.
\newblock Cambridge University Press, 2011.

\bibitem[Xiang et~al.(2022)Xiang, Peng, Liu, and Yao]{xiang2022self}
Zixue Xiang, Wei Peng, Xu~Liu, and Wen Yao.
\newblock Self-adaptive loss balanced physics-informed neural networks.
\newblock \emph{Neurocomputing}, 496:\penalty0 11--34, 2022.

\bibitem[Yang(1949)]{yang1949kinetic}
Li-Ming Yang.
\newblock Kinetic theory of diffusion in gases and liquids. i. diffusion and
  the brownian motion.
\newblock \emph{Proceedings of the Royal Society of London. Series A,
  Mathematical and Physical Sciences}, pages 94--116, 1949.

\bibitem[Zhang et~al.(2022)Zhang, Hu, Meng, Wang, Zhu, Chen, Ma, and
  Liu]{zhang2022drvn}
Rui Zhang, Peiyan Hu, Qi~Meng, Yue Wang, Rongchan Zhu, Bingguang Chen, Zhi-Ming
  Ma, and Tie-Yan Liu.
\newblock Drvn (deep random vortex network): A new physics-informed machine
  learning method for simulating and inferring incompressible fluid flows.
\newblock \emph{Physics of Fluids}, 34\penalty0 (10):\penalty0 107112, 2022.

\bibitem[Zhang et~al.(2024)Zhang, Meng, and Ma]{zhang2024deciphering}
Rui Zhang, Qi~Meng, and Zhi-Ming Ma.
\newblock Deciphering and integrating invariants for neural operator learning
  with various physical mechanisms.
\newblock \emph{National Science Review}, 11\penalty0 (4):\penalty0 nwad336,
  2024.

\bibitem[Zhang(2012{\natexlab{a}})]{zhang2012stochastic}
Xicheng Zhang.
\newblock Stochastic lagrangian particle approach to fractal navier-stokes
  equations.
\newblock \emph{Communications in Mathematical Physics}, 311\penalty0
  (1):\penalty0 133--155, 2012{\natexlab{a}}.

\bibitem[Zhang(2012{\natexlab{b}})]{zhang2012stochastic2}
Xicheng Zhang.
\newblock Stochastic functional differential equations driven by l{\'e}vy
  processes and quasi-linear partial integro-differential equations.
\newblock \emph{The Annals of Applied Probability}, 22\penalty0 (6):\penalty0
  2505--2538, 2012{\natexlab{b}}.

\bibitem[Zhao et~al.(2022)Zhao, Lindell, and Wetzstein]{qzhao2022graphpde}
Qingqing Zhao, David~B. Lindell, and Gordon Wetzstein.
\newblock Learning to solve pde-constrained inverse problems with graph
  networks.
\newblock In \emph{International Conference on Machine Learning}, pages
  26895--26910, 2022.

\bibitem[Zwicker(2020)]{py-pde}
David Zwicker.
\newblock py-pde: A python package for solving partial differential equations.
\newblock \emph{Journal of Open Source Software}, 5\penalty0 (48):\penalty0
  2158, 2020.
\newblock \doi{10.21105/joss.02158}.

\end{thebibliography}

\end{document}